\documentclass[preprint,12pt,authoryear]{elsarticle}

\usepackage{makecell}
\usepackage{amssymb}
\usepackage{xcolor}
\usepackage{hyperref}

\usepackage[flushleft]{threeparttable}

\journal{Expert Systems with Applications (Accepted)}

\begin{document}

\begin{frontmatter}



\title{Explanations Based on Item Response Theory (\textit{eXirt}):
A Model-Specific Method to Explain Tree-Ensemble Model in Trust Perspective}


% \author[1,2,3]{Anonymous}
% \author[1,2]{Anonymous}
% \author[4,5]{Anonymous}
% \author[2]{Anonymous}
% \author[2]{Anonymous}
% \author[1,2]{Anonymous}

%\author[1,2,3]{José de Sousa Ribeiro Filho\corref{aaa}}
%\ead{jose.ribeiro@ifpa.edu.br}
%\author[1,2]{Lucas Felipe Ferraro Cardoso}
%\ead{lucas.cardoso@pq.itv.org}
%\author[4,5]{Raíssa Lorena Silva da Silva}
%\ead{r.lorenna@gmail.com}
%\author[2]{Nikolas Jorge Santiago Carneiro}
%\ead{nikolas.carneiro@itv.org}
%\author[2]{Vitor Cirilo Araujo Santos}
%\ead{vitor.cirilo.santos@itv.org}
%\author[1,2]{Ronnie Cley de Oliveira Alves}
%\ead{ronnie.alves@itv.org}
\author[1,2,3]{José de Sousa Ribeiro Filho\corref{aaa}}
\ead{jose.ribeiro@ifpa.edu.br}
\author[1,2]{Lucas Felipe Ferraro Cardoso}
\ead{lucas.cardoso@icen.ufpa.br}
\author[4,5]{Raíssa Lorena Silva da Silva}
\ead{raissa.silva@inserm.fr}
\author[2]{Nikolas Jorge Santiago Carneiro}
\ead{nikolas.carneiro@itv.org}
\author[2]{Vitor Cirilo Araujo Santos}
\ead{vitor.cirilo.santos@itv.org}
\author[1,2]{Ronnie Cley de Oliveira Alves}
\ead{ronnie.alves@itv.org}

\affiliation[1]{organization={Federal University of Pará (UFPA), Postgraduate Program in Computer Science (PPGCC)},
            %addressline={},
            city={Belém},
            postcode={66075-10},
            state={Pará},
            country={Brazil}}

\affiliation[2]{organization={Vale Institute of Technology (ITV)},
            %addressline={},
            city={Belém},
            postcode={66055-090},
            state={Pará},
            country={Brazil}}

\affiliation[3]{organization={Federal Institute of Education, Science and Technology of Pará (IFPA)},
            %addressline={},
            city={Ananindeua},
            postcode={67125-000},
            state={Pará},
            country={Brazil}}

\affiliation[4]{organization={University of Montpellier},
            %addressline={},
            city={Montpellier},
            postcode={34090},
            state={Hérault},
            country={France}}

\affiliation[5]{organization={La Ligue Contre le Cancer},
            %addressline={},
            city={Montpellier},
            postcode={34000},
            state={Hérault},
            country={France}}

\cortext[aaa]{Corresponding Author. Tel.: +55 (91) 98185-3166}

% \affiliation[1]{organization={Anonymous},
%              %addressline={},
%              city={Anonymous},
%              postcode={Anonymous},
%              state={Anonymous},
%              country={Anonymous}}

% \affiliation[2]{organization={Anonymous},
%              %addressline={},
%              city={Anonymous},
%              postcode={Anonymous},
%              state={Anonymous},
%              country={Anonymous}}

% \affiliation[3]{organization={Anonymous},
%              %addressline={},
%              city={Anonymous},
%              postcode={Anonymous},
%              state={Anonymous},
%              country={Anonymous}}

% \affiliation[4]{organization={Anonymous},
%              %addressline={},
%              city={Anonymous},
%              postcode={Anonymous},
%              state={Anonymous},
%              country={Anonymous}}

% \affiliation[5]{organization={Anonymous},
%              %addressline={},
%              city={Anonymous},
%              postcode={Anonymous},
%              state={Anonymous},
%              country={Anonymous}}

\begin{abstract}

Solutions based on tree-ensemble models represent a considerable alternative to real-world prediction problems, but these models are considered black box, thus hindering their applicability in problems of sensitive contexts (such as: health and safety). Explainable Artificial Intelligence (XAI) aims to develop techniques that generate explanations of black box models, since these models are normally not self-explanatory. Methods such as \textit{Ciu, Dalex, Eli5, Lofo, Shap} and \textit{Skater} emerged with the proposal to explain black box models through global rankings of feature relevance, which based on different methodologies, generate global explanations that indicate how the model's inputs explain its predictions. This research aims to present an innovative XAI method, called \textit{eXirt}, capable of carrying out the process of explaining tree-ensemble models, based on Item Response Theory (IRT). In this context, 41 datasets, 4 tree-ensemble algorithms (\textit{Light Gradient Boosting, CatBoost, Random Forest}, and \textit{Gradient Boosting}), and 7 XAI methods (including \textit{eXirt}) were used to generate explanations. In the first set of analyses, the 164 ranks of global feature relevance generated by \textit{eXirt} were compared with 984 ranks of the other XAI methods present in the literature, being verified that the new method generated different explanations from other existing methods. In a second analysis, exclusive local and global explanations generated by \textit{eXirt} were presented that help in understanding the model trust, since in this explanation it is possible to observe particularities of the model regarding difficulty (if the model had difficulty predicting the test dataset), discrimination (if the model understands the test dataset as discriminative) and guesswork (if the model got the test dataset right by chance). Thus, it was verified that \textit{eXirt} is able to generate global explanations of tree-ensemble models and also local and global explanations of models through \textit{IRT}, showing how this consolidated theory can be used in machine learning in order to obtain explainable and reliable models.

\end{abstract}




\end{frontmatter}


\section{Introduction}\label{label_introducao}



%%\bibliographystyle{elsarticle-harv} 


%introdução básica
Technology has been evolving and today artificial intelligence is already a reality in the daily life of society. There are many real-world problems that machine learning algorithms solve, making human life more automated and intelligent \citep{shalev2014understanding,ghahramani2015probabilistic}.

%%\color{magenta}
%tree-ensemble
Machine learning models based on tree-structured bagging and boosting algorithms are known to provide high performance and high generalization capabilities, and thus being widely used in intelligent systems embedded in real-world problems \citep{maclin1997empirical_bagging_boosting,haffar2022explaining_ensemblerf_ai}. 

%Mesmo sendo popularmente utilizado em problemas das mais diferentes naturezas, os algoritmos baseados em tree-ensemble não contam com um elevado número de metodos de XAI capazes de criar explicações de suas predições, assim como as redes neurais por exemplo \citep{ibrahim2023explainable_cnn,samek2021explaining_dnn}
Even though they are popularly used in problems of the most different natures, tree-ensemble-based algorithms do not have a high number of XAI methods capable of creating explanations of their predictions, as well as neural networks, for example \citep{ibrahim2023explainable_cnn,samek2021explaining_dnn}
%%\color{black}
%
Tree-ensemble algorithms are not considered transparent\footnote{Transparent Algorithms: Algorithms that naturally generate explanations of how a particular output was produced. Such examples include Decision Tree, Logistic Regression, and K-Nearest Neighbors.}, their predictions are not self-explanatory, thus being considered black box algorithms\footnote{Black box algorithms: machine learning algorithms that have the steps of classification or regression decisions hidden from the user.} and, therefore, less used in problems related to sensitive contexts, such as health and safety, for example \citep{shojaei2023evolutionary_esa,ghosh2023role_esa,ribeiro_pred2town_et_al_2021}.

With the increasing need for high-performance models --- which implies low transparency \citep{arrieta_explainable_2019_20} --- in sensitive contexts, there is currently a growing need to develop methods or tools that can provide information about local explanations (feature relevance explanation generated around each data instances) and global explanations (when it is possible to understand the logic of all instances of the model generating in global way) as a means to make predictions more easily interpretable and also more trustworthy by humans \citep{guidotti2018survey,darpa_2019,xai_local_global_2020,lime_ref,wang2021trust_esa}.

%Efforts have been made by the community that researches Explainable Artificial Intelligence - XAI in developing different methods to explain black box models. Many of these efforts are defined in XAI methods that use the previously trained model, its input data, and its outputs in order to explain the model. This approach is called \textit{Post-hoc Explanation} \citep{molnar2020interpretable}.

In this regard, methods such as \textit{Ciu} \citep{ciu_ref}, \textit{Dalex} \citep{dalex_book}, \textit{Eli5} \citep{eli5_ref}, \textit{Lofo} \citep{lofo_ref}, \textit{Shap} \citep{xai_local_global_2020} e \textit{Skater} \citep{skater_ref} have emerged to promote the creation of model-agnostic and model-specific explanations. Note, a model-agnostic is a XAI method that it does not depend on type of model to be explained \citep{arrieta_explainable_2019_20}, and a model-specific is a XAI method that apply to a specific type of machine learning model \citep{khan2022model_specific}.

The main advantage of methods that use the model-agnostic approach is related to its independence related the type of model to explained. In other way, the main advantage of the model-specific approach is related the possibility of developing specific explanations for certain types of algorithms or even certain problems \citep{khan2022model_specific,molnar2020interpretable}.

It should be noted that each of the methods mentioned above is capable of explaining models using different techniques and methodologies, but one fact they have in common is that they all generate global relevance rankings of features related to the explanation of a model. And, therefore, are likely to have their results compared in quantitative way \citep{ribeiro_complexity_et_al_2021}.

The terminologies Feature Relevance Ranking and Feature Importance Ranking are widely used as synonyms in the computing community, but have different definitions herein, as shown in \citep{arrieta_explainable_2019_20}. Since feature rankings are regarded as ordered structures whereby each feature of the dataset used by the model appears in a position indicated by a score. The main difference being that, in relevance ranking, the calculation of the score is based on the model output, whereas to calculate the importance ranking of features, the correct label to be predicted is used \citep{arrieta_explainable_2019_20,molnar2020interpretable}.

%Thus, while feature relevance ranking acts as an explanation of how features contribute to reaching a particular model output, feature importance ranking acts as a performance method related to how features contribute to reaching the correct prediction \citep{arrieta_explainable_2019_20}.  

Global feature relevance ranking represents a  significant part of this study because they allow for general analyses of how a given model generalizes a specific problem, along with analyses of how a given methodology explains a specific model, without the need for a preliminary understanding of the context in which the model is embedded \citep{ribeiro_complexity_et_al_2021}.

Despite being limited, the global feature relevance ranks carry general explanations about the analyzed model, and for this reason they were selected as a basic structure of explanation to compare results of different XAI methods in a quantitative way, without the need to use knowledge of a human expert of the context of each analyzed problem \citep{molnar2020interpretable}. Because, in XAI there is no baseline definition for good or bad model explanations \citep{review_xai_2021}.

As shown in previous study \citep{ribeiro_complexity_et_al_2021}, explanations originating from different XAI methods may present specific similarities between themselves or also significant differences. This, considering the properties of the model to be explained and the particularities existing in each XAI method used. Given this fact, when there are several explanations for a set of models, the question naturally arises \textit{``Which model and explanation should I trust?''}.

%Além das explicações globais, existem as explicações locais que são criadas a nível de instância do dataset do modelo, permitindo um nível maior de entendimento de como um modelo realiza predições \citep{arrieta_explainable_2019_20}.
In addition to global explanations, there are local explanations that are created at the model dataset instance level, allowing a greater level of understanding of how a model performs predictions \citep{arrieta_explainable_2019_20}.

%Na literatura existem diferentes metodos voltados para explicação local,  como por exemplo: Local Surrogate \cite{lime_ref}, Scoped Rules \cite{ribeiro2018anchors_anchor}, Counterfactual explanations \cite{wachter2017counterfactual}, Shapley \cite{kernel_shap_ref}, and \textit{Explanation-by-example} \cite{keane2019case_explanation_by_example}. Dentre estes, destaca-se o último que é utilizado pela proposta defendida por esta pesquisa.
%In the literature there are different methods for local explanation, such as: \textit{Local Surrogate} \cite{lime_ref}, \textit{Scoped Rules} \cite{ribeiro2018anchors_anchor}, \textit{Counterfactual explanations} \cite{wachter2017counterfactual}, \textit{Shapley} \cite{kernel_shap_ref}, and \textit{Explanation-by-example} \cite{keane2019case_explanation_by_example}. Among these, the last one is highlighted, which is used by the proposal defended by this research.

%\textit{Explanation-by-example} é uma técnica que identifica instâncias específicas do dataset do modelo e gera explicações que ajudam no entendimento da predição \cite{molnar2020interpretable}. Valendo-se ressaltar, que este método se mostra como um caminho viavel para a criação de explicações que forneçam insights sobre como um humano pode confiar em uma predição do modelo, ou mesmo no modelo como um todo \cite{lime_ref}. 
\textit{Explanation-by-example} is a type of model explanation technique focused on local instances of significant examples from a dataset, which through specific techniques produce explanations that help in the process of interpreting model predictions by a human \citep{molnar2020interpretable}. It is worth noting that this method is a viable way to create explanations that provide insights into how a human can trust a model prediction, or even the model as a whole \citep{lime_ref,cardoso2022explanationbyexample_irt}.

%Focando-se nos algoritmos tree-ensemble, esta pesquisa identifica a necessidade e a oportunidade de criar um método model-specific, capaz de gerar explicações globais e locais visando uma maior confiabilidade no modelo \citep{chatzimparmpas2020state_trust,lime_ref}. Com isso, surge a necessidade de se ter uma forma avaliativa de modelos diferenciadas dos demais métodos de XAI existentes.
Focusing on tree-ensemble algorithms, this research identifies the need and opportunity to create a model-specific method, capable of generating global and local explanations aiming for greater reliability in the \citep{chatzimparmpas2020state_trust,lime_ref} model. With this, there is a need to have a way of evaluating models that is different from other existing XAI methods.

%This research understands that the interesting way to develop the XAI method with the desired differential in the context presented above is to base the method on \textit{Item Response Theory - IRT}\citep{de2000teoria_irt_ref}, since the exclusive properties for evaluating the model that this theory presents, will be able to answer questions about the confidence that a model and its explanation can present, similar to the one developed in \cite{cardoso2022explanationbyexample_irt} --- which presents an interesting \textit{IRT}-based strategy for producing \textit{Explanation-by-example} using a pool of diverse models.

The \textit{Item Response Theory}, is a very widespread theory, generally used in the process of evaluating candidates in selection processes. The theory uses the properties ``discrimination'', ``difficulty'' and ``guessing'' to enable evaluation of latent characteristics, which cannot be observed directly, of the responses of candidates in a selection process. This is intended to establish the relationship of hit probability to the candidate's ability \citep{de2000teoria_irt_ref}.

This research proposes a new method for explaining tree-ensemble models based on Item Response Theory, called \textit{eXirt}. Seeking to validate this method, global feature relevance ranks were generated for models created from 4 different algorithms (\textit{Light Gradient Boosting, CatBoost, Random Forest}, and \textit{Gradient Boosting}) and 41 different datasets (binary classification), which were compared to the results of 6 XAI methods already known in the literature, aiming to show similarities and differences in several contexts of problems. Then, analyzes of local explanations uniquely generated by the \textit{eXirt} method are also presented, which provide insights on how to trust the analyzed models.

%Toda a metodologia apresentada este artigo, assim como a estratégia em torno da proposta de método de XAI aqui apresentada se baseiam em artigos previamente publicados, que são \citep{ribeiro_complexity_et_al_2021,cardoso2022explanationbyexample_irt}. 
This research is the continuation of previous studies carried out in: \textit{``Expla-nation-by-Example Based on Item Response Theory''}\citep{cardoso2022explanationbyexample_irt}, \textit{``Does Dataset Complexity Matters for Model Explainers?''}\citep{ribeiro_complexity_et_al_2021} and \textit{``Decoding Machine Learning Benchmarks''}\citep{cardoso2020decoding_irt}, which have already been duly published.

The main contributions to the studies in Explainable Artificial Intelligence that this research generates are as follows:

\begin{itemize}
  \item An innovative XAI method, called \textit{eXirt}, which is based on \textit{Item Response Theory}, an interesting theory still under-explored in machine learning;
  
  \item Innovative explanations of tree-ensemble models generated by the \textit{eXirt} method, capable of generating global feature relevance ranks based in \textit{IRT}, along with local information on model discrimination, difficulty and guessing, enabling unique insights into its reliability;
  
  \item Comparisons of the features relevance ranks generated by the \textit{eXirt} method with the results generated by the \textit{Ciu, Dalex, Eli5, Lofo, Shap} and \textit{Skater} methods.
\end{itemize}

This article is organized in follows sections: Section \ref{label_trabalhos_relacionados} presents work related to this research; Section \ref{label_background} introduces key background concepts about XAI and \textit{IRT}; Section \ref{label_materiais_metodos} presents the methodological and conceptual part on which the experiments and the central proposal of the article are based; Section \ref{label_resultados} results of analysis involving identification of dataset profiles, and the production of global and local explanations; Section \ref{label_conclusoes} Conclusions of paper; Section \ref{label_trabalhos_furutos} Presentation of future works.

%%\color{blue}
\section{Related Works}\label{label_trabalhos_relacionados}

%newnewnew
%Nesta seção serão apresentados trabalhos relacionados a pesquisa que dão suporte as análises comparativas realizadas entre os diferentes métodos de XAI, são apresentados referências quanto a taxonomia utilizada, e faz a apresentação de um levantamento bibliográfico sobre os principais métodos de de XAI presentes na literatura.   
This section will present: The research-related works that support the comparative analyzes carried out between the different XAI methods (subsection \ref{lab_analisis_involvin_xai}); And references regarding the taxonomy used, and a bibliographical survey will be presented on the main XAI methods present in the literature (subsection \ref{lab_xai_literatura}).

\subsection{Analysis involving XAI methods}\label{lab_analisis_involvin_xai}
%newnewnew
%Nos últimos anos, observa-se um crescente interesse pelo tema Explainable Artificial Intelligence por parte da sociodade, uma prova disso é a crescente quantidade de publicações sobre estes temas em repositórios de pesquisa como \citep{vine2006googlescholar_repository,hunter1998sciencedirect_repository,durniak2000welcome_ieeexplore_repository} ou mesmo o volumes de buscas na internet sobre o tema \citep{coogle_trends}.
In recent years, there has been a growing interest in the topic of Explainable Artificial Intelligence among society, proof of this is the growing number of publications on these topics in research repositories such as \citep{vine2006googlescholar_repository,hunter1998sciencedirect_repository,durniak2000welcome_ieeexplore_repository} or even volumes of internet searches on the topic \citep{coogle_trends}.

%newnewnew
%Um dos motivos que impulsionam os interesses por esta área de pesquisa, é a presença cada vez maior de sistema inteligentes no cotidiano da vida humana, como por exemplo os sistemas voltados para cidades inteligentes, casas inteligentes, controle de tráfego, controle de saúde humana, segurança pública, educação, finanças entre outros \citep{shalev2014understanding,ghahramani2015probabilistic}.
One of the reasons for the interest in this area of research is the increasing presence of intelligent systems in everyday human life, such as systems aimed at: Smart Cities, Smart Homes, Traffic Control, Human Health Control, Public Security , Education, Economy, among others \citep{shalev2014understanding,ghahramani2015probabilistic}.

%newnewnew
%Com isso, a área de XAI vem ganhando cada vez mais atenção, cria-se a necessidade cada vez maior de se explicar os sistemas citados, principalmente pelo fato de que na maioria dos casos, estes sistemas são baseados em algoritmos caixa-preta \citep{arrieta_explainable_2019_20}.
As a result, the area of XAI is gaining more attention, creating an increasing need to explain the aforementioned systems, mainly due to the fact that in most cases, these systems are based on black box algorithms, due their high performances \citep{arrieta_explainable_2019_20}.

%newnewnew
%Com a rápida ascensão da áre de XAI, fica notável diferenças taxonômicas do uso dos termos utilizados por diferentes trabalhos. Buscando uniformizar o entendimento dos termos taxonômicos utilizados aqui, optou-se por adotar a taxonomia utilizada no trabalho \citep{arrieta_explainable_2019_20} com principal base para esta pesquisa.
With the rapid rise of the XAI area, the taxonomic differences used by different works are notable. Seeking to standardize the understanding of the taxonomic terms used here, we chose to adopt the taxonomy used in the work \cite{arrieta_explainable_2019_20} as the main basis for this research.

%A pesquisa utilizada com base de definição para os diferentes tipos de métodos XAI que geram explicações globais model-agnostic e model-specific foi a \citep{molnar2020interpretable}, que apresenta a divisão de tipos/formas de explicações \textit{Partial Dependence Plot (PDP)}, \textit{Accumulated Local Effects (ALE), Feature Interaction, Global Surrogate, Prototypes and Cristicisms,} and \textit{Permutantion Feature Relevance (Importance)}, sendo este último o foco da pesquisa aqui descrita.

%Como pode-se notar acima, existem pelo menos sete tipos diferentes de explicações model-agnostic e model-specific de modelo de aprendizagem de máquina, sendo que estes tipos de explicações surgiram através do desenvolvimento de novas pesquisas envolvendo a apresentação de métodos XAI, como por exemplo o popular \citep{tree_shap_ref,kernel_shap_ref}.

%newnewnew
%Com o crescente surgimento de novas propostas que buscam explicar modelos caixa-preta, uma necessidade acaba surgindo neste meio, que é a comparação quantitativa entre estes métodos de XAI, pelo fato de se precisar entender se tais explicações são diferentes entre os modelos ou mesmo similares \citep{sahatova2022overview_comparacao_shap_lime,chadaga2023artificial_compare,hariharan2023xai_compare,jouis2021_anchors}.
With the increasing emergence of new proposals that seek to explain black box models, a need arises in this context, which is the quantitative comparison between these XAI methods, due to the need to understand whether such explanations are different between the models or even similar \citep{sahatova2022overview_comparacao_shap_lime,chadaga2023artificial_compare,hariharan2023xai_compare,jouis2021_anchors}.

%newnewnew
%Definir se a explicação de um modelo é boa ou ruim é um tarefa ainda mais difícil, pois discussões na área de XAI já mostram que não existe uma explicação que pode atender as expectativas de todos indivíduos envolvidos em um contexto/problema, ou seja, não existe um baseline quanto a explicação ideal que pode ser gerada de um modelo \citep{peterflach_humanintheloop}.
Defining whether a model's explanation is good or bad is a complicated task, as discussions in the area of XAI already show that there is no explanation that can meet the expectations of all individuals involved in a context/problem, that is, there is no baseline as to the ideal explanation that can be generated from a model \citep{peterflach_humanintheloop}.

%newnewnew
%Neste sentido, pesquisas que buscam realizar a proposição de novos métodos de XAI, na maioria das vezes não chegam a comparar resultados de seus métodos propostos com os resultados de outros métodos, voltando-se na maioria das vezes as descrições dos embasamentos teóricos dos quais seus métodos se baseiam \citep{tree_shap_ref, lofo_ref,eli5_ref, dalex_book}.
In this sense, research that seeks to propose new XAI methods, most of the time does not compare the results of their proposed methods with the results of other methods, most often focusing on descriptions of the theoretical basis of which their methods are based on \citep{tree_shap_ref, lofo_ref,eli5_ref, dalex_book}.

%newnewnew
%Pesquisas que buscam comparar diferentes métodos de XAI, baseados em ranque de relevância, é algo novo e ainda pouco explorado pela comunidade de machine learning \citep{holzinger2020explainable,krishna2022disagreement}. Pois, os desafios em torna da comparação dos métodos são significativos e a não existência de um baseline torna o problema ainda maior, como visto em \citep{ribeiro_complexity_et_al_2021}. Mesmo assim, pesquisas envolvendo comparações vem surgindo, porém com restrições a um único dataset e alguns poucos modelos que podem ser criados a partir destes dados \citep{sahatova2022overview_comparacao_shap_lime,chadaga2023artificial_compare,hariharan2023xai_compare,jouis2021_anchors}.
Research that seeks to compare different XAI methods, based on relevance ranking, is something new and still little explored by the machine learning community \citep{holzinger2020explainable,krishna2022disagreement}. Therefore, the challenges involved in comparing methods are significant and the lack of a baseline makes the problem even greater, as seen in \citep{ribeiro_complexity_et_al_2021}. Even so, research involving comparisons has been emerging, but with restrictions on a single dataset and a few models that can be created from this data \citep{sahatova2022overview_comparacao_shap_lime,chadaga2023artificial_compare,hariharan2023xai_compare,jouis2021_anchors}.

%newnewnew
%Por exemplo, em \cite{sahatova2022overview_comparacao_shap_lime} são feitas comparações visuais das explicações locais produzidas pelos métodos \textit{Shap} e \textit{Lime} \citep{lime_ref} sob um modelo de aprendizagem de máquina voltado para detecção de objetos em imagens de tumografias. As principais comparações realizadas entre os métodos são de natureza visual, já que os dados de entrada do modelo são em formato de imagem
For example, in \cite{sahatova2022overview_comparacao_shap_lime} visual comparisons are made of the local explanations produced by the methods \textit{Shap} and \textit{Lime} \citep{lime_ref} under a machine learning model aimed at detecting objects in images of tumorography. The main comparisons made between the methods are visual in nature, as the model input data is in image format.

%newnewnew
%No estudo \cite{chadaga2023artificial_compare}, é feita a criação de um sistema de apoio a decisão que utiliza os métodos \textit{Shap}, \textit{Eli5}, \textit{QLattice}, \textit{Anchor}, and \textit{Lime} com o objetivo de criar explicações de um modelo de aprendizagem de máquina voltado para o diagnóstico de COVID 2019. Aqui diferentes tipos (formatos) de explicações são gerados e são usados para discussãode diferentes perspectivas do problema proposto.
In the study \cite{chadaga2023artificial_compare}, a decision support system is created based on the methods \textit{Shap}, \textit{Eli5}, \textit{QLattice}, \textit{Anchor}, and \textit{ Lime} with the aim of creating explanations of a machine learning model aimed at diagnosing \textit{COVID 2019}. Here different types/formats of explanations are generated and are used to discuss different perspectives of the proposed problem.

%newnewnew
%Na pesquisa \cite{hariharan2023xai_compare} é feita a utilização dos métodos de \textit{Permutation Importance}, \textit{Shap}, \textit{Lime}, and \textit{Ciu} sob um modelo de aprendizagem de máquina voltado ao problema de Detecção de Intrusão voltado para segurança cibernética. Neste estudo, são realizadas comparações das explicações por meio de acurácia, consistência e estabilidade.
In the research \cite{hariharan2023xai_compare} the methods of \textit{Permutation Importance}, \textit{Shap}, \textit{Lime}, and \textit{Ciu} are used under a machine learning model aimed at the problem of Intrusion Detection focused on cybersecurity. In this study, comparisons of explanations are made using accuracy, consistency and stability.

%newnewnew
%Em \cite{jouis2021_anchors} são utilizados os métodos \textit{Anchors} and \textit{Attention} sob um modelo de predição de ofertas de emprego. No estudo, são usados diversos tipos (formatos) de  explicações devidamente comparadas de maneira quantitativa e qualitativas.
In \cite{jouis2021_anchors} the \textit{Anchors} and \textit{Attention} methods are used under a job offer prediction model. In the study, different types and formats of explanations are used, duly compared in a quantitative and qualitative way.

%newnewnew
%Diferenciando-se de pesquisas existentes na atual literatura, esta pesquisa utiliza de descrições de conceitos teóricos e da comparação quantitativa de resultados, para apresentar e validar um novo método de XAI chamado \textit{eXirt}. Para isso, descreveu-se nas sessões seguintes os principais conceitos em tordo de XAI, principais conceitos da Teoria de Resposta ao Item, ainda mais a frente são apresentados resultados envolvendo um número elevado de dados e modelos tree-ensemble.
Differentiating itself from existing research in the current literature, this research uses descriptions of theoretical concepts and quantitative comparison of results, to present and validate a new XAI method called \textit{eXirt}. To this end, the main concepts surrounding XAI and the main concepts of Item Response Theory were described in the following sessions. Further on, results involving a large number of data and tree-ensemble models are presented.

\subsection{XAI methods in the literature}\label{lab_xai_literatura}

%newnewnew
%Esta pesquisa fez uma segunda pesquisa bibliográfica, foca na identificação de pesquisa voltadas a proposições dos métodos de XAI existentes, assim it was possible to find the main XAI methods specifically aimed at generating global feature relevance rankings in a model-agnostic and model specific manner that support tabular data.
This research carried out a second bibliographical search, focusing on identifying research focused on propositions of existing XAI methods, so it was possible to find the main XAI methods specifically aimed at generating global feature relevance rankings in a model-agnostic and model specific manner that support tabular data.

As a result, a total of six XAI methods were found to be properly validated and compatible with one another (at library and code execution dependencies  level). These methods include: \textit{CIU} \citep{ciu_ref}, \textit{Dalex} \citep{dalex_python_ref}, \textit{Eli5} \citep{eli5_ref}, \textit{Lofo} \citep{lofo_ref}, \textit{SHAP} \citep{xai_local_global_2020} and \textit{Skater} \citep{skater_ref}.

This survey found other tools aimed at model explanation, including: \textit{Alibi} \citep{alibi_ale_ref}, \textit{Lime} \citep{lime_ref}, \textit{IBM Explainable AI 360} \citep{ibm_xai360}, \textit{Anchor} \citep{anchors_aaai18}, \textit{Attention} \citep{lin2017structured_attention} e \textit{Interpreter ML} \citep{interpretML_arxiv}. However, due to incompatibilities and technical problems, they ended up not being used in this research.

%\color{blue}
The main problems and incompatibilities found were: No generation of global ranks; Rank generation based on another XAI method already existing in the pipeline; Non-compatibility with pipeline dependencies (at library version level); Lack of method library updates; No support for the \textit{Light Gradient Boosting, CatBoost, Random Forest} and \textit{Gradient Boosting} models. 
%\color{black}

Due to incompatibilities between XAI methods libraries and versions of machine learning model libraries and dependencies \textit{scikit-lean,scikit-learn}, only the \textit{Light Gradient Boosting, CatBoost, Random Forest} and \textit{Gradient Boosting} algorithms were used herein as models to be explained by the aforementioned methods. In other words, this problem is directly linked to the way these tools were programmed rather than the methodologies they advocate. 

Note that the six methods presented herein generate relevance rankings based on the same previously trained machine learning models (with the same training and testing split), manipulate their inputs and/or produce new intermediate models copies. Therefore, they are required to be compatible with each other so that a fair comparison of their final rankings of explanations can be made.

Table \ref{tab_resumo_xai} shows a general comparison between the techniques found during bibliographic research.

% \usepackage{graphicx}

\begin{table*}[!h]
%%\color{blue}
\centering
\caption{Researched XAI methods}
\resizebox{.99\textwidth}{!}{%
\begin{tabular}{c|c|c|c|c|c|c}
\hline
\textbf{\begin{tabular}[c]{@{}c@{}}Name\end{tabular}} & 
\textbf{Base algorithm} &
\textbf{\makecell{Explanation\\ technique}} & 
\textbf{\makecell{Global \\ explanation\\ (by rank)}} &
\textbf{\makecell{Local \\ explanation}} &
\textbf{\makecell{Model\\ Specific or \\ Agnostic?}} &
\textbf{\makecell{Compatible?}} \\ \hline
\textit{Alibi} &  \makecell{Out-of-bag\\error} & \makecell{Feature\\Permutation\\and accuracy and f1} & Yes & Yes & Agnostic & No \\ \hline
\textit{Anchor} &  \makecell{if-Then\\Rules} & \makecell{Rules} & No & Yes & Agnostic & No \\ \hline
\textit{Attention} &  \makecell{Structured\\Self-attentive\\embedding} & \makecell{Multiple\\Vector\\Representations} & No & Yes & Specific & No \\ \hline
\textit{CIU} &  \makecell{Decision\\Theory} & \makecell{Feature Permutation \\and Multiple Criteria\\Decision Making} & Yes & No & Agnostic & Yes \\ \hline
\textit{Dalex} & \makecell{Leave-one\\ covariate out} & Feature Permutation & Yes & Yes & Agnostic & Yes \\ \hline
\textit{Eli5} & \makecell{Assigning \\weights\\ to decisions} & \makecell{Feature Permutation\\and Mean Decrease\\ Accuracy} & Yes & Yes & Specific & Yes \\ \hline
\textit{eXirt} & \makecell{Item\\ Response\\ Theory} & \makecell{Feature Permutation\\and Model Ability} & Yes & Yes & Specific & Yes \\ \hline
\textit{\makecell{IBM\\Explainable\\AI 360}} &  \makecell{Same of \textit{Shap}} & \makecell{Same of \textit{Shap}} & Yes & Yes & Specific & No \\ \hline
\textit{\makecell{Interpreter\\ML}} &  \makecell{Same of\\\textit{Lime} and \textit{Shap}} & \makecell{Same of\\\textit{Lime} and \textit{Shap}} & Yes & Yes & \makecell{Specific and\\ Agnostic} & No \\ \hline
\textit{Lime} &  \makecell{local linear\\approximation} & \makecell{Perturbation of\\the Instance} & No & Yes & Agnostic & No \\ \hline
\textit{Lofo} & \makecell{Leave One\\ Feature Out} & Feature Permutation & Yes & No & Specifc & Yes \\ \hline
\textit{Shap} & \makecell{Game\\Theory} & Feature Permutation & Yes & Yes & Specific & Yes \\ \hline
\textit{Skater} & \makecell{Information\\ Theory} & Feature Relevance & Yes & Yes & Agnostic & Yes \\ \hline
\end{tabular}
}
\label{tab_resumo_xai}
\end{table*}

Note, the \textit{eXirt} in table \ref{tab_resumo_xai}, explainable base in \textit{Item Response Theory}, is the XAI method defended by this article, it appears prematurely here only at the level of global comparison with the other methods.

Still in table \ref{tab_resumo_xai}, it can be seen that most existing XAI methods use the ``Feature Permutation'' technique to perform the model explanation process. However, it should be emphasized at this moment, that the \textit{eXirt} differs from other methods by having base in \textit{IRT}.

%newnewnew
%Detalhe importante, até a data da produção deste artigo não foi encontrado um método de XAI capaz de explicar modelos de aprendizagem de máquina por meio de explicações globais baseadas em ranque de relevância de atributos utilizando como base a Teoria de resposta ao Item.
An important detail, at the time of this article's production, an XAI method capable of explaining machine learning models through global explanations based on attribute relevance ranking using \textit{Item Response Theory} as a basis was not found.

%%\color{black}
%%\color{orange}
%\color{blue}
\section{Background}\label{label_background}

%Nesta seção, são apresentados conceitos e taxonomias relacionadas aos assuntos  ``Explainable Artificial Intelligence'' e ``Teoria de Resposta ao Item'',  que serão utilizadas ao decorrer deste trabalho.
In this section, concepts, taxonomies and theories related to Explainable Artificial Intelligence (subsection \ref{lab_xai}) and \textit{Item Response Theory} (subsection \ref{lab_irt}) are presented, which will be used throughout this work.

\subsection{Explainable Artificial Intelligence} \label{lab_xai}

The so-called post-hoc explanation is the currently most widely used existing XAI method category in the computing community. The post-hoc techniques can be divided into different strategies: \textit{Text Explanations}, \textit{Visual Explanations}, \textit{Local Explanations}, \textit{Explanation-by-simplification}, \textit{Feature Relevance Explanations} and \textit{Explanation-by-example}. Among these six types of techniques, this research is based on the last two, \textit{Feature Relevance Explanation} and \textit{Explanation-by-example}, as they are properly used by the XAI method proposed here.

\subsubsection{Feature Relevance Explanation}

The \textit{CIU}, \textit{Dalex}, \textit{Eli5}, \textit{Lofo}, \textit{SHAP} and \textit{Skater} are capable of generating several types of model explanations --- however, for quantitative comparisons, only the ranking generation processes are described. %However, since one focus of this article is to compare in quantitative way this important structure of model explanation, the rank, it is restricted to comparing only this type of result with the result with the proposed method.
%De maneira a facilitar o entendimento de como o cada métod XAI listado acima gera as feature relevances explanations, a seguir são apresentadas explicações sobre seus funcionamentos básicos.
Thus, in order to facilitate the understanding of how each XAI method listed above generates the feature relevance explanations, descriptions about their basic operations are presented below.

%\subsubsection{\textit{CIU}\\}

The \textit{Contextual Importance and Utility (CIU)} is a XAI method based on Decision Theory \citep{decisions_1993} that focuses on serving as a unified metric of model-agnostic explainability based on the implementation of two different scores: \textit{Contextual Importance (CI)} and \textit{Contextual Utility (CU)} \citep{ciu_ref}.

%A Importância está ligada a ideia de se mensurar, através de diferentes valores, quanto uma feature é importante frente a um contexto específico geral do dataset usada como base para o modelo.

%removido
%Contextual Importance is linked to the idea of measuring, through different values, how important a feature is in a specific general context of the dataset used as a basis for the model \citep{ciu_git}.

%A Utilidade está ligada a ideia de se mensurar, através de diferentes valores de utilidade, quanto um atributo é utilizado por diferentes contextos menores presentes no dado que o dataset se basea.

%removido
%Contextual Utility is linked to the idea of measuring, through different utility values, how much an attribute is used by different smaller contexts present in the data on which the dataset is based \citep{ciu_git}.

As verified in preliminary tests carried out by this research, these two indices generate equal ranks. Thus, it was decided to use the \textit{CI} \citep{ciu_ref}, since its definition shows that this score is more general to the context of the model data \citep{ciu_ref}.

%\subsubsection{\textit{Dalex}\\}

The method \textit{Dalex} is a set of XAI tools based on the \textit{LOCO} (\textit{Leave one Covariate Out}) approach and can generate explainabilities from this approach. This method receives the model and the data to be explained, calculates model performance, performs new training processes with new generated data sets, and makes the inversion of each feature of the data in a unitary and iterative way, methods what features are important to the model, evaluates its performance obtained according to the inversions of the features \citep{dalex_python_ref,dalex_book}.

%As inversões de features realizadas por este método de XAI são literais ao se observar um dataset --- ou seja a feature possue seus índices invertidos ---, pois por exemplo: em um dataset que possui as features $F1$ e $F2$ juntamente com seus respectivos valores V1, V2, V3, ..., Vn e V1, V2, V3, ..., Vn, ao se inverter os valores presentes em F1 os seus valores são Vn, Vn-1, Vn-2, ...,V1.

%removido
%The feature inversions performed by this XAI method are literal when observing a dataset --- that is, the feature has its indexes inverted ---, for example: in a dataset that has the features $F1$ and $F2$ together with their respective values $V_1, V_2, V_3, ..., V_n$ and $V_1, V_2, V_3, ..., V_n$, when inverting the values present in $F_1$ their values are $V_n, V_{n-1}, V_{n-2} , ...,V_1$ \citep{dalex_python_ref}.

%\subsubsection{\textit{Lofo}\\}

A little less popular but very powerful is the \textit{Leave One Feature Out (Lofo)}, a XAI method with a similar proposal to that of \textit{Dalex}, but no feature inversion is performed here, because in the \textit{Lofo} metric the iterative step is based on iterative removal of the features to find its global relevance to the model. This method also analyzes the performance of the model \citep{lofo_ref}.

%O LOFO em um primeiro passo avalia o desempenho do modelo com todos as features de entrada incluídos, posteriormente remove iterativamente um recurso de cada vez, treina novamente o modelo e avalia seu desempenho em um dataset específico de validação. A média e o desvio padrão da importância de cada feature são então relatados.
\textit{Lofo} initially evaluates the model's performance with all input features included, then iteratively removes one feature at a time, retrains the model and evaluates its performance on a specific validation dataset. The mean and standard deviation of the relevance of each feature are then reported \citep{lofo_ref}.

%Esta pesquisa optou por utilizar a versão \textit{Fast Lofo}, que executa o mesmo algoritmo presente no \textit{Lofo} converncional, porém sem necessáriamente treinar novos modelos. Isto foi necessário devido seu menor custo computacional da execução da versão \textit{Fast Lofo}.
%This research chose to use the \textit{Fast Lofo} version, which runs the same algorithm present in the conventional \textit{Lofo}, but without necessarily training new models. This was necessary due to the lower computational cost of running the \textit{Fast Lofo} version.

%\subsubsection{\textit{Eli5}\\}

%resumido
A very popular and quick method to be performed, the \textit{Explain Like I’m Five (Eli5)} is a complete tool that helps explore machine learning classifiers and explains the predictions \citep{eli5_ref}.%Dentre as várias maneira diferentes de explicar um modelo de aprendizagem de máquina, o Eli5 é capaz de executar o algoritmo de \textit{Mean Decrease Accuracy - MDA} com a finalidade de gerar um ranque de relevância de atributos.   
Among the many different ways to explain a machine learning model, Eli5 is capable of executing the \textit{Mean Decrease Accuracy} algorithm to generate an attribute relevance ranking \citep{eli5_git}.

%A ideia central do algoritmo que calcula o ranque de relevância do Eli5 é a seguinte: calcular a relevância da feature observando o quanto a performance (precisão, F1, R$^2$, etc.) diminui quando uma feature não está disponível para o modelo. %Assim, iterativamente cada feature é removida (apenas da parte de teste do dataset) e em seguida é calculada a performance do modelo sem que ele usasse uma feature específica. 
The central idea of the algorithm is to calculate the relevance of the feature by observing how much the performance (accuracy, F1, R$^2$, etc.) decreases when a feature is not available for the model. Thus, each feature is removed (only from the test part of the dataset) and then the model's performance is calculated without it using a specific feature \citep{eli5_ref}.

%removido
%Quando se diz remover a feature, deve-se entender que os valores desta são substituídos por ruído aleatório, não contendo assim informações úteis ao modelo. Neste método o ruído é extraído da mesma distribuição dos valores originais da feature (caso contrário, o estimador pode falhar). E assim a relevância de permutação é calculada no Eli5.
%When it is said to remove the feature, it must be understood that its values are replaced by random noise, thus not containing useful information for the model. In this method, the noise is extracted from the same distribution as the original feature values (otherwise, the estimator may fail). And so the permutation relevance is calculated in Eli5 \citep{eli5_ref}.

%\subsubsection{\textit{Shap}\\}

One of the most popular and currently used methods is the \textit{SHapley Additive exPlanations (SHAP)}, proposed as a unified method of feature relevance that explains the prediction of an instance $X$ from the contribution of an feature, based in the game theory of \textit{Shapley Value} \citep{roth1988shapley,tree_shap_ref,shap_ref}.

%Para o Shap, cada feature é considerada um jogador de um jogo e o objetivo deste jogo é alcançar a saída do modelo, ou seja, a predição em si. Desta forma, de maneira iterativa a subconjuntos de features, são re-treinados novos modelos com e sem os subconjuntos de features e assim calculados os Shapley Values de cada feature, gerando ao final um ranque de relevância de atributos.
For \textit{Shap}, each feature is considered a player in a game and the objective of this game is to achieve the model's output. In this way, in an iterative manner on feature subsets, new models are re-trained with and without the feature subsets and the \textit{Shapley Values} of each feature are calculated, ultimately generating an attribute relevance ranking \citep{shap_ref,shap_doc}.

%Esta pesquisa optou por utilizar a versão do Shap que é específico para modelos baseados em árvore de decisão, devido um menor custo computacional do que a versão agnóstica.
%This research chose to use the version of \textit{Shap} that is specific for decision tree-based models, due to a lower computational cost than the agnostic version.

%\subsubsection{\textit{Skater}\\}

Last but not least is the method \textit{Skater}, a set of tools capable of generating ranks of the relevance of model features, differing from the other methods to calculate its explanation index based on Information Theory \citep{info_theory_1994}, through measurements of entropy in changing predictions through a disturbance of a certain feature \citep{skater_ref}.

%Diferentes dos demais métodos de XAI, o Skater foi desenvolvido por uma iniciativa privada e hoje é um software de código fechado. Informações aprofundadas sobre como este método funciona são bem escaças, porém mesmo assim ele é bastante utilizado pela comunidade de XAI.
Unlike other XAI methods, \textit{Skater} was developed by a private initiative and is now closed source software. In-depth information about how this method works is very scarce, but it is still widely used by the XAI community \citep{skater_ref,skater_git}.

%O que se sabe, é que as explicações globais do modelo, o Skater faz uso de relevância de variável de maneira independente do modelo utilizado e faz uso de Gráficos de Dependência Parcial para julgar o viés de um modelo e compreender seu comportamento geral.

%removido
%What is known is that the global explanations of the model, \textit{Skater} makes use of feature relevance independently of the model used and makes use of \textit{Partial Dependency Plots (PDP)} to judge the bias of a model and understand its general behavior \citep{skater_ref,skater_citation}.

%\color{black}

\subsubsection{Explanation-by-example}

%Métodos de baseados em \textit{Explanation-by-example} selecionam instâncias específicas do conjunto de dados com o objetivo de explicar o comportamento dos modelos ou para explicar a distribuição de dados através de explicações locais \citep{molnar2020interpretable_bookref}.
Methods based on \textit{Explanation-by-example} select specific instances of the dataset in order to explain the behavior of the models or to explain the distribution of data through local explanations \citep{molnar2020interpretable}.

%Existem explicações baseadas em exemplos model-agnostic e também model-specific, sendo este último geralmente voltado para redes neurais. De acordo com \cite{molnar}, os principais tipos de explicação por exemplo são baseadas em: Counterfactual explanations \cite{wachter2017counterfactual}, Adversarial examples\cite{biggio2018wild_adversal}, Prototypes\cite{kim2016examples_critsism}, and Influential instances\cite{koh2017understanding_influence}. 
There are explanations based on model-agnostic and also model-specific examples, the latter being generally aimed at neural networks. According to \cite{molnar2020interpretable}, the main types of explanations for example are based on: \textit{Counterfactual Explanations} \citep{wachter2017counterfactual}, \textit{Adversarial Examples} \citep{biggio2018wild}, \textit{Prototypes} \citep{kim2016examples}, and \textit{Influential Instances} \citep{koh2017understanding}.

%Cada uma destas propostas, busca realizar o processo de identificação de instâncias relevantes do dataset analisado, que de maneira direta ou mesmo indireta explicam e justificam a saída do modelo seja esta correta ou incorreta\cite{molnar2020interpretable_bookref}.
Each of these proposals seeks to carry out the process of identifying relevant instances of the analyzed dataset, which directly or even indirectly explain and justify the output of the model, whether it is correct or incorrect \citep{molnar2020interpretable}.

%Com base no apresentado, esta pesquisa entende a não necessidade de comparar os resultados de explanation-by example apresentados neste artigo com resultados de ferramentas já existentes na atualidade, pois conforme visto em \cite{complexity}, é de se esperar que os resultados de explicações sejam meramente diferentes em casos específicos e ao considerar o número de comparações a serem realizadas (a nível de instâncias dos datasets) as conclusões se tornariam imprecisas. Diante disso, os resultados do estudo voltado para \textit{Explanation-by-example} são justificados diretamente pelas interpretações das propriedades (adivinhação, discriminação e dificuldade) da TRI, uma vez que o estudo proposto se baseia nela para indicar a instâncias mais significativas para a explicação do modelo analisado.
Based on what was presented above, this research understands the under-need to compare the \textit{Explanation-by-example} results presented in this article with results from tools that already exist today, because, as seen in \cite{ribeiro_complexity_et_al_2021}, it is to be expected that the results of explanations are merely different in specific cases and when considering the number of comparisons to be performed (at the level of dataset instances) the conclusions would become imprecise. 

%newnewnew
%Portanto, os resultados referentes a \textit{Explicação por exemplo} são diretamente justificados pelas interpretações das propriedades (adivinhação, discriminação e dificuldade) da \textit{IRT}, uma vez que o método proposto se baseia nesta teoria e como ela pode ajuda no processo de explicação local e global do modelo, gerando informações importantes referentes a confiança do modelo.
Therefore, the results regarding \textit{Explanation by example} are directly justified by the interpretations of the properties (guessing, discrimination and difficulty) of \textit{IRT}, since the proposed method is based on this theory and how it can help in process of local and global explanation of the model, generating important information regarding model confidence.

%\color{orange}
\subsection{Item Response Theory} \label{lab_irt}

%A área de Teoria da Resposta ao Item (IRT) pertence à Psicometria e oferece modelos matemáticos para a estimação de traços latentes, relacionando a probabilidade de um indivíduo dar uma resposta específica a um item com as características dos itens na área de conhecimento analisada \citep{PASQUALI2003_tracos_latentes,de2000teoria_irt_ref}.
The area of Item Response Theory (IRT) belongs to \textit{Psychometrics} and offers mathematical models for estimating latent traits, relating the probability of an individual giving a specific response to an item with the characteristics of the items in the area of knowledge analyzed \citep{PASQUALI2003_tracos_latentes,de2000teoria_irt_ref}.

%Métodos avaliativos tradicionais medem o desempenho de indíviduos em testes baseando-se no número total de respostas corretas, mas essa abordagem tem limitações. Como por exemplo, como lidar com respostas corretas obtidas por acerto ao acaso, ou mesmo como avaliar a dificuldade de cada pergunta do teste \cite{de2000teoria_irt_ref}.
Traditional assessment methods measure individuals' performance on tests based on the total number of correct answers, but this approach has limitations. For example, how to deal with correct answers obtained by chance, or even how to evaluate the difficulty of each test question \cite{de2000teoria_irt_ref}.

%Ao contrário de avaliações tradicionais, a IRT concentra-se nos itens do teste, avaliando o desempenho com base na capacidade de se acertar itens específicos, não apenas na contagem total de respostas corretas \cite{hambleton1991fundamentals_irt_2}.
Unlike traditional assessments, IRT focuses on test items, evaluating performance based on the ability to get specific items correct, not just the total count of correct answers \cite{hambleton1991fundamentals_irt_2}.

%A IRT busca avaliar características latentes não observáveis de um indivíduo, fornecendo a relação entre a probabilidade de resposta correta e seus traços latentes, ou seja, a habilidade do indivíduo na área de conhecimento avaliada \cite{hambleton1991fundamentals_irt_2}.
IRT seeks to evaluate unobservable latent characteristics of an individual, providing the relationship between the probability of a correct answer and their latent traits, that is, the individual's ability in the area of knowledge evaluated \cite{hambleton1991fundamentals_irt_2}.

%Em resumo, a IRT consiste em modelos matemáticos que representam a probabilidade de um indivíduo acertar um item, considerando parâmetros do item e a habilidade do respondente. Diferentes implementações da IRT existem na literatura, como o "\textit{Modelo Dicotômico de Rasch}" \citep{kreiner2012rasch_params_item} e o "\textit{Modelo Tridimensional de Birnbaum}" \citep{birnbaum1968some_parameters_irt}. Neste estudo, foi utilizada a última implementação, conhecida como modelo logístico $3PL$, que realiza cálculos probabilísticos em duas etapas: \textit{Estimação de Parâmetros do Item} e \textit{Estimação de Habilidade}.
In summary, IRT consists of mathematical models that represent the probability of an individual getting an item correct, considering item parameters and the respondent's ability. Different implementations of IRT exist in the literature, such as the ``\textit{Rasch Dichotomous Model}'' \citep{kreiner2012rasch_params_item} and the ``\textit{Birnbaum Three-Dimensional Model}'' \citep{birnbaum1968some_parameters_irt}. In this study, the latest implementation was used, known as the $3PL$ logistic model, which performs probabilistic calculations in two stages: \textit{Item Parameter Estimation} and \textit{Skill Estimation}.

%Detalhe, entender os processos realizados na Teoria de Resposta ao Item não é uma tarefa trivial para um pesquisador de computação voltado para machine learning, fato natural devido ser um conhecimento da área de Psicometria \cite{}. Com isso, nos dois tópicos a seguir serão descritos como os principais processos desta teoria podem ser entendido e abstraído para a área de machine learning através de um exemplo hipotético.
In the two topics below, it will be described how the main processes of this theory can be understood and abstracted to the area of machine learning through a hypothetical example.

\subsubsection{Estimation of Item Parameters}

%O cálculo dos parâmetros de item no modelo logístico de três parâmetros (3PL) é realizado por meio de técnicas de estimação, como o método da Maximum Likelihood Estimation - MLE. Neste cálculo o objetivo é encontrar os valores dos parâmetros que maximizam a probabilidade de observar as respostas reais dos indivíduos aos itens.
The calculation of item parameters (discrimination, difficulty, and guessing) in the \textit{3PL} is performed using estimation techniques, such as the \textit{Maximum Likelihood Estimation - MLE} \citep{myung2003tutorial_mle}. In this technique, the objective is to find the values of the parameters that maximize the probability of observing the actual responses of individuals to the items. Each property has the following definition:

%A MLE é responsável por estimar os valores de discriminação, dificuldade e adivinhação para cada item de um teste. Sendo estes valores utilizados no cálculo da habilidade dos respondentes que será mostrado mais a frente.
%removido
%The \textit{MLE} is responsible for estimating discrimination, difficulty, and guessing values for each item on a test. These values are used to calculate the ability of the respondents, which will be shown later. 

\begin{itemize}
    \item \textit{Discrimination:} consists in how much a specific item $i$ is able to differentiate between highly and poorly skilled respondents. It is understood that the higher its value, the more discriminative the item is. Ideally, a test should feature a gradual and positive discrimination;
    
    \item \textit{Difficulty:} represents how much a specific item $i$ is hard to be responded correctly by respondents. Higher difficulty values represent more difficult items to answer;
  
    \item \textit{Guessing:} representing the probability that a respondent gets a specific item $i$ right randomly. It can also be understood as the probability that a respondent with low ability will get the item right. It is also the smallest possible chance that an item will be correct regardless of the estimated ability of the respondent.
    
\end{itemize}

%Para estimar os parâmetros de item, é necessário ter um conjunto de dados referentes as respostas dos indivíduos para cada item, como indicado em \ref{table_irt}
To facilitate the abstraction of the procedure performed, we will use a simple example as a small case study containing 6 individuals, 4 hypothetical questions and the answers of each one arranged in a response matrix presented in the table \ref{table_irt_mr}.

%Este exemplo hipotético é somente para facilitar a abstração dos processos realizados, desta forma todos os valores calculados a partir deles são ilustrativos. Isso se faz necessario devido os resultados da aplicação daIRTsão mais confiáveis ao se ter entre 200 a 300 respondentes.

%resumido
Note, in the hypothetical example the values presented are illustrative, because the results of the \textit{IRT} are accurate when you have more then 100 individuals \citep{de2000teoria_irt_ref,prudencio2015analysis_irt_ml_1,baylari2009design_irt_esa}.

\begin{table}[!h]
\begin{center}
\caption{Simple example of a response matrix of 6 individual and 4 questions.}
\resizebox{.7\textwidth}{!}{%
\begin{tabular}{c|c|c|c|c}
\hline
             & Question 1 & Question 2 & Question 3 & Question 4 \\ \hline
Individual 1 & 1          & 1          & 1          & 1          \\ %\hline
Individual 2 & 1          & 1          & 1          & 0          \\ %\hline
Individual 3 & 1          & 1          & 0          & 0          \\ %\hline
Individual 4 & 1          & 1          & 0          & 0          \\ %\hline
Individual 5 & 1          & 0          & 0          & 0          \\ %\hline
Individual 6 & 1          & 0          & 0          & 1          \\ \hline
\end{tabular}
}
\label{table_irt_mr}
\end{center}
\end{table}

In the table \ref{table_irt_mr} the number ``1'' indicates a correct answer and ``0'' indicates a wrong answer by the respondent (individual). From this response matrix, the parameter estimation can be done in 4 steps: 

\textit{Parameter initialization:} Item parameters are initialized with arbitrary values.

\textit{Calculation of expected probabilities}: From the initial item parameters, the expected correct answer probabilities for each item are calculated for each individual.

\textit{Parameters Update:} Item parameters are updated iteratively, using the \textit{MLE} to maximize the likelihood of the data, by comparing the observed responses of individuals with the expected probabilities calculated in the previous step.

\textit{Iterations:} The previous two steps are repeated until the parameter estimates do not change significantly between iterations.

In the table \ref{table_irt_ip}, follow the hypothetically extracted item parameter values for the responses of individuals in the table \ref{table_irt_mr}.

\begin{table}[!h]
\begin{center}
\caption{Hypothetical values of item parameters estimated by Table \ref{table_irt_mr}.}
\resizebox{.6\textwidth}{!}{%
\begin{tabular}{c|c|c|c}
\hline
           & Discrimination & Difficulty & Guessing \\ \hline
Question 1 & 0.00           & $-\infty$  & 0.00     \\ %\hline
Question 2 & 0.19         & 0.36       & 0.00     \\ %\hline
Question 3 & 0.58         & 0.39      & 0.00     \\ %\hline
Question 4 & 0.86         & 1.28      & 0.25     \\ \hline
\end{tabular}
}
\label{table_irt_ip}
\end{center}
\end{table}

%O exemplo hipotetico descrito na tabela \ref{table_irt_mr}, mostra que uma teste com 4 questões foi aplicado para um total de 6 individuos, através de uma simples contagem de respostas corretas, pode-se afirmar que o individuo 1 acertou mais questões (total de 4). Também pode-se afirmar que o candidato 5 foi o que acertou menas questões (total de 1), já os demais candidatos tiveram resultados intermediários.
The hypothetical example described in the Table \ref{table_irt_mr}, shows that a test with 4 questions was applied to a total of 6 individuals, through a simple counting of correct answers, it can be stated that individual 1 got more questions right (total from 4). It can also be said that candidate 5 was the one who got the fewest questions right (total of 1), while the others candidates had intermediate results.

%Ao observar as estimativas de parâmetros de item apresentadas na tabela \ref{table_irt_ip}, verifica-se que a discriminação, a dificuldade e a adivinhação fornecem informações mais específicas sobre cada questão do teste, como:
When observing the hypothetical item parameter estimates presented in the Table \ref{table_irt_ip}, it is verified that discrimination, difficulty and guessing provide more specific information about each test question, such as below.

\textit{Discrimination:} ``Question 1'' cannot be considered the best to discriminate individuals because the value 0 (discrimination null), this happened because all the respondents got the question right. ``Question 4'' can be considered the most discriminating question, followed by ``Question 3'' and ``Question 2'' respectively.
    
\textit{Difficulty:} higher difficulty values indicate more difficult questions, so ``Question 4'' is considered the most difficult, followed by 2, 3 (with very close values)  and 1 (with value $-\infty$ due to all individuals having got it right).

\textit{Guessing:} the question that presented significant value was ``Question 4'', as it is observed that ``Individual 6'' even without correctly answering ``Question 2'' and ``Question 3'' (which present intermediate difficulties similar) correctly answered question 4 (considered the most difficult).

%Conforme pode-se observar acima, o processo de estimação dos parâmetros de item fornecem por sí só informações importantes sobre questões e individuos relacionados a um teste avaliativo. No passo seguinte, serão apresentados procedimentos relacionados a estimação da habilidade latente de cada indivíduo.
As noted above, the process of estimating item parameters provides important information about questions and individuals related to an evaluative test. Below are procedures for calculating the ability of individuals.

\subsubsection{Estimation of ability}

The logistic model $3PL$, presented in the equation \ref{eq:ml3}, consists of a model capable of evaluating the respondents of a test from the estimated ability ($\theta_{j}$), together with the correct answer probability $P(U_{ij} = 1\mid\theta_{j})$ calculated as a function of the individual skill $j$ and the parameters of the item $i$.

\begin{equation}
    \label{eq:ml3}
    P(U_{ij} = 1\mid\theta_{j}) = c_{i} + (1 - c_{i})\frac{1}{1+ e^{-a_{i}(\theta_{j}-b_{i})}}
\end{equation}

%O modelo logístico de três parâmetros 3PL é usado para modelar a relação entre a habilidade dos indivíduos e a probabilidade de responder corretamente a um item em um teste ou questionário. Ele assume que a probabilidade de resposta correta depende de três parâmetros de item: habilidade do indivíduo, dificuldade do item e discriminação do item. Seeking to better explain what each of the item parameters shown in the equation \ref{eq:ml3} is, the following definitions are given:
The 3PL is used to model the relationship between individuals' ability and the likelihood of correctly answering an item on a test. It assumes that the probability of a correct answer depends on three item parameters: item discrimination, item difficulty, and item guessing.

%Na equação \ref{eq:ml3} as propriedades discriminação, dificuldade e adivinhação dos itens $i$, são representadas respectivamente pelas letras $a$,$b$, e $c$
In the equation \ref{eq:ml3} the properties discrimination, difficulty and guessing of the items $i$, are represented respectively by the letters $a_i$,$b_i$, and $c_i$

%\begin{equation}
%   \label{eq:calc_item_param}
%    \pi_i = c_i + (1 - c_i) \frac{exp[a_{i}(\theta - b^*_{i})]}{1+exp[a_{i}(\theta - b^*_{i})]}
%\end{equation}

%Onde $\pi_i$ denota a probabilidade condicional de responder corretamente ao $i$-ésimo item dado $z$, $c_i$ é o parâmetro adivinhação, $\beta_{1i}$ é o parâmetro dificuldade, $\beta_{2i}$ é o parâmetro discriminação, e $\theta$ denota a habilidade latente.

%O ajuste do modelo, representado pela equação \ref{eq:calc_item_param}, é baseado na máxima verossimilhança marginal aproximada, usando a regra de quadratura de Gauss Hermite \citep{liu1994note_gaussiana_quad} para a aproximação das integrais requeridas.

The $\theta_j$ is the ability of the individual $j$, which is a continuous parameter representing the latent trait being measured. $P(U_{ij} = 1\mid\theta_{j})$ represents the probability of correctly answering item i for an individual $j$ with ability $\theta$;

%Agora com o conhecimento da equação $PL3$, retorna-se ao exemplo hipotético mostrando-se o valor de \theta de cada indivíduo. 
Now with the knowledge of the $PL3$ equation, we return to the hypothetical example showing the value of $\theta$ for each individual, table \ref{table_irt_theta}.

\begin{table}[!h]
\caption{Result of $\theta$ estimation using the hypothetical example as a basis.}
\begin{center}
\resizebox{.3\textwidth}{!}{%
\begin{tabular}{c|c}
\hline
Respondent & Ability ($\theta$) \\ \hline
Individual 1        & 0.905          \\ %\hline
Individual 2        & 0.281         \\ %\hline
Individual 3        & 0.195         \\ %\hline
Individual 4        & 0.195         \\ %\hline
Individual 5        & 0.091         \\ %\hline
Individual 6        & 0.091         \\ \hline
\end{tabular}
}
\end{center}
\label{table_irt_theta}
\end{table}

%O valor calculado no processo de estimação de abilidade \theta, apresentado na tabela \ref{table_irt_theta} é adaptado por \cite{lucas} e indica de os individuos com maiores habilidade (maiores valores) e os indivíduos com menores habilidades (menores valores). Perceba que o ``Individuo 1'' é o mais habilidoso dentre os respondentes, seguido respectivamente pelo ``Individual 2'', ``Individual 3'', ``Individual 4'' and ``Individual 5'' (estes três últimos empatados com score de theta igual a -1.095), ficando em ultima posição o ``Individuo 6'' que foi penalizado por acertar uma das questões por adivinhação.
The value calculated in the $\theta$ ability estimation process, presented in the table \ref{table_irt_theta}, indicates individuals with greater ability (higher values) and individuals with lower abilities (lower values). Note that ``Individual 1'' is the most skilled among the respondents, followed respectively by ``Individual 2'', ``Individual 3'', ``Individual 4'', ``Individual 5'' and ``Individual 6'', these last two tied, as the question agreed by guessing (``Question 4'' by ``Individual 6'') was disregarded.

\begin{figure*}[!h]
%%\color{teal}
\begin{center}
\includegraphics[scale=0.35]{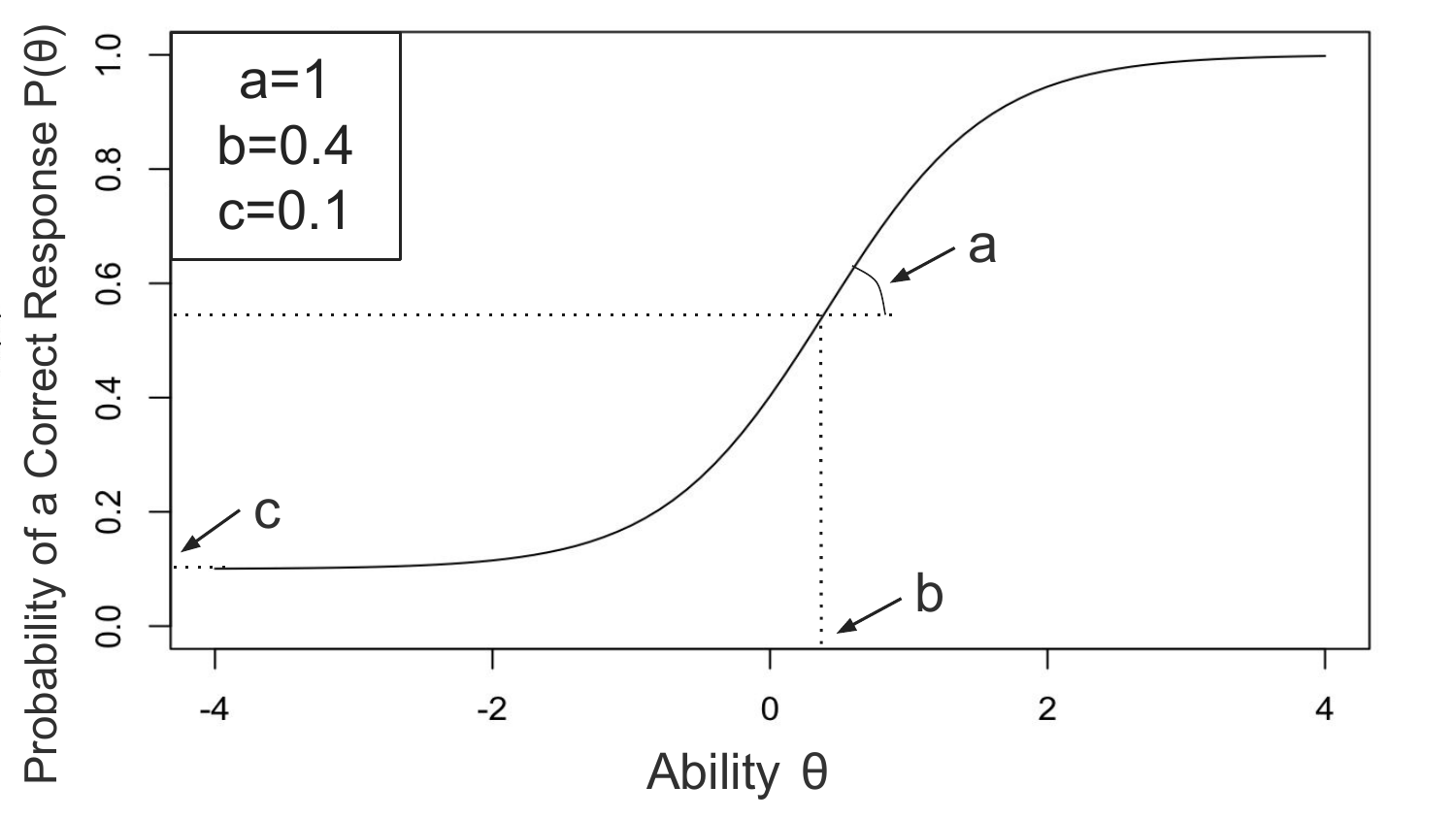}
\caption{Example of the representation of the parameter values of an item arranged on the Item Characteristic Curve - ICC. The letters $a$, $b$ and $c$ represent the discrimination, difficulty and guessing properties, respectively.}
\label{fig_cci}
\end{center}
\end{figure*}

Thus, once the item parameters are estimated and the hit probability is calculated using the equation \ref{eq:ml3}, the \textit{Item Characteristic Curve (ICC)} can be obtained. The \textit{ICC} defines the behavior of an item's hit probability curve according to the parameters describing the item ($a_i$, $b_i$ e $c_i$) and the respondents' skill variance, figure \ref{fig_cci}.

As can be seen in figure \ref{fig_cci}, the hit probability on axis $y$ is calculated by adding the values of the properties $a_i$, $b_i$ e $c_i$ found in an item and the variation of the skill $\theta$.

Thus, the property $a_i$ (discrimination) is responsible for the slope of the logistic curve; the property $b_i$ (difficulty) plots the curve as a function of skill in the logistic function; and the property $c_i$ (guessing) places the basis of the logistic function relative to the axis $y$.

%removido
%The values found in the discrimination, difficulty and guessing properties are found for each instance and enable \textit{IRT} to measure the ability $\theta$ of each respondent in the evaluation process, resulting in a rank of respondents arranged according to their ability, like Table \ref{table_irt_theta}. 

\begin{figure*}[h]
\begin{center}
\includegraphics[scale=0.58]{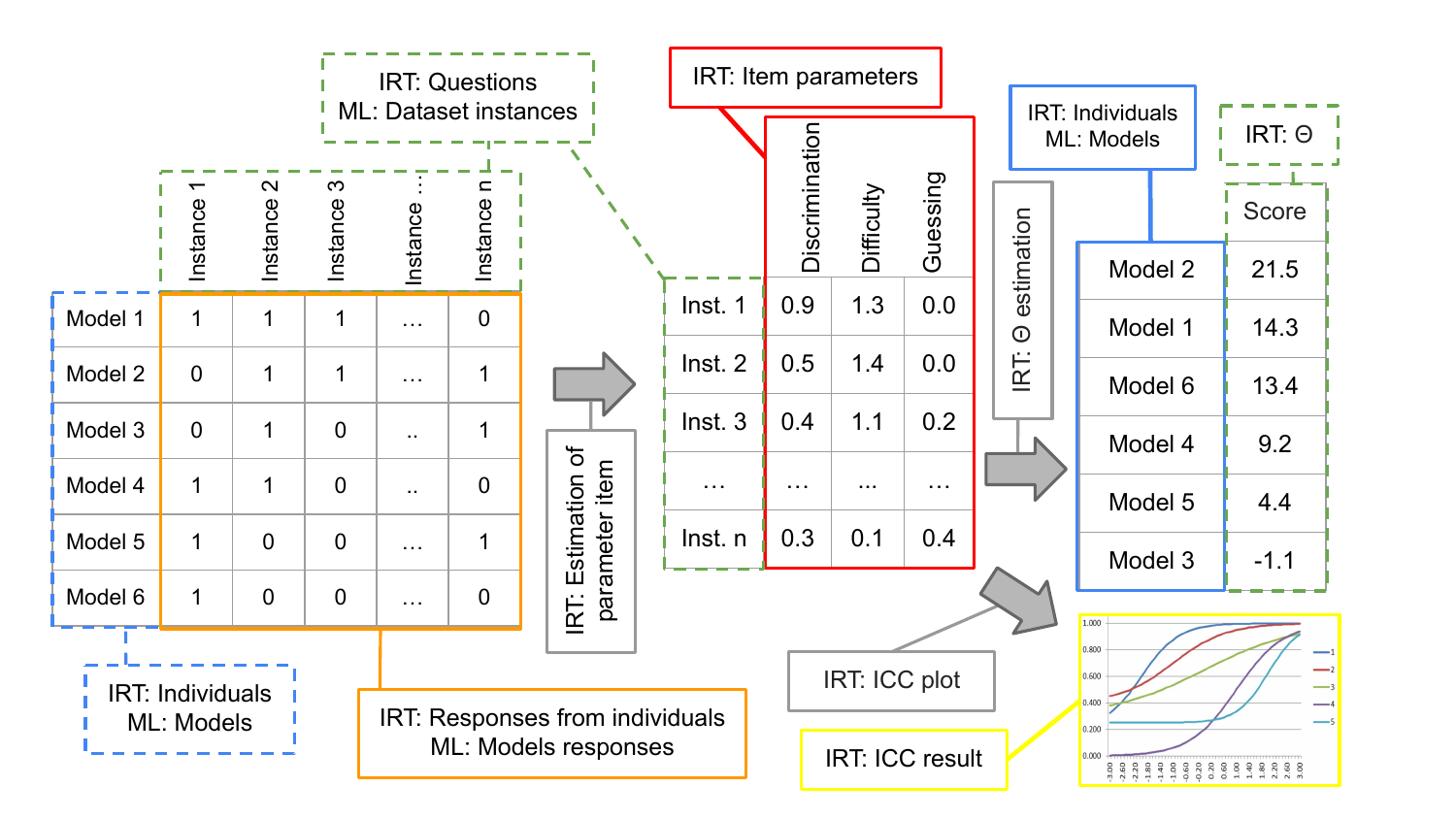}
\caption{Associations of \textit{IRT} terms and Machine Learning terms.}
\label{fig_irt_ml}
\end{center}
\end{figure*}

As stated herein, the \textit{IRT} emerges as a consolidated theory from the field of \textit{Psychometrics}, which can be adapted to the field of Machine Learning (ML). Therefore, it is sufficient to consider that a ``test'' is a dataset, each ``test item'' is the instances of this dataset, ``independent variables'' are the questions and the ``dependent variable'' is the answer, ``each answer'' can be evaluated as right or wrong, and each ``individual'' is a separate machine learning model. This association can be better understood from figure \ref{fig_irt_ml}.

That is, from this abstraction above, this important theory can be used to evaluate computational models, thus obtaining the evaluative benefits that \textit{IRT} provides. Similar abstractions to this one have been made in machine learning research that uses this theory \citep{martinez2016making_irt_ml_2,martinez2019item_implement_irt,kline2021item_irt_ml_4_raissa,baylari2009design_irt_esa,chen2006personalized_irt_esa,araujo2023quest_vitor}.

%removido
%\textit{Item Response Theory} currently has many applications in computing and its use goes beyond the abstractions presented above, such as  \citep{cardoso2022explanationbyexample_irt,chen2006personalized_irt_esa,cardoso2020decoding_irt,martinez2019item_implement_irt,bergner2012model_implement_irt,baylari2009design_irt_esa,araujo2023quest_vitor}. It is worth mentioning that up to the time this article was written, the literature survey had not identified any research related to the use of \textit{IRT} as an XAI method to explain black box models based on feature relevance ranking.

%\color{black}
%\color{purple}
\section{Materials and Methods}\label{label_materiais_metodos}

%Nesta seção, serão apresentados aspectos metodológicos da pesquisa, relacionados a: Descrição do pipeline desenvolvido (subseção \ref{lab_pipeline}); Definição dos datasets utilizados e seus préprocessamentos (subseção \ref{lab_preprocess}); Descrição da clusterização (subseção \ref{lab_clustering}); Descrição da \textit{MCA} (subseção \ref{lab_mca}); Definição da medida de correlação (subseção \ref{lab_correlation}); E por último, a definição do método de XAI \textit{eXirt} (subseção \ref{lab_exirt}).
In this section, methodological aspects of the research will be presented, related to: Description of the developed pipeline (subsection \ref{lab_pipeline}); Definition of the datasets used and their preprocessing (subsection \ref{lab_preprocess}); Description of clustering (subsection \ref{lab_clustering}); Description of \textit{MCA} (subsection \ref{lab_mca}); Definition of the correlation measure (subsection \ref{lab_correlation}); And finally, the definition of the XAI method \textit{eXirt} (subsection \ref{lab_exirt}).

%%\color{orange}
\subsection{Pipeline overview} \label{lab_pipeline}
%Uma visão macro dos principais pontos metodologicos abordado neste estudo, pode ser vista a partir da figura \ref{fig_metodologia}, que é apresentada a fim de ser facilitar a abstração dos passos realizados.

\begin{figure*}[!h]
\begin{center}
\includegraphics[scale=0.54]{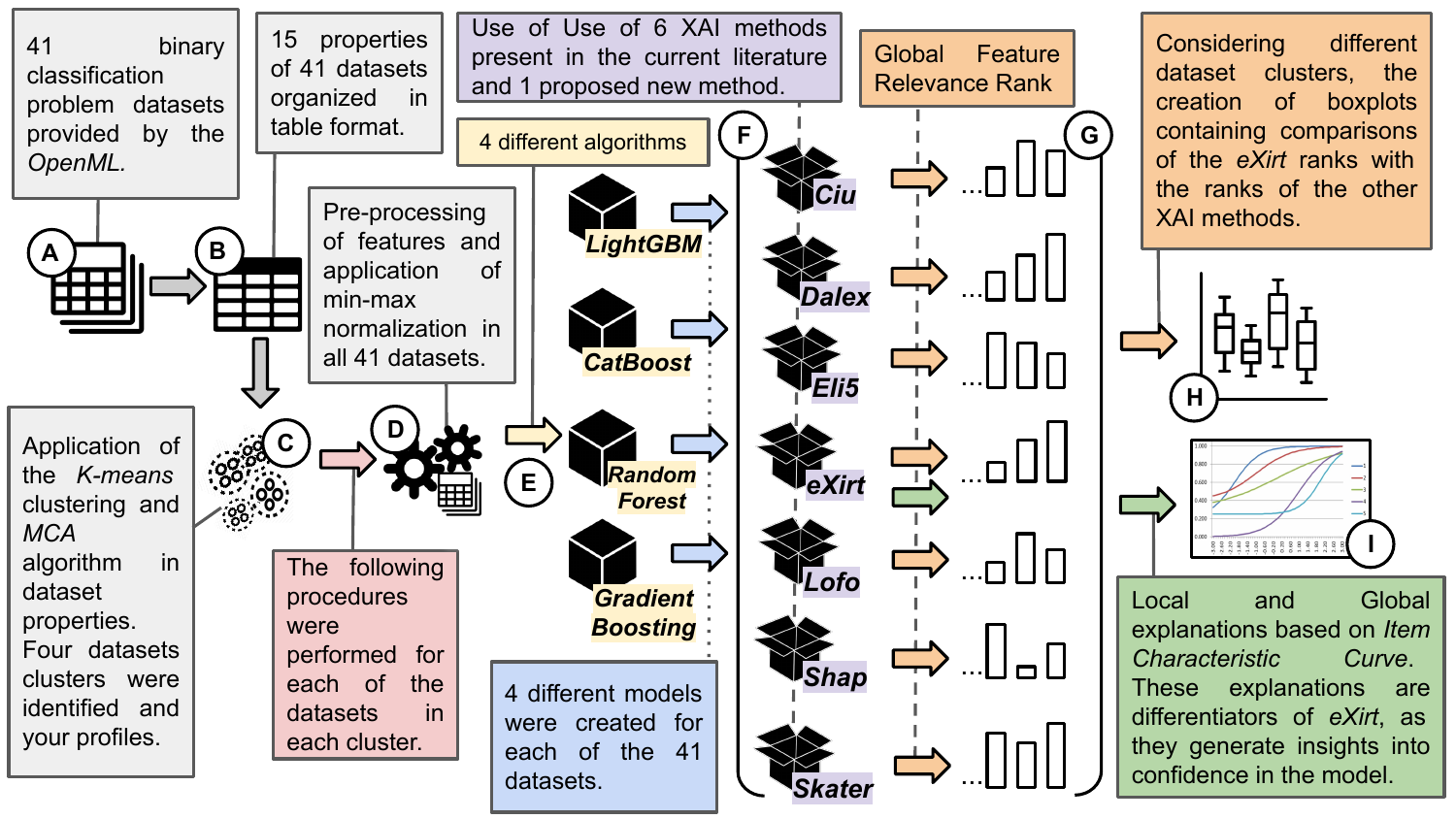}
\caption{Visual scheme of all steps and processes performed by the pipeline.}
\label{fig_metodologia}
\end{center}
\end{figure*}

A general view of the main methodological points addressed in this study (in pipeline format) can be seen from the figure \ref{fig_metodologia}, which is presented in order to facilitate the abstraction of the steps performed in a methodology.

The pipeline, figure \ref{fig_metodologia}, presents the steps: (A) Selection of 41 datasets referring to binary classification problems present in the OpenML platform \citep{openml}; (B) Utilization of 15 property parameters from all 41 analyzed datasets (provided by OpenML OpenML platform); (C) Application of the \textit{K-means} \citep{kmeans} clustering algorithm in the dataset properties table, followed by a Multiple Correspondence Analysis \citep{ref_mca}; (D) Pre-processing of the 41 datasets selected in the initial step; (E) Use of 4 tree-ensemble algorithms in each of the 41 datasets to create black-box models;  (F-G) Use of 6 XAI methods present in the literature and \textit{eXirt} (method proposed in this article) to create global explanations based on feature relevance ranks; (H) Comparison of the feature relevance ranks found by \textit{eXirt} with the ranks of the other XAI methods (considering the existing dataset clusters); (I) Use the item parameter estimates created by \textit{eXirt} in step (F) and create local and global explanations of the analyzed models.

%Detalhe, todos os processos realizados no pipeline podem ser reproduzidos a partir de seu repositório descrito nas Informações Suplementares (D).
Detail, all processes carried out in the pipeline can be reproduced from its repository described in \textit{Supplementary Information} (D).

%%\color{black}

\subsection{Dataset's Preprocess} \label{lab_preprocess}

Were used 41 datasets, selected from the \textit{OpenML} \citep{openml} basis, while observing the fact that they refer to binary classification problems, without data loss and with a greater number of executions by the community, in order to contribute to research that already uses these datasets present in this repository, which is considered the gold standard of data quality \citep{cardoso2020decoding_irt}. 

The datasets used were as follows: \textit{australian, analcatdata-lawsuit, phishing websites, spec, satellite, banknote authentication, blood transfusion service center, churn, climate model simulation crashes, credit-g, delta ailerons, diabetes, eeg-eye-state, haberman, heart-statlog, ilpd, ionosphere, jEdit-4.0-4.2, kc1, kc2, kc3, kr-vs-kp, mc1, monks-problems-1, monks-problems-2, monks-problems-3, mw1, mozilla4, ozone-level-8hr, pc1, pc2, pc3, pc4, phoneme, prnn crabs, qsar-biodeg, sonar, spambase, steel-plates-fault, tic-tac-toe} and \textit{wdbc}.

In all the datasets mentioned above, the following pre-processing and/or coding were applied: conversion of independent variables from categorical to numeric, application of min-max normalization (using 0 as min and 1 as max by feature) of independent variables and coding of the target variable (dependent) in binary to ``0'' or ``1''.

\subsection{Dataset's Clustering} \label{lab_clustering} %Clusters

%Analisar as caracteristicas de um dataset sem levar em conta o contexto (problema do mundo real que o dataset representa) não é uma tarefa nada fácil, pois está envolto a diversas propriedades técnicsa do dado. Até por que um dataset é constituido por diversas propriedades que refletem como ele foi construido \citep{machine_learning}.
Analyzing the characteristics of a dataset without taking into account the context (real world problem represented by the dataset) is not a trivial task, as it involves several technical properties of the data. Even because a dataset is made up of several properties that reflect how it was built \citep{dataset_characteristics_effects}.

%Esta pesquisa optou por evitar uma análise individual e meramente descritiva de cada um dos 41 datasets selecionados anteriormente, a fim de tornar os resultados mais genéricos e passíveis de serem transcritos no limite de espaço de escrita deste artigo.%Seguindo esta ideia, optou-se por agrupar tais datasets através do processo de clusterização utilizando as propriedades deles, buscando ao final do processo a identificação do perfil de cada cluster de dataset por meio de análises de correlações existentes.
This research chose to avoid an individual and merely descriptive analysis of each of the 41 previously selected datasets, in order to make the results generic. Following this idea, it was decided to group such datasets through the clustering process using their properties, seeking at the end of the process to identify the profile of each dataset cluster through analysis of existing correlations.

%Os dados relacionados as propriedades dos datasets, foram fornecidas diretamente pela API do OpenML, que por questões de consistência limitou-se a 15 propriedades. %As 15 propriedades dos 41 datases foram consolidadas em um dataset intermediário chamado de `propriedades dos datasets'.
The data related to the properties of the datasets were provided directly by the OpenML API, which for consistency reasons was limited to 15 properties. This properties, were consolidated into an intermediate dataset called ``dataset properties''. On this dataset, the \textit{K-means} \citep{kmeans} algorithm was executed in order to identify how many clusters the data could be better grouped. In \textit{Supplementary information (B)}, it is possible to observe the visual of mentioned dataset.

\begin{figure}[!h]
\begin{center}
\includegraphics[scale=0.5]{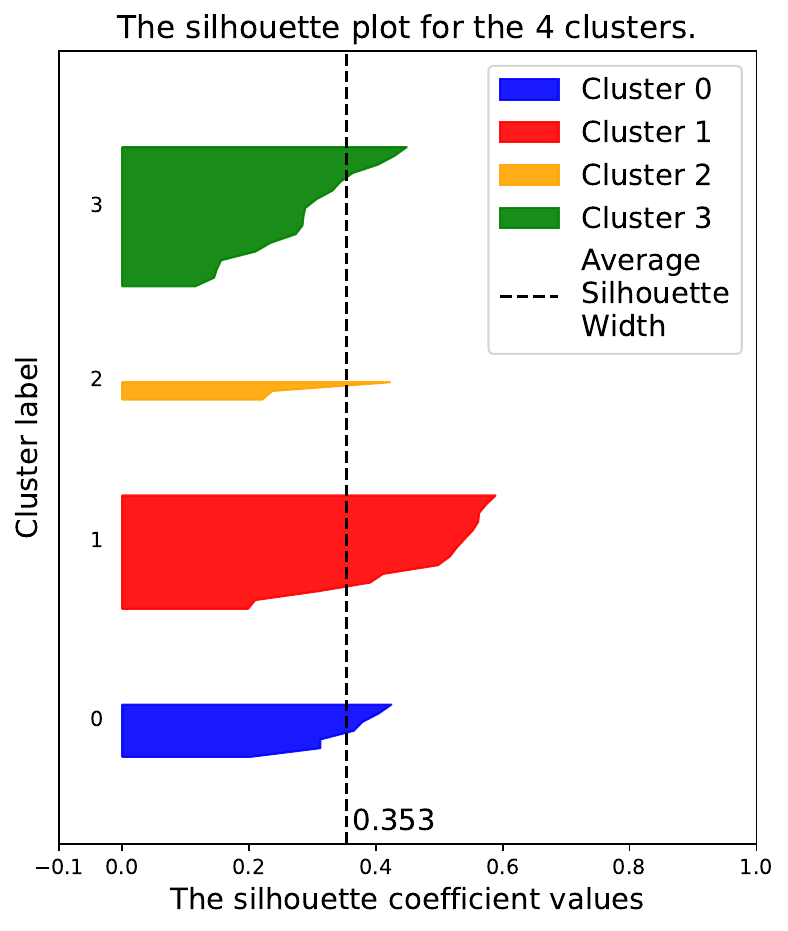}
\caption{Silhouette coefficients for clustering, using the \textit{K-means} algorithm, for $K = 4$. Distance means (axis $x$) and label of clusters $0, 1, 2$ and $3$ (axis $y$).}
\label{fig_silhuette}
\end{center}
\end{figure}

Seeking to identify the optimal number of clusters to best separate the 41 datasets analyzed, the algorithm for interpretation and validation between data clusters was used, which is called Silhouettes \citep{silhouettes}, by varying the value $K$ (clusters) between  2 and 10. In the end, $K = 4$ was found with \textit{Average Silhouette Width} $= 0.353$, figure \ref{fig_silhuette}.

Figure \ref{fig_silhuette} shows the result of running the silluette algorithm, where it can be seen that for $k=4$ the distances between each cluster (0, 1, 2 and 3) are above the average (red line), so this is an appropriate value of $k$.

\subsection{Dataset's Multiple Correspondence Analysis} \label{lab_mca}

%No passo apresentado seção acima, clusterização, encontrou-se 4 grupos de datasets que apresentaram características similares entre sí e por isso foram identificados como pertencentes a um mesmo cluster após a execução do algoritmo \textit{K-means}. Porém, fez necessária uma análise mais profunda, a fim de se conseguir definir quais perfis os datasets existentes em cada cluster tinham. Com isso, surgiu a necessidade de se utilizar a Multiple Correspondence Análises - MCA \cite{mca}.
In the step presented in the section above, 4 groups of datasets were found that presented similar characteristics to each other and, therefore, were identified as belonging to the same cluster after the execution of the \textit{K-means} algorithm. However, a deeper analysis was necessary in order to be able to define which profiles the existing datasets in each cluster had. With that, the need arose to use the \textit{Multiple Correspondence Analysis (MCA)} \citep{ref_mca}.

In the \textit{MCA} \citep{ref_mca} analysis applied on the properties datasets, whereby the lines in this table are the observations or individuals $(n)$ concerned --- the datasets are here --- and the columns are the different categories of nominal variables $(p)$ --- here are the properties of each dataset. In this analysis, the label of the cluster each dataset belongs to was taken into consideration.

%Para que a análise MCA pudesse ser executada, ouve a necessidade de se aplicar o processo de binarização ao dataset contendo as 15 propriedades analisadas. Para isso, recorreu-se ao processo de binarização em diferente bins que é relativamente simples, a fim de minimizar qualquer interferência sobre os achados ao fim da análise. 
For the \textit{MCA} analysis to be performed, there is a need to apply the binarization process to the dataset containing the 15 analyzed properties. For this, the binarization called ``cut'' \citep{cut_ref}, this function is also useful for going from a continuous variable to a categorical variable. Supports binning into an equal number of bins, or a pre-specified array of bins.

%Essa função oferece suporte ao agrupamento em um número igual de compartimentos ou em uma matriz de compartimentos pré-especificada. Aqui usamos o número igual de compartimentos, calculado a partir do número de compartimentos de igual largura no intervalo da variável $x$.
This function, ``cut'', supports binning into an equal number of bins, or a pre-specified array of bins. Here we used the equal number of bins, calculated from the number of equal-width bins in the range of $x$ variable.

%Durante o processo de binarização, os valores apresentados em bins menores receberam a codificação `s', já os valores existentes em bins maiores receberam a codificação `h'.
During the binarization process, the values presented in smaller bins were coded ``s'', while the values in higher bins were coded ``h''. The result of the binarization, on the dataset with the 15 properties, can be seen in the section \textit{Supplementary Information} (C). 

Note, this research initially considered working with \textit{Principal Component Analysis - PCA} \citep{pca_ref} instead of \textit{MCA}, but only this last technique allowed the identification of the dataset profile of each cluster, based on two ranges of bin values (low value ``s'' and high value ``h'') for the 15 analyzed properties. Therefore, this is the technique responsible for the basis of the results that will be presented in section \ref{data_profile}.

\subsection{Rank Correlations} \label{lab_correlation}

For all 41 datasets (clustered into 4 different clusters), 4 machine learning models were created, based on the \textit{Light Gradient Boosting}, \textit{CatBoosting}, \textit{Random Forest}, and \textit{Gradient Boosting} algorithms. In all, a total of 164 models and 1.148 ranks were generated. 

In order to calculate the correlation between \textit{eXirt} ranks and ranks from others XAI methods (for each of the 4 models), the \textit{Spearman's Rank Correlation Coefficient} \citep{spearman_ref} was selected. The reason for using this particular algorithm is that it measures the correlation between pairs of ranks considering the idea of ranks (positions) where different values (in this case, dataset features) may appear. 

\subsection{The \textit{eXirt} Method} \label{lab_exirt}

%E capaz de realizar explicações de maneira global (usando ranque de relevância de atributos) e local (através do explicações baseadas em parâmetros de item e na Curva característica do Item ) ambas baseadas em propriedes da \textit{IRT} (discriminação, dificuldade e adivinhação).
The method Explainable Based on \textit{Item Response Theory} - \textit{eXirt} is one of the XAI methods performed in the developed pipeline. This method is a new proposal to generate explanations for tree-ensemble models. It is able to perform explanations based on feature relevance rank (global) and based on item parameters and the Item Characteristic Curve (local and global), both based on \textit{IRT} properties (discrimination, difficulty and guessing).

\begin{figure*}[!h]
\begin{center}
\includegraphics[scale=0.55]{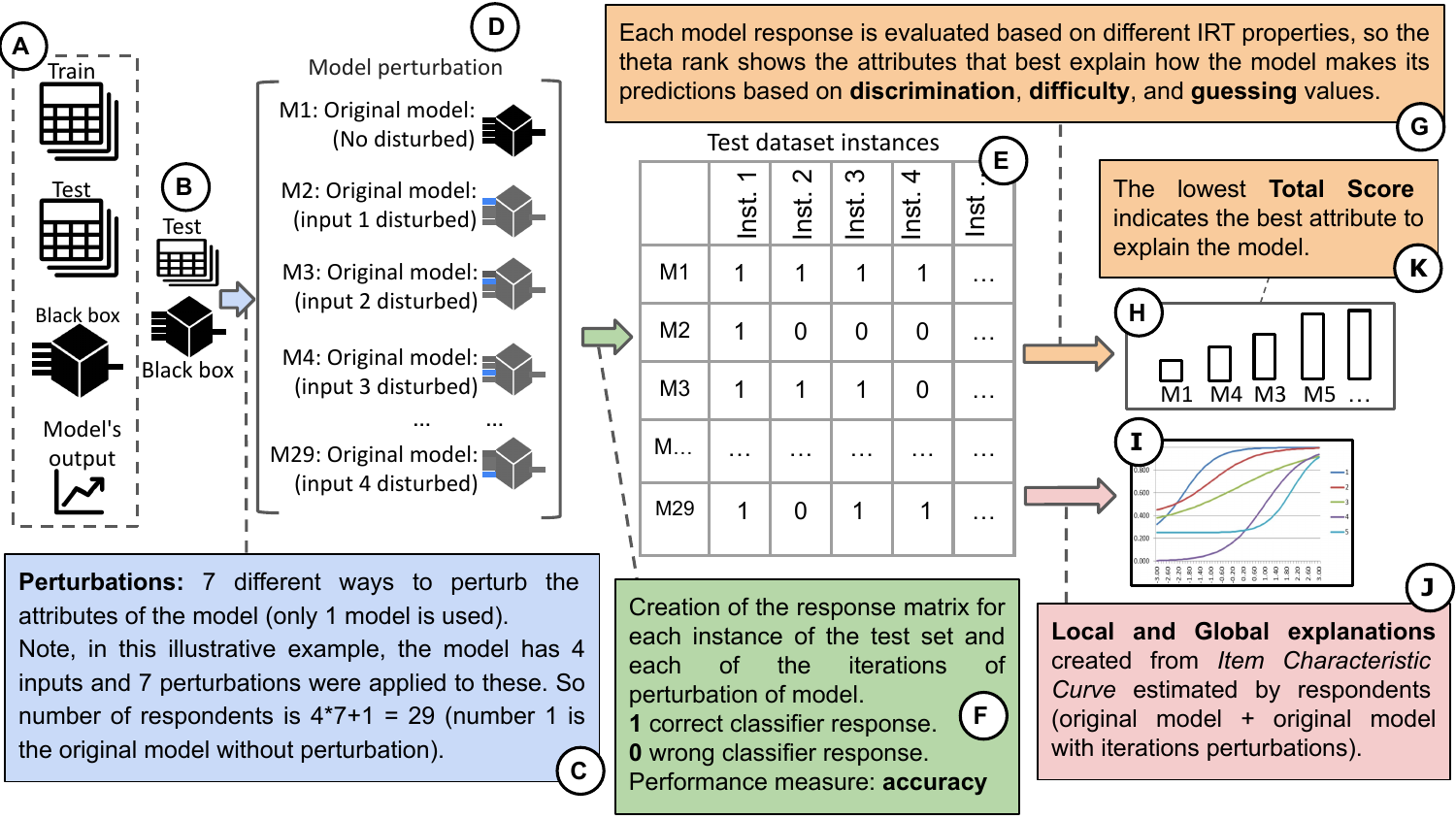}
\caption{Visual summary of all steps and processes performed by the XAI measure, called \textit{eXirt}.}
\label{fig_metodologia_eXirt}
\end{center}
\end{figure*}

Just like other XAI methods, \textit{eXirt} uses the training data, test data, the model itself together with its outputs, figure \ref{fig_metodologia_eXirt} (A). In (B) the test data and a model are passed on to the \textit{eXirt}. In (C-D) the method creates the so-called ``Model Perturbation'' (responsible for simulating the response of several respondents using only a single model). In (E-F) the creation of the response matrix used in the \textit{IRT} run.

At the end of the process, about figure \ref{fig_metodologia_eXirt}, in (G-K) a global explanations are generated based on relevance feature rank (calculated from the theta value). In (I-J) the local explanations are also created from the item parameters and also the curve characteristic of the item.

%resumido
The ``Model Perturbation'' process, figure \ref{fig_metodologia_eXirt} (D) is inspired on methods \textit{Dalex} \citep{dalex_book} and Lofo \citep{lofo_ref}, for in this process iterative perturbations are performed on all features of the model (iteratively, one feature at a time), and at each iteration the answers of the model prediction are collected and, thus, it is possible to have a high number of responding candidates (since a high number of respondents is a necessity for the\textit{IRT}). %It should be noted that the responses of each iteration may be different or the same as the existing ones, but it is expected that perturbations of the model inputs generate greater errors (calculated based on the predictions of the original model) in its outputs in proportion to the relevance of the feature.

Still referring to the ``Model Perturbation'', 7 different ways to perturb input features of model have been used, figure  \ref{fig_metodologia_eXirt} (C). Each of these perturbations are applied to each input of the model (individually). These perturbations can be divided into three types: \textit{Feature Perturbation, Single Value Perturbation} and \textit{Permutation Perturbation}.

The \textit{Feature Perturbation} is a technique used to evaluate the relative importance of the inputs features in machine learning model, testing the model's tolerance to different values of data perturbations. This technique involves perturbing all instances of an feature by changing values or adding noise \citep{molnar2020interpretable,robnik2018perturbation_1}. The perturbations implemented were: sign inversion (multiplies all values by -1), application of bining (divides feature values into fixed-width bins), and application of standardization (modifies the feature value using z-score standardization).

%``Single Value Perturbation'': técnica utilizada para avaliar a sensibilidade do modelo de aprendizagem de máquina a mudanças pontuais nos dados de entrada. Nessa técnica, um único valor específico é atribuido a todas instâncias de um atributo, permitindo-se investigar o impacto dessa perturbação nos resultados do modelo \cite{chang2018explaining_perturbation_2}. As perturbações implementadas foram: zeros (valor 0 em todas instâncias) e desvio padrão (valor do desvio padrão em todas instâncias);

\textit{Single Value Perturbation} is a technique used to evaluate the machine learning model's sensitivity to specific changes in the input data. In this technique, a single specific value is assigned to all instances of an feature, allowing to investigate the impact of this perturbation on the model results \citep{chang2018explaining_perturbation_2}. The perturbations implemented were: zeros (value 0 in all instances) and standard deviation (standard deviation value in all instances).

%\item ``Perturbação por Permutação'': técnica de perturbação de entradas usada para avaliar a importância dos atributos em modelos de aprendizado de máquina dos tipos baggin e boosting. Essa técnica envolve a permutação aleatória ou não aleatória dos valores de um atributo específico \cite{breiman2001random_perturbation_3,permutation_feature_importance_perturbation_3}. As perturbações implementadas foram: ordenação crescente (ordena os valores de um parâmetro para forma crescente) e inverção (modifica os valores de um parâmetro invertendo o vetor do atributo de cima para baixo).

Lastly, the \textit{Permutation permutation} is a most common perturbation technique used to evaluate the feature importance in bagging machine learning models. This technique involves random (or non-random) permutation of the values in specific feature \citep{breiman2001random_perturbation_3,permutation_feature_importance_perturbation_3}. The perturbations implemented were: ascending ordering (orders the values of a feature in ascending order) and inversion (modifies the values of a parameter by inverting the feature vector from top to bottom).

%removido
%Note, optou-se por utilizar perturbações de natureza não aleatória a fim de tornar o processo ``Model Perturbation'' mais estável, sendo capaz de gerar perturbações iguais (em igual proporção), mesmo em casos onde se tenha modelos diferentes.
%Note, we chose to use perturbations of a non-random nature in order to make the ``Model Perturbation'' process more stable, being able to generate equal perturbations (in equal proportions), even in cases where there are different models.

%removido
%Continuing the explanation of the figure \ref{fig_metodologia_eXirt} (D), illustratively, figures of several black-box models are shown, but it must be understood that only 1 (one) model is used iteratively in this step.

%resumido
The creation of the response matrix, figure \ref{fig_metodologia_eXirt} (E) and (F), contains the answers of all iterations of model (understood as different \textit{IRT} respondents). The columns refer to the instances of the dataset that was passed on, while the rows refer to the different answers. %Note, values equal to ``0'' (zero) are wrong answers of the prediction, while values ``1'' (one) are correct answers of the prediction. This matrix is used to calculate the values of the item parameters (discrimination, difficulty and guessing) for each instance, in figure \ref{fig_metodologia_eXirt} (G).

The implementation of the \textit{IRT} used was \cite{cardoso2020decoding_irt}, called decodIRT, in a code developed exclusively for the purpose of this paper, as the code first receives the answer matrix, performs the calculations to generate the item parameter values --- different algorithms can be used to calculate the \textit{IRT} (such as: ternary, dichotomous, fibonacci, golden, brent, bounded or golden2) \citep{catsim} --- and, after this step, generates the rank of most skilled classifiers, figure  \ref{fig_metodologia_eXirt} (G).

Among the new features of \textit{decodIRT} is a new score calculation that involves the calculated ability of all respondents and their respective hits and misses, called \textit{Total Score}. This score can be understood as an adaptation of the \textit{True-Score} \citep{lord1984comparison}, whereby the score is calculated by summing up all the hit probabilities for the test items. However, in cases where respondents have a very close ability, the True-Score result can be very similar or even equal, since only the hit chance is considered. To avoid equal score values and to give more robustness to the models' final score, the \textit{Total Score} also considers the respondent's probability of error, given by: $1- P(U_{ij} = 1\vert\theta_{j})$. By rewarding the model when it gets the classification right and penalizing every time it misses the classification of an instance, the objective is to show when the disturbance in a certain feature impacted the classification.

Thus, every time the model gets it right, the hit probability is added, and if the model gets it wrong, the error probability is subtracted. The calculation of the Total Score $t$ is defined by the Equation \ref{eqn:total_score}, where $i'$ corresponds to the set of items answered correctly, and $i''$ corresponds to the set of items answered incorrectly.

\begin{equation}
\label{eqn:total_score}
t = \sum_{i=1}^{i'} P(U_{ij} = 1\vert\theta_{j}) - \sum_{i=1}^{i''} 1 - P(U_{ij} = 1\vert\theta_{j})
\end{equation}

In this regard, a skilled model with high hit probability that ends up getting an item wrong will not have its score heavily discounted. However, for a low ability model with low hit probability, the error will result in a greater discount of the score value. For, it is understood that the final score value should consider both the estimated ability of the respondent and his own performance on the test.

The \textit{Total Score} resulting from the execution of decodIRT is not yet the final rank of explainability of the model, because in this case it is necessary to calculate the average of the skills found for each feature, figure \ref{fig_metodologia_eXirt} (H), involving the different perturbations of features used in the previous steps.

Ultimately, figure \ref{fig_metodologia_eXirt} (H) and (K), an explanation rank is generated where each feature appears with a skill value. In this case, the lower the ability values, the more the feature explains the analyzed model. Equation $T_{(f,r)}$ is presented, which represents the processes performed by \textit{eXirt}, equation \ref{eqn:exirt_sigma}. 

\begin{equation}
\label{eqn:exirt_sigma}
T_{(f,r)} = \sum_{j=1}^{(v*f)+1} e_{j} \sum_{i=1}^{r} (I_{ji}+p_{ji}) + \sum_{l=1}^{f}t
\end{equation}

Where $v$ refer to the number of perturbation used. The value of $j$ represents the respondent's index of the iteration in model perturbation. While $f$ represents the total input features of the model, where $(v*f)$ the total number of respondents with perturbations and $+1$ is the respondent without perturbation (original model response).

While $e_j$ represents the process of building the response matrix, which is used in the following estimation of item parameter $I_{ji}$, where $j$ is the analyzed respondent and $i$ is the analyzed instance, and $r$ is the quantity of items to be considered in the estimation. In this step, item parameter estimates are already considered local explanations of the analyzed model.

The calculation of the ability is performed in $p_{ji}$, where $j$ is the analyzed respondent and $i$ is the analyzed instance. Finally, there is the iterative process for calculating the \textit{Total Score} in $t$, where $f$ is the quantity of input features of the model. In this step, the global feature relevance rank is created.

%Observação importante, todos os processos envolvendo os cálculos da \textit{IRT} são realizados com base nas respostas da população de respondentes usada no eXirt. Como a resposta do modelo original sem perturbação está na população, esta é usada como baseline para as repostas geradas por meio de perturbações. Por isso, o \textit{eXirt} sempre irá considerar como a melhor resposta da população a resposta do modelo original sem perturbação. Permitindo assim, que os resultados  do processo de extração de parâmetros de item sejam utilizados como  explicações locais válidas para o modelo analisado, assim como o ranque global de relevância de atributo.
Important note, all processes involving \textit{IRT} calculations are performed based on responses from the respondent population used in \textit{eXirt}. As the response of the original model without perturbation is in the population, this is used as a baseline for the responses generated through perturbation. Therefore, \textit{eXirt} will always consider the response of the original model without perturbation as the best population response. Thus allowing the results of the item parameter estimation process to be used as valid local explanations for the analyzed model, as well as global rank of feature relevance.

Based on the above, it can be seen that the execution of the method \textit{eXirt} depends directly on the number of respondents in the iteration ``pertubation of model'' and also on the number of instances in the dataset. For this reason, 7 different ways of perturbing the inputs of the machine learning model were used, which already makes it possible to create a considerable number of respondents, figure \ref{fig_metodologia_eXirt} (C).

%Já que o \textit{eXirt} usa somente 1 modelo, e nele aplica perturbações iterativas a fim de criar ``respondentes virtuais'', pode-se considerar que as perturbações aumentam as complexidades (no sentido de dificuldade) das instâncias de entrada do modelo analisado. Ou seja, no mundo real este processo pode ser visto como se as complexidades das questões de um teste pudessem ser aumentadas gradativamente em várias provas, que em seguida são aplicadas a um 1 único indivíduo que terá sua estabilidade avaliada a partir de cada prova aplicada.
Since \textit{eXirt} uses only 1 model, and applies iterative perturbations to it in order to create ``virtual respondents'', it can be considered that the perturbations increase the complexities (in the sense of difficulty) of the input instances of the analyzed model. That is, in the real world this process can be seen as if the complexities of the questions of a test could be gradually increased in several tests, which are then applied to a single individual who will have their stability evaluated from each test applied.

With the above, since the operation of \textit{eXirt} is based on the estimation of \textit{IRT} properties (discrimination, difficulty and guessing), on the estimation of ability values, and on model perturbations, further processes will help: Identifying features that best explain how the model works in a global way; Identifying instances that explain how stable and reliable the model is, in a local and global way.

%newnewnew
%Esta pesquisa declara que todos os processos executados pelo método proposto, \textit{eXirt}, são compatíveis com os passos metodológicos realizados por métodos XAI model-agnostic. Porém, em virtude das análises que são apresentadas neste artigo serem focadas em modelos tree-ensemble, considera-se o \textit{eXirt} inicialmente model-specific.
%%\color{red}
This research declares that all processes performed by the proposed method, \textit{eXirt}, are compatible with the methodological steps performed by XAI model-agnostic methods. However, because the analyzes presented in this article are focused on tree-ensemble models, \textit{eXirt} is initially considered model-specific. More information about \textit{eXirt} in the section \textit{Supplementary Information} \textit{(F)}.

%\color{black}

%\color{violet}
\section{Results and Discussion}\label{label_resultados}

%Esta seção apresenta os resultados desta pesquisa divididos em três momentos diferentes: Seção \ref{data_profile} onde é apresentado um estudo sobre os perfis de datasets utilizados; Seção \ref{exirt_vs_xai} onde são apresentados resultados das comparações dos ranques globais de relevância de atributo gerados pelo \textit{eXirt} com os resultados dos demais métodos XAI; E a seção \ref{exirt_and_trust} que apresenta resultados específicos do método eXirt, relacionados a explicações locais de modelos. 
This section presents the results divided into three different moments: A study on the used dataset profiles is presented (subsection \ref{data_profile}); The results of comparisons of the global features relevance ranks generated by \textit{eXirt} with the results of the other XAI methods (subsection \ref{exirt_vs_xai}); And a specific results of the \textit{eXirt} method, related to local and global explanations of models (subsection \ref{exirt_and_trust}).

\subsection{Dataset's Profile}\label{data_profile}

The first step to better understand the results is observe clustering and multiple correspondence analysis processes being used. In clustering, 4 clusters were found, with the dataset arrangements presented:

\begin{itemize}
    \item\textit{Cluster 0:} ``Australian'',``credit-g'', ``haberman'', ``monks-problems-1'', ``monks-problems-2'', ``monks-problems-3'', and ``tic-tac-toe'';

    \item\textit{Cluster 1:} ``analcatdata-lawsuit'',``churn'', ``climate-model-simulation-crashes'', ``kc1'',``kc2'', ``kc3'', ``mc1'', ``mw1'', ``ozone-level-8hr'', ``pc1'', ``pc2'', ``pc3'', ``pc4'', and ``Satellite'';
    
    \item\textit{Cluster 2:} ``kr-vs-kp'', ``PhishingWebsites'', and ``SPECT'';
    
    \item\textit{Cluster 3:} ``banknote-authentication'', ``blood-transfusion-service-center'', ``delta-ailerons'',``diabetes'', ``eeg-eye-state'', ``heart-statlog'', ``ilpd'', ``ionosphere'', ``jEdit-4.0-4.2'',  ``mozilla4'', ``phoneme'', ``prnn-crabs'', ``qsar-biodeg'', ``sonar'', ``spambase'', ``steel-plates-fault'', and ``wdbc''.
\end{itemize}

\begin{figure*}[p]
\includegraphics[scale=0.62]{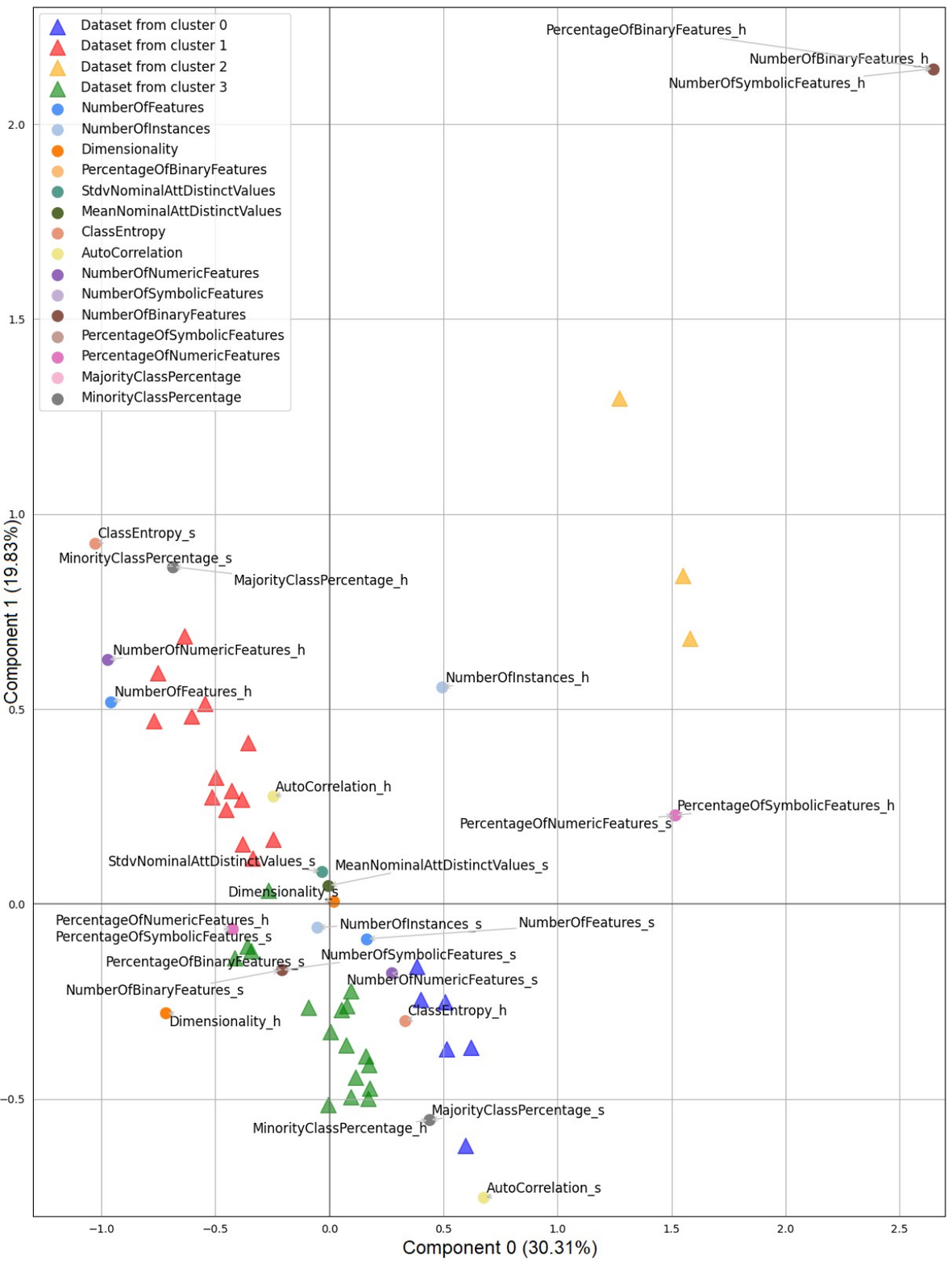}
\caption{Multiple Correspondence Analysis - MCA with rows (data sets) and columns (properties) the axis $x$ and $y$ are respectively the 0 and 1 components of the analysis.}
\label{fig_mca}
\end{figure*}

In order to consolidate the cluster profile analysis, a \textit{MCA} was performed, as advocated in the literature \cite{ref_mca}, and the relationship between the datasets in each cluster and the value ranges of the 15 properties analyzed was verified, figure \ref{fig_mca}.

%Com base na inspeção do resultado do MCA, mostrado na figura \ref{fig_mca}, pode-se notar diferentes relações e/ou influências de propriedades (circulos coloridos) específicas analisadas em relação ao espalhamento dos datasets (símbolos: triângulos azuis, lozangos vermelhos, estrelas laranjas, e vezes verdes) de cada cluster. A partir desta analise, pode-se entender melhor quais propriedades são mais significativas em cada cluster, ou seja, quais propriedades melhor diferenciam os datasets de um cluster para outro.
Based on the inspection of the \textit{MCA} result, shown in figure \ref{fig_mca}, one can notice different relationships and/or influences of specific properties (colored circles) analyzed in relation to the spread of datasets of each cluster (blue, red, orange, and green triangles). From this analysis, one can better understand which properties are most significant in each cluster, that is, which properties best differentiate the datasets from one cluster to another.

%Para caracterizar o perfil, não foi fixado um quantitativo igual de propriedades para cada cluster, mas sim usou-se as propriedades que mais influenciaram a posição dos datasets sobre as ordenadas das componentes 0 e 1. Entendendo-se que desta forma, as propriedades encontradas seriam as mais significativas por cluster:
To characterize the profile, an equal number of properties were not fixed for each cluster, but the properties that most influenced the position of the datasets on the ordinates of components 0 and 1 were used. The identified profiles were:

\begin{itemize}
    %\item \textbf{Cluster 0 profile:} datasets com maiores balanceamentos (``MajorityClassPercentage'' = ``s'' and ``MinorityClassPercentage'' = ``h''), menores autocorrelações (``AutoCorrelation'' = ``s''), maiores entropias na classe target (``ClassEntropy'' = ``h''), e menores valores de atributos númericos (``NumberOfNumericFeatures'' = ``s'');
    \item \textit{Cluster 0 profile:} datasets with higher class balances (``MajorityClassPercentage'' = ``s'' and ``MinorityClassPercentage'' = ``h''), lower autocorrelations (``AutoCorrelation'' = ``s''), higher entropy in the target class (``ClassEntropy'' = ``h''), and fewer numerical features (``NumberOfNumericFeatures'' = ``s'');
    
    %\item \textbf{Cluster 1 profile:} datasets com maiores desbalanceamentos (``MajorityClassPercentage'' = ``h'' and ``MinorityClassPercentage'' = ``s''), menores valores de entropia de classe (``ClassEntropy'' = ``s''), e maiores valores de número de atributos e também valores de atributos numéricos (``NumberOfNumericFeatures'' = ``h'' and ``NumberOfFeatures'' = ``h'');
    \item \textit{Cluster 1 profile:} datasets with larger class imbalances (``MajorityClassPercentage'' = ``h'' and ``MinorityClassPercentage'' = ``s''), smaller class entropy values (` `ClassEntropy'' = ``s''), many features (``NumberOfFeatures'' = ``h''), and many numeric features (``NumberOfNumericFeatures'' = ``h'');

    %\item \textbf{Cluster 2 profile:} datasets com os maiores valores de atributos não numéricos (``PercentageOfBinaryFeatures'' = ``h'', ``NumberOfBinaryFeatures''=``h'', and ``NumberOfSymbolicFeatures'' = ``h'');
    \item \textit{Cluster 2 profile:} datasets with the largest non-numeric features values (``PercentageOfBinaryFeatures'' = ``h'', ``NumberOfBinaryFeatures''=``h'', and ``NumberOfSymbolicFeatures' ' = ``h'');
    
    %\item \textbf{Cluster 3 profile:} datasets com os menores valores de atributos não numéricos e menores valores de número de instâncias (``NumberOfFeatures'' = ``s'', ``NumberOfSymbolicFeatures'' = ``s'', ``NumberOfBinaryFeatures'' = s, and ``NumberOfInstances'' = ``s'').
    \item \textit{Cluster 3 profile:} few datasets with non-numeric features and few instances (``NumberOfFeatures'' = ``s'', ``NumberOfSymbolicFeatures'' = ``s'', ``NumberOfBinaryFeatures'' = s, and ``NumberOfInstances'' = ``s'').
    
\end{itemize}

It is worth noting that some properties were not mentioned above because they appear at very similar distances for the four clusters.

%removido
%A partir de cada 4 clusters e seus respectivos perfis, nas próximas etapas do pipeline a fim de se comparar as explicações do \textit{eXirt} com as explicações dos demais métodos de XAI (seção \ref{exirt_vs_xai}) e mais a frente gerou-se explicações locais exclusivas do \textit{eXirt} para datasets selecionados destes clusters (seção \ref{exirt_vs_trust}).
%From each 4 clusters and their respective profiles, in the next stages of the pipeline in order to compare the \textit{eXirt} explanations with the other XAI methods (section \ref{exirt_vs_xai}) and later on, local and global explanations were generated \textit{eXirt} exclusives for selected datasets from these clusters (section \ref{exirt_and_trust}).

%%\color{violet}

\subsection{The \textit{eXirt} vs XAI Methods}\label{exirt_vs_xai}

%Os resultados apresentados abaixo mostram as comparações dos ranques gerados pelo \textit{eXirt} com os ranques gerados pelas demais medidas de XAI, isto levando-se em conta os diferentes clusters (C0 até C3) e seus datasets. Se não fosse o processo de clusterização e identificação dos perfis dos datasets, se tornaria dificultosa a avaliação das correlações do modelos baseados nos 41 datasets por meio de boxplot, pois os espalhamentos e outliers impediriam qualquer análise.
The results presented below show the comparisons of the ranks generated by \textit{eXirt} with the ranks generated by the other XAI measures, this taking into account the different clusters datasets ``C0'', ``C1'', ``C2'', and ``C3''. Note, without the process of clusterization and identification of datasets profiles, it would be difficult to evaluate the correlations of the models based on the 41 datasets through boxplots, as the scattering and outliers would prevent any analysis.

%A ideia é mostrar a estabilidade dos ranques de relevância de atributos gerados pelo eXirt, frente a diferentes perfis de datasets e também diferentes algoritmos de aprendizagem de máquina (M1: LightGBM, M2: CatBoost, M3: Random Forest, and M4: Gradient Boosting). Mostrando também que os resultados gerados pelo \textit{eXirt} não são replicações de resultados encontrados por outras medidas de XAI já existentes.
The idea is to show the stability of feature relevance ranks generated by \textit{eXirt}, against different dataset profiles and also different machine learning algorithms. Also showing that the results generated by \textit{eXirt} are not replications of results found by other existing XAI methods, even in conditions where the other XAI methods agree with each other, as seen in \citep{ribeiro_complexity_et_al_2021}. For this, each of the 4 models were created using their default settings indicated by their documentation, the models are: ``M1'' is \textit{Light Gradient Boosting} \citep{lightgbm}, ``M2'' is \textit{CatBoost} \citep{catboost}, ``M3'' is \textit{Random Forest} \citep{breiman2001random_perturbation_3}, and ``M4'' is \textit{Gradient Boosting} \citep{natekin2013gradientboost_gb}.

%Para facilitar as análises, irá se considerar a escala de correlação de Spearman, que definida com os seguintes valores: 0 --- 0.19 (very weak), 0.20 --- 0.39 (weak), 0.40 --- 0.59 (moderative), 0.60 --- 0.79 (strong), and 0.8 --- 1.0 (very strong). A fim de se avaliar a validade estatística dos resultados, também serão considerarados os \textit{p-values} gerados nos cálculos de correlações, já que a partir dele dos \textit{p-values} se torna possível a identificação dos valores de intervalos de confiança \textit{ic} estatística de determinada afirmação em porcentagem, $ic = (1 -$ \textit{p-value}) $* 100$.  
To facilitate the analyses, the \textit{Spearman correlation} (\textit{sp}) scale will be considered: 0 --- 0.19 (very weak), 0.20 --- 0.39 (weak), 0.40 --- 0.59 (moderative), 0.60 --- 0.79 (strong), and 0.8 --- 1.0 (very strong). In order to evaluate the statistical validity of the results, the \textit{p-values} generated in the correlation calculations will also be considered, since from this \textit{p-values} it becomes possible to identify Confidence Intervals \textit{ic} statistics of a given statement in percentage, $ic = (1 -$ \textit{p-value})$* 100$.

%Em momentos específicos se recorrerá a medidas de desempenho como acurária ou mesmo recall, a fim de se apontar indícios que mostram o modelo com os melhores desempenhos. Sobre os resultados desta medidas de performance dos modelos, será realizado o Teste de Friedman \cite{friedman} a fim de se verificar a significância estatísticas das doferentes performances dos modelos. 
At specific times, performance measures such as accuracy, precision, and recall will be used in order to present evidence that shows the model with the best performances. On the results of these three model performance measures, the \textit{Friedman Test} \citep{demvsar2006statistical_compare_mult_classifier_friedman} will be carried out in order to verify the statistical significance of the different model performances.

\subsubsection{Model performance}

\begin{figure*}[!h]
\begin{center}
\includegraphics[scale=0.53]{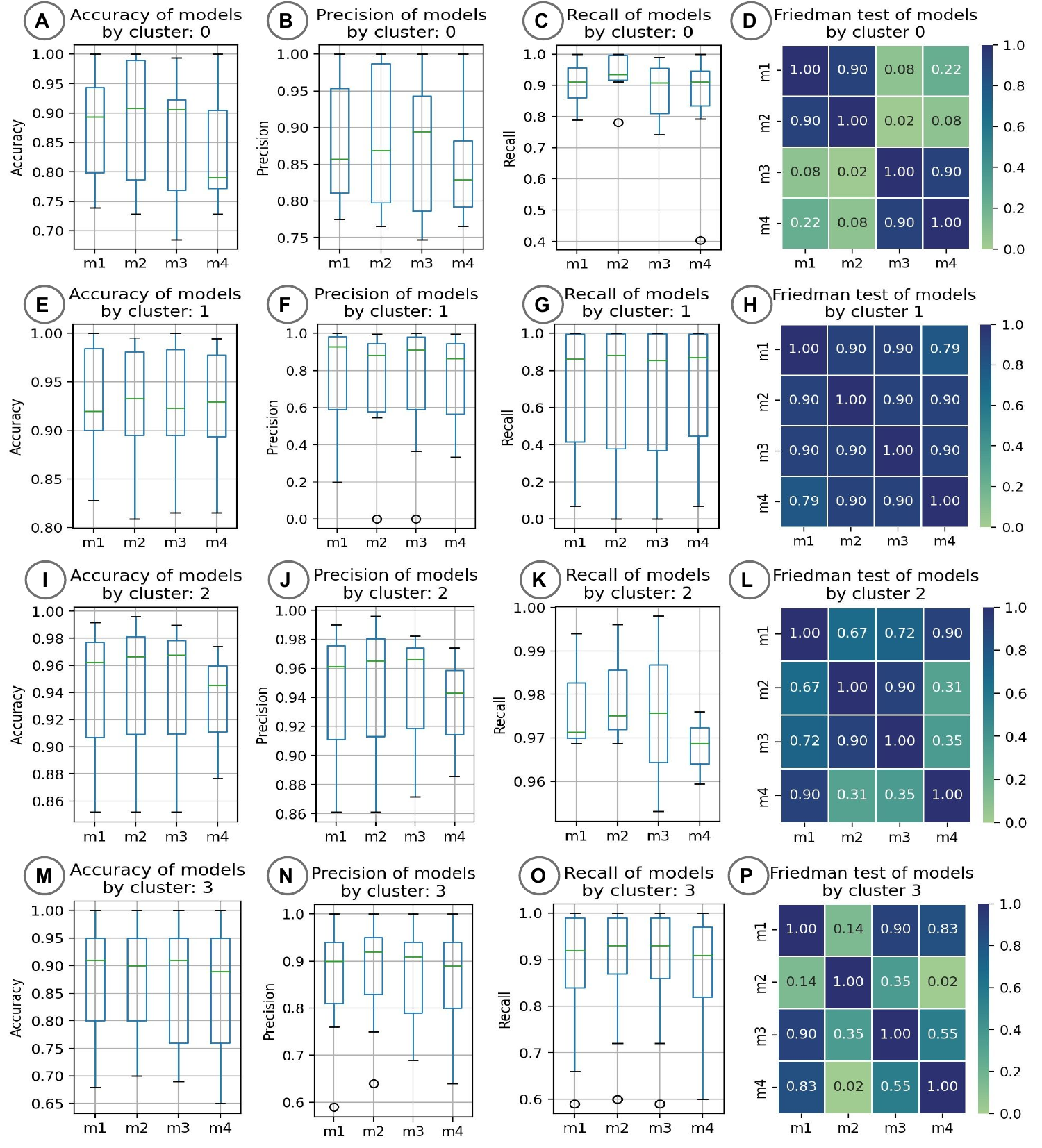}
\caption{\textit{Accuracy, precision}, \textit{Recall} and \textit{p-values} (Friedman Test) of the models ``M1'', ``M2'', ``M3'', and ``M4'' for all clusters.}
\label{fig_painel}
\end{center}
\end{figure*}

%Tomando-se como base os modelos ``M1'', ``M2'', ``M3'', and ``M4'', buscou-se realizar análises referentes a performance encontrado-se valores de acuária, precision e recall. Sobre os valores encontrados, executou-se o teste de frieman, a fim de se encontrar se as diferenças existentes entre os diferentes modelos são estatisticamente significantes. O resultado desta análise é disposto na figura \ref{fig_painel}.
Taking as a basis the models ``M1'', ``M2'', ``M3'', and ``M4'' from different clusters, analyzes of their performances were carried out, finding values of accuracy, precision and recall. In the values found, the Frieman test was performed in order to find the differences between the models, indicating whether these are statistically significant. The result of this analysis is shown in figure \ref{fig_painel}.

%newnewnew
%Inicia-se as análises dos modelos do cluster ``C0'', observando-se os valores de accuracy, precision and recall dispostos na figura \ref{fig_performance_c0}, ficando evidente que o modelo ``M2'' foi o que apresentou a melhor performance. 
The analysis of the models in the ``C0'' cluster begins, \ref{fig_painel} (A-C), observing the accuracy, precision and recall values displayed, making it evident that the ``M2'' model was the one that presented the best performance.

%Conforme apresentado na figura \ref{fig_performance_c0}, os valores de acurácia flutuam entre $\approx0.99$ e $\approx0.68$, considerando-se os limites inferiores e superiores de cada boxplot. Mostrando que todos os modelos apresentam consideravel performance. Porém, observando-se os intervalos em que os valores de precisão e o recall fluturam, entre $\approx0.99$ e $\approx0.74$, nota-se que os modelos apresentaram dificuldades na predição correta da classe positiva (1).
As shown in figure \ref{fig_painel} (A-C), accuracy values fluctuate between $\approx0.99$ and $\approx0.68$, considering the lower and upper limits of each boxplot. Showing that all models present considerable performance. However, observing the intervals in which the precision and recall values fluctuate, between $\approx0.99$ and $\approx0.74$, it is noted that the models presented difficulties in correctly predicting the positive class (1) .

%Buscando dar um embasamento estatístico na comparação dos modelos, figura \ref{fig_friedman_c0}, o \textit{Test de Friedman} foi realizado e verificou-se que o modelo ``M2'' apresentou diferenças com elevados intervalor de confiança para os modelos ``M3'' (\textit{p-value} $=0.02 \therefore ic=98\%$) e ``M4'' (\textit{p-value} $=0.08 \therefore ic=92\%$). Podendo-se afimar que a diferença de performance do modelo ``M2'' para o modelo ``M1'' não é significante, já as diferenças do modelo ``M2'' para os modelos ``M3'' e ``M4'' são estatisticamente significantes.
Seeking to provide a statistical basis for comparing the models, figure \ref{fig_painel} (D), the \textit{Friedman test} was carried out and it was found that the ``M2'' model presented differences with high confidence intervals for the models `M3'' (\textit{p-value} $=0.02 \therefore ic=98\%$) and ``M4'' (\textit{p-value} $=0.08 \therefore ic=92\%$). It can be stated that the difference in performance between the ``M2'' model and the ``M1'' model is not significant, as for the differences between the model ``M2'' for models ``M3'' and ``M4'' are statistically significant.

%newnewnew
%Com base nos resultados apresentados nas figuras \ref{fig_performance_c0} e \ref{fig_friedman_c0}, as próximas análises referentes aos modelos pertencentes ao cluster ``C0'' focarão nos modelos ``M1'' e M2'', já que estes apresentam as melhores performances sobre os datasets utilizados. Porém, as análises são apresentados na íntegra (todos modelos) em \textit{Supplementary Information} (E).
Based on the results presented in figures \ref{fig_painel} (A-D), the next subsections analyzes referring to the models belonging to the cluster ``C0'' will focus on the models ``M1'' and ``M2'', since these present the best performances on the datasets used. 

%################## C1

%newnewnew
%Na figura \ref{fig_performance_c1}, os valores de acurácia flutuam entre $\approx0.99$ e $\approx0.81$, considerando-se os limites inferiores e superiores de cada boxplot. O que mostra que todos os moelos apresentam elevada performance. Porém, observando-se os intervalos em que os valores de precisão e o recall fluturam, entre $\approx0.99$ e $\approx0.0$, nota-se que alguns dos modelos apresentaram problemas na predição correta da classe positiva (1). Isto já era esperado, devido os datasets do cluster ``C1'' terem alto desbalanceamento. 
In figure \ref{fig_painel} (E-G), the accuracy values fluctuate between $\approx0.99$ and $\approx0.81$, considering the lower and upper limits of each boxplot. Which shows that all models present high performance. However, observing the intervals in which precision and recall values fluctuate, between $\approx0.99$ and $\approx0.0$, it is noted that some of the models presented problems in correctly predicting the positive class (1). This was already expected, as the datasets in the ``C1'' cluster have high imbalance.

%newnewnew
%Por inspessão da figura \ref{fig_performance_c1}, não é possível notar clara diferença entre os modelos, sendo que na aplicação do \textit{Teste de Friedman}, figura \ref{fig_friedman_c1}, foi comprovada a não existência de diferença estatisticamente significante entre os modelos. Portanto, as próximas análises dos modelos pertencentes ao cluster ``C0'' abordarão os modelos ``M1'' até ``M4''. 
By inspecting the figure \ref{fig_painel} (H), it is not possible to notice a difference between the models, and when applying the \textit{Friedman test}, it was proven that there was no statistically significant difference between the models. Therefore, the next subsections analyzes of the models belonging to the ``C0'' cluster will address models ``M1'' to ``M4''.

%################## C2

%newnewnew
%Analisando-se as performances dos modelos que são mostrados na figura \ref{fig_performance_c2}, pode-se verificar que os modelos ``M1'', ``M2'' e ``M3'' apresentam resultados similares entre si, já o modelo ``M4'' apresenta uma performance ligeiramente menor que os demais. Porém, ao se analisar os \textit{p-values} provenientes do \textit{Teste de Friedman}, nota-se que a diferença existente entre os modelos em questão não é estatisticamente significante (observe os \textit{p-values} entre $0.31$ e $1$), figura \ref{fig_friedman_c2}.
Analyzing \ref{fig_painel} (I-L), it can be seen that models ``M1'', ``M2'' and ``M3'' present similar results to each other, while the Model ``M4'' presents a slightly lower performance than the others. However, when analyzing the \textit{p-values} from the \textit{Friedman Test}, it is noted that the difference between the models in question is not statistically significant (note the \textit{p-values} between $0.31$ and $1$).

%newnewnew
%Na figura \ref{fig_performance_c2}, os valores de acurácia, precisão e recall flutuam entre $\approx0.86$ e $\approx0.99$, considerando-se os limites inferiores e superiores de cada boxplot. O que mostra que todos os modelos apresentam elevada performance quanto aos problemas a que são aplicados em ambas as classes (0 e 1).
In figure \ref{fig_painel} (I-K), the accuracy, precision and recall values fluctuate between $\approx0.86$ and $\approx0.99$, considering the lower and upper limits of each boxplot. This shows that all models present high performance regarding the problems to which they are applied in both classes (0 and 1).

%################## C3

%newnewnew
%Considerando os valores de acurária, precisão e recall calculados para cada um dos modelos de ``C3'', figura \ref{fig_performance_c3}, pode-se perceber valores de acurácia similares entre os modelos ``M1'', ``M2'' e ``M3'' (medianas mais elevadas), assim como valores ligeiramente mais baixos em ``M4''. Já precisão e o recall, evidenciaram maiores diferenças de performance dos modelos, apresentando novamente os valores mais baixos para ``M4''.
Considering the accuracy, precision and recall values calculated for each of the ``C3'' models, figure \ref{fig_painel} (M-O), similar accuracy values can be seen in the models ``M1'', ``M2'' and ``M3'' (higher medians), as well as slightly lower values in ``M4''. Precision and recall, on the other hand, showed greater differences in the performance of the models, again presenting the lowest values for ``M4''.

%Na figura \ref{fig_performance_c3}, os valores de acurácia, precisão e recall flutuam entre $\approx0.6$ e $\approx0.99$, considerando-se os limites inferiores e superiores de cada boxplot. Apesar da diferença visual dos boxplots ser minima, a análise estatística baseada no \textit{Teste Friedman} foi capaz de evidenciar diferenças estatisticamente significantes entre os modelos ``M2'' e ``M4'' \ref{fig_friedman_c3}.
In figure \ref{fig_painel} (M-P), the accuracy, precision and recall values fluctuate between $\approx0.6$ and $\approx0.99$, considering the lower and upper limits of each boxplot. Although the visual difference in the boxplots is minimal, the statistical analysis based on the \textit{Friedman Test} was able to highlight statistically significant differences between the models ``M2'' and ``M4''.

%newnewnew
%Observando-se os \textit{p-values} apresentados na figura \ref{fig_friedman_c3}, percebe-se que os modelos ``M1'' (\textit{p-value} $=0.14 \therefore ic = 86\%$), ``M3'' (\textit{p-value} $=0.35 \therefore ic = 65\%$) não apresentam diferença estatisticamente significante (\textit{p-value} $< 0.05$) em relação ao modelo ``M2''. Sendo que ``M4'' (\textit{p-value} $=0.02 \therefore ic = 98\%$) apresenta diferença estatisticamente significamente se comparado com o modelo ``M2''. Portando, observando-se os resultados do \textit{Teste de Friedman} e a surperioridade (mesmo que mínima) de performance dos ``M1'', ``M2'', e ``M3'' define-se estes modelos como os que apresentaram melhores performances sobre os dados do cluster ``C3''.
Observing the \textit{p-values} presented in figure \ref{fig_painel} (P), it is clear that the models ``M1'' (\textit{p-value} $=0.14 \therefore ic = 86\%$ ), ``M3'' (\textit{p-value} $=0.35 \therefore ic = 65\%$) do not present a statistically significant difference (\textit{p-value} $< 0.05$) in relation to the model ` `M2''. Therefore, ``M4'' (\textit{p-value} $=0.02 \therefore ic = 98\%$) presents a statistically significant difference compared to the model ``M2''. Therefore, observing the results of the \textit{Friedman Test} and the superiority (even if minimal) of performance of ``M1'', ``M2'', and ``M3'', these models are defined as those that presented better performances on data from the ``C3'' cluster.

\subsubsection{Models by cluster ``C0''}

%Antes de analisar os resultados do pipeline para o cluster ``C0'', precisa-se considerar que os modelos baseados nestes datasets tendem a apresentar predições mais precisas em função de duas características principais do perfil do cluster, que é o fato destes datasets apresentarem: valores mais balanceados de classe e também apresentarem alta entropia de classe.
Before analyzing the results of the pipeline for the ``C0'' cluster, it should be considered that the models based on these datasets tend to present more accurate predictions based on two main characteristics of the cluster profile, which is the fact of these datasets they will present: more balanced class values and also present high class entropy \citep{zhou2021machine_book}.

\begin{figure*}[!h]
\begin{center}
\includegraphics[scale=0.55]{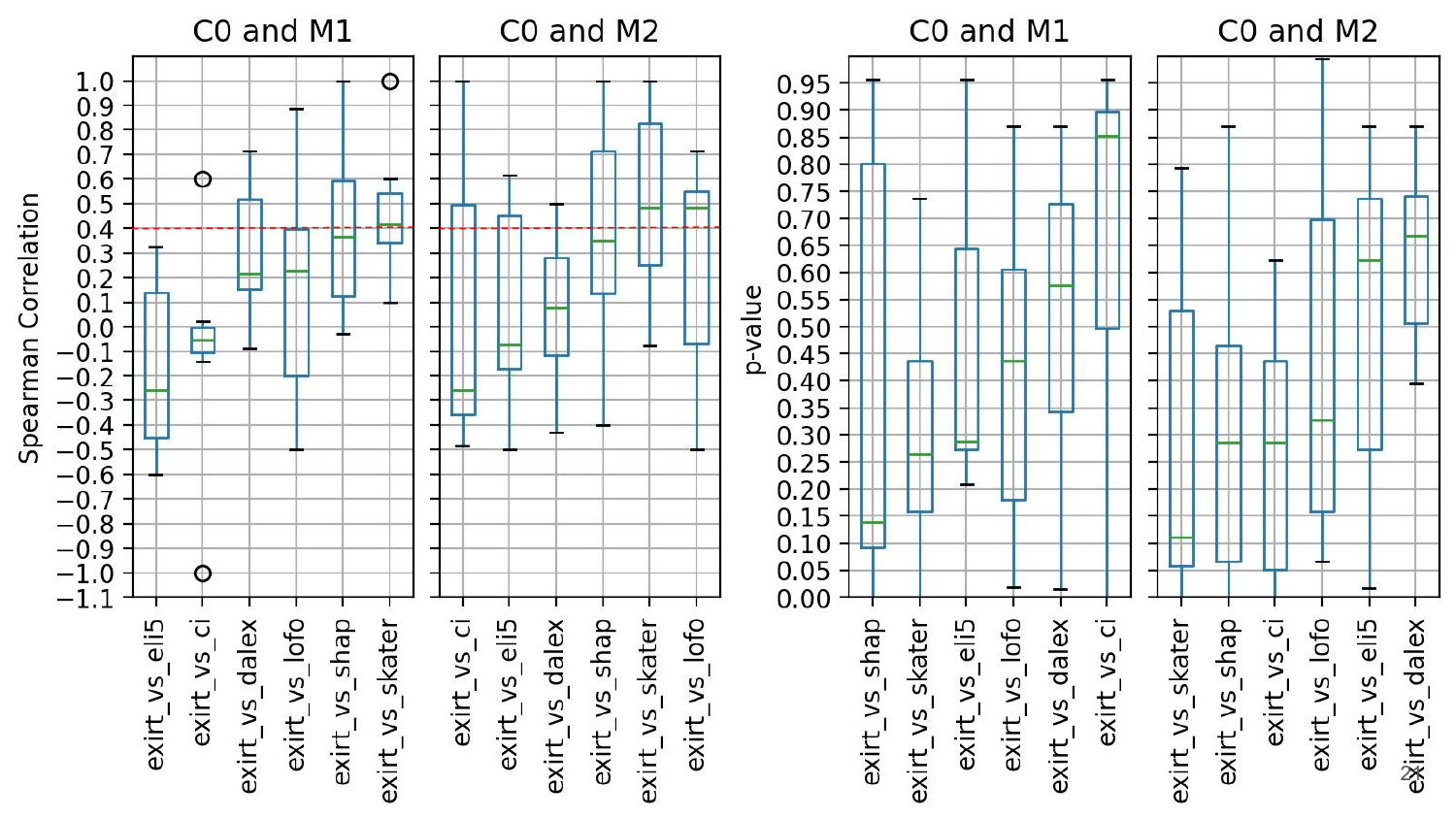}
\caption{Results of \textit{Spearman} correlations (left) and \textit{p-values} (right) for models ``M1'' and ``M2'' based on the ``C0'' cluster datasets. On the left, the dashed red line divides the correlations between above and below ``moderate''. On the right, values close to 0.05 are more significant.}
\label{fig_resume_c0}
\end{center}
\end{figure*}

%newnewnew
%A figura \ref{fig_resume_c0} a esquerda, mostra as flutuações dos valores de correlações de spearman encontratos a partir das comparações dos ranques criados pelo \textit{eXirt} e os demais métodos de XAI. Já a direita, são mostradas as flutuações dos \textit{p-values} encontrados a cada comparação, indicando se determinandas comparações apresentam significância estatistica.
The figure \ref{fig_resume_c0} on the left shows the fluctuations in the Spearman correlation values found from the comparisons of the ranks created by \textit{eXirt} and the other XAI methods. On the right, the fluctuations in the \textit{p-values} found in each comparison are shown, indicating whether certain comparisons are statistically significant.

%newnewnew
%É possível notar, figura \ref{fig_resume_c0} esquerda, que o \textit{eXirt} apresentou correlações acima de 0.4 (linha vermelha tracejada) em algumas comparações de ranques (boxplots que ultrapassam linha tracejada vermelha). Porém, observando-se a figura \ref{fig_resume_c0} direita, que as as medianas  dos \textit{p-values} (linhas verdes dos boxplots) apontam as comparações com resultados mais significantes, dentro de intervalos de confiança específicos, que são: 
It is possible to notice, figure \ref{fig_resume_c0} on the left, that \textit{eXirt} presented correlations above 0.4 (dashed red line) in some ranking comparisons (boxplots that exceed the red dashed line). However, observing the figure \ref{fig_resume_c0} on the right, the medians of the \textit{p-values} (green lines of the boxplots) point to comparisons with more significant results, within specific confidence intervals, which are:

\begin{itemize}
    \item ``M2'': \textit{skater} (medians of $sp=0.44$ and \textit{p-value} $=0.11 \therefore ic=88\%$).
    \item ``M1'': \textit{shap} (medians of $sp=0.38$ and \textit{p-value} $=0.14 \therefore ic=86\%$);
\end{itemize}

%newnewnew
%Os valores de correlações $0.38$ e $0.44$ são considerados respectivamente ``weak'' e ``moderative'', seguindo a escala da correlação de \textit{Spearman}, mostrando que o \textit{eXirt} foi capaz de gerar ranques de explicações dos modelos de aprendizagem de máquina pertencentes ao cluster ``C0'' de maneira moderadamente correlacionada as explicações dos demais métodos de XAI existentes --- em outras palavras, as explicações do \textit{eXirt} são diferentes das explicações geradas pelos demais métodos de XAI, observando-se modelos do cluster ``C0''.
The correlation values $0.38$ and $0.44$ are considered respectively ``weak'' and ``moderative'', following the \textit{Spearman} correlation scale, showing that \textit{eXirt} was able to generate explanations of the machine learning models belonging to the ``C0'' cluster in a moderately correlated way to the explanations of the other existing XAI methods --- in other words, the explanations of \textit{eXirt} are different from the explanations generated by the other methods of XAI, observing models from the ``C0'' cluster.

%newnewnew
%Fica evidente, observando-se os boxplots da figura \ref{fig_resume_c0}, que os resultados das explicações geradas por todas os métodos de XAI testados são diferentes, mesmo se tratando de modelos computacionais que não apresentam diferença de performance estatisticamente significante. A prova desta afirmação, é que os boxplots dispostos em ``C0 and M1'' e ``C0 and M2'' são diferentes um dos outros.
It is evident, looking at the boxplots in figure \ref{fig_resume_c0}, that the results of the explanations generated by all the XAI methods tested are different, even when dealing with computational models that do not present a statistically significant difference in performance. The proof of this statement is that the boxplots arranged in ``C0 and M1'' and ``C0 and M2'' are different from each other.

%newnewnew
%Os diferentes valores de correlações e também diferentes tamanhos de boxplots, figura \ref{fig_resume_c0} esquerda, mostram de maneira indireta que os métodos de XAI presentes na literatura explicaram os modelos do cluster ``C0'' utilizando ranques diferentes um dos outros. Pois, ao se comparar dois boxplots por exemplo, quanto maiores as diferenças (tamanho e posição) entre eles, maiores serão as diferenças das explicações dos métodos. Isso, pode ser melhor observado, ao se analisar todas as comparações de ranques gerados na pesquisa, section \textit{Supplementary information} (E). 
The different correlation values and also different sizes of boxplots, figure \ref{fig_resume_c0} on the left, indirectly show that the XAI methods present in the literature explained the models in the ``C0'' cluster using different ranks from each other. Because, when comparing two boxplots for example, the greater the differences (size and position) between them, the greater the differences in the explanations of the methods. This can be better observed when analyzing all the ranking comparisons generated in the research, section \textit{Supplementary information} (E).

%O fato de cada método de XAI explicar um mesmo modelo e assim gerar diferentes explicações, impacta diretamente na confiabilidade do mesmo, pois se para um mesmo problema gera-se várias explicações diferentes, se torna difícil um humano confiar em uma determinada explicação. Isto mostra, o quão complexa é a missão dos explicadores de modelos de aprendizagem de máquina caixa preta.
The fact that each XAI method explains the same model and thus generates different explanations has a direct impact on its reliability, because if several different explanations are generated for the same problem, it becomes difficult for a human to trust a given explanation. This shows how complex the mission of explainers of black box machine learning models is.

\subsubsection{Models by cluster ``C1''}

%Antes de analisar os resultados do pipeline para o cluster ``C1'', precisa-se levar em consideração que naturalmente os modelos que se baseiam em datasets deste grupo tendem a apresentar tendências de previsão para uma classe específica, pois se tratam de datasets com alto desbalanceamento, assim como uma baixa entropia de classe. Outra característica dos datasets deste cluster, é o número alto de features, que impacta diretamente no processo de calculo das correlações dos rânques de relevância de atributo, já que ao se ter mais features se tem mais possibilidades de explicações do modelos que usa o dataset.
Before analyzing the results of the pipeline for the ``C1'' cluster, it should be taken into consideration that naturally the models that are based on datasets of this cluster tend to present forecast trends for a specific class, because they are datasets with high imbalance, as well as a low class entropy \citep{zhou2021machine_book}. Another characteristic of the datasets of this cluster, is the high number of features, which directly impacts the process of calculating the correlations between the ranks of feature relevance, since having more features has more possibilities of explanations of the model (higher possible combination of ranks).

\begin{figure*}[!h]
\begin{center}
\includegraphics[scale=0.55]{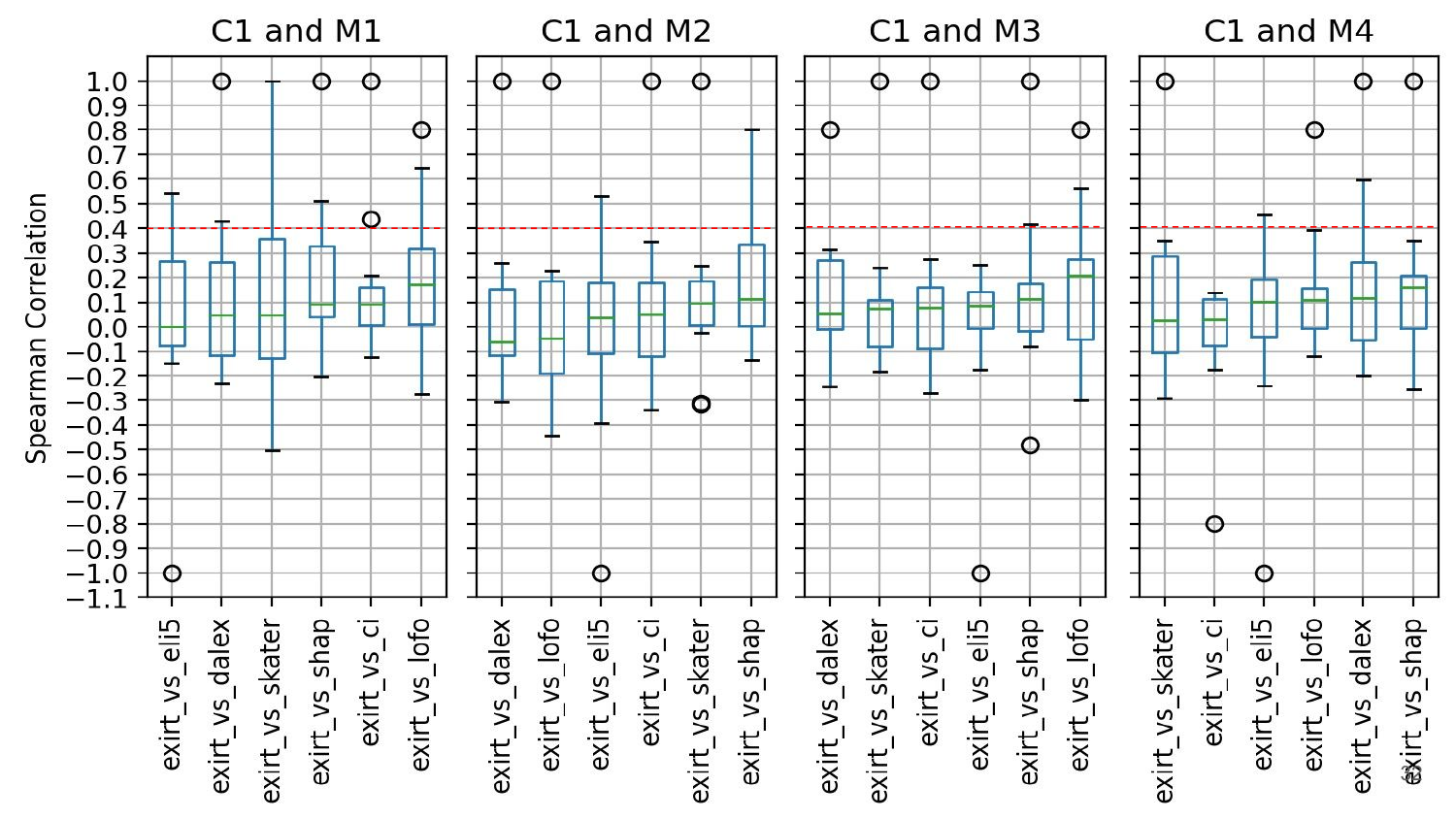}
\caption{Results of \textit{Spearman} correlations for models ``M1'', ``M2'' ``M3'', and ``M4'' based on the datasets cluster ``C1''. The dashed red line divides the correlations between above and below ``moderate''.}
\label{fig_resume_c1}
\end{center}
\end{figure*}

%Os resultados apresentados na figura \ref{fig_resume_c1} são geradados a partir dos modelos que usam datasets do cluster ``C1'', e mostram correlações ``weak'' ou mesmo ``very weak'' para a maioria dos ranques gerados pelo \textit{eXirt} em comparação com os demais métodos de XAI. O que significa dizer que para este grupo de modelos, o \textit{eXirt} gerou ranques de explicações extramamente diferentes dos demais métodos (com exceção de poucos outliers). 
The results shown in figure \ref{fig_resume_c1} are generated from models that use datasets from cluster ``C1'', and show ``weak'' or even ``very weak'' correlations for most of the ranks generated by the \textit{eXirt} compared to the other XAI methods. Which means that for this cluster of models, \textit{eXirt} generated extremely different ranks of explanations from the other methods (except for a few outliers).

%Uma particularidade da figura \ref{fig_resume_c1} é a quantitade elevada de outliers, o que indica que em pelo menos 1 execução, de cada boxplot, foi identificada uma correlação ``moderate'' ou mesmo ``strong'' diferente das demais.
A particularity of the figure \ref{fig_resume_c1} is the high number of outliers, which indicates that in at least 1 execution of each boxplot, a ``moderate'' or even ``strong'' correlation (positive or negative) different from the others was identified.

%newnewnew
%Como os resultados apresentados na figura \ref{fig_resume_c1} apontam para flutuações das correlações com força em torno de ``very weak' ($sp < 0.19$), não identifica-se a necessidade de avaliar os \textit{p-values} dado que eles somente mostrarão significancia estatisticas para correlações positivas ou negativas. Mesmo assim, as análises envolvendo os \textit{p-values} podem ser vistas na section \textit{Supplementary information} (E).
As the results presented in figure \ref{fig_resume_c1} point to fluctuations in correlations with strength around ``very weak'' ($sp < 0.19$), there is no need to evaluate the \textit{p-values} since they will only show statistical significance for positive or negative correlations. Even so, the analyzes involving \textit{p-values} can be seen in section \textit{Supplementary information} (E).

%Nos resultados referentes aos modelos do cluster ``C1'', ficam evidentes que as explicações geradas pelo \textit{eXirt} são extremamente diferentes das explicações geradas pelos demais métodos de XAI. Ficando evidente também que os atuais métodos de XAI geraram explicações diferentes entre si (notar a diferença de tamanhos dos boxplots).
In the results referring to the models in the ``C1'' cluster, it is evident that the explanations generated by \textit{eXirt} are extremely different from the explanations generated by other XAI methods. It is also evident that the current XAI methods generated different explanations (note the difference in the sizes of the boxplots).

%newnewnew
%Novamente os resultados mostram que mesmo para modelos de aprendizagem de máquina com performance similares (sem diferença estatistivamente significante) as explicações geradas pelos métodos de XAI são diferentes entre si.
Once again, the results show that even for machine learning models with similar performance (without statistically significant differences) the explanations generated by the XAI methods are different from each other.

\subsubsection{Models by cluster ``C2''}

%newnewnew
%Antes de se iniciar as análises dos modelos criados a partir dos datasets do cluster ``C2'', deve-se considerar que este é o cluster com o menor quantitativo de datasets (3 ao total), e estes datasets apresentam alto número de features não numéricas (em sua maioria binária). Detalhe, o fato de se ter um $n$ pequeno não impede a execução do pipeline, porém os boxplots serão mais sensíveis a mudanças das correlações.
Before starting the analysis of the models created from the datasets of the ``C2'' cluster, it should be considered that this is the cluster with the smallest number of datasets (3 in total), and these datasets have a high number of features non-numeric (mostly binary). Detail, the fact of having a small $n$ does not prevent the execution of the pipeline, however the boxplots will be more sensitive to changes in correlations.

\begin{figure*}[!h]
\begin{center}
\includegraphics[scale=0.55]{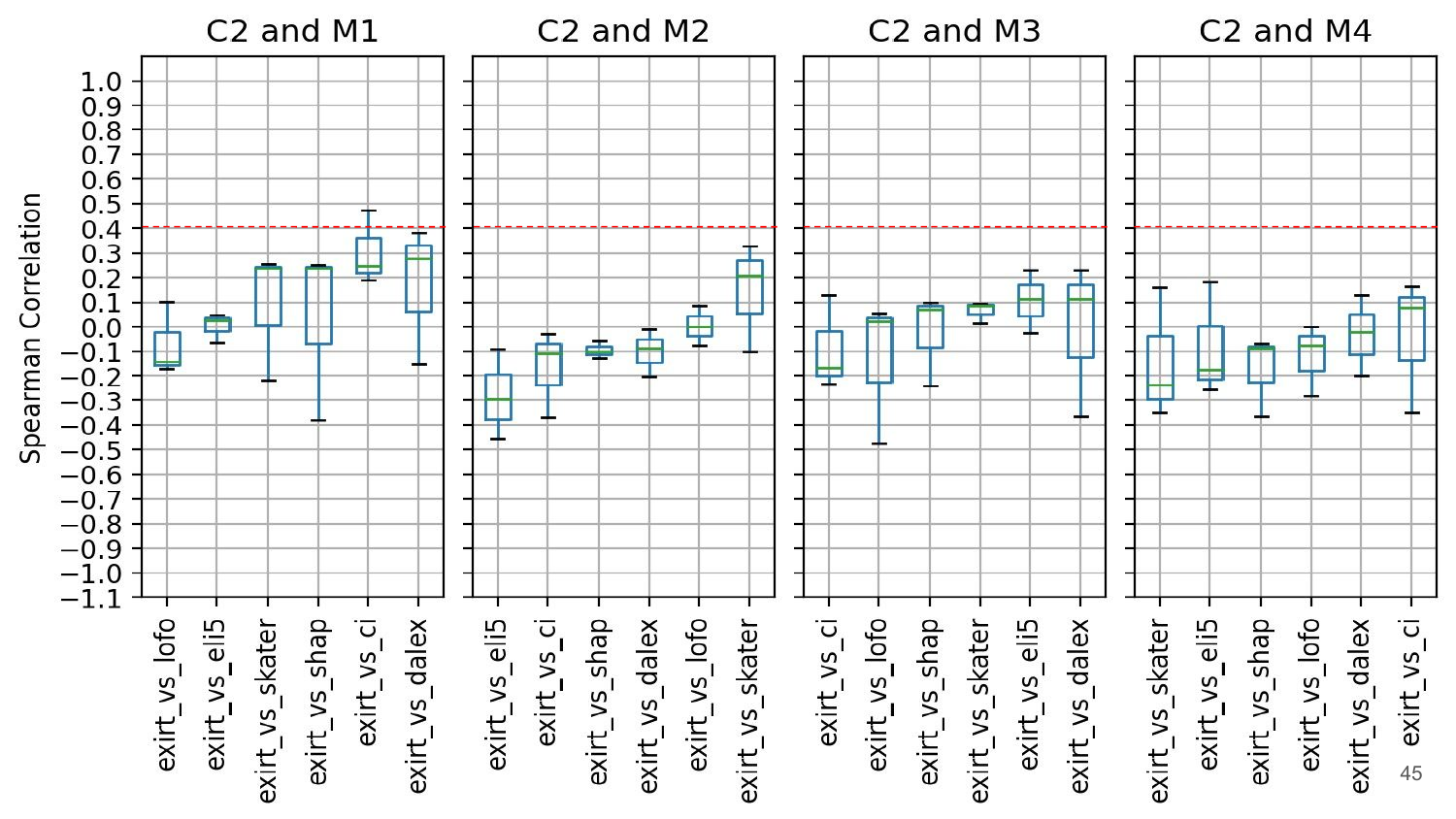}
\caption{Results of \textit{Spearman} correlations for models ``M1'', ``M2'' ``M3'', and ``M4'' based on the datasets cluster ``C2''. The dashed red line divides the correlations between above and below ``moderate''.}
\label{fig_resume_c2}
\end{center}
\end{figure*}

%newnewnew
%Os resultados apresetados anteriormente na figura \ref{fig_resume_c1} referentes ao ``C1'', são similares aos apresentados na figura \ref{fig_resume_c2}, porém sem a existência de outliers, o que indica que em todos as comparações realizadas, o \textit{eXirt} apresentou resultados extremamente diferentes dos demais métodos de XAI.
The results previously presented in figure \ref{fig_resume_c1} referring to ``C1'' are similar to those presented in figure \ref{fig_resume_c2}, but without the existence of outliers, which indicates that in all comparisons carried out, \ textit{eXirt} presented results that were extremely different from other XAI methods.

%newnewnew
%As correlações apresentadas na figura \ref{fig_resume_c2} apontam para flutuações das correlações com força em torno de ``very weak' ($sp < 0.19$), não identifica-se a necessidade de avaliar os \textit{p-values} dado que eles somente mostrarão significancia estatisticas para correlações positivas ou negativas. Mesmo assim, as análises envolvendo os \textit{p-values} podem ser vistas na section \textit{Supplementary information} (E).
The correlations presented in figure \ref{fig_resume_c2} point to fluctuations in correlations with strength around ``very weak' ($sp < 0.19$), the need to evaluate the given \textit{p-values} is not identified that they will only show statistical significance for positive or negative correlations. Even so, the analyzes involving \textit{p-values} can be seen in section \textit{Supplementary information} (E).

%newnewnew
%Com base nos resultados gerados nas análises dos modelos de ``C1'' e ``C2'', pode-se considerar os modelos provenientes destes dois clustes como os mais difíceis de serem explicados, pois de maneira geral, os métodos de XAI utilizados (incluindo o eXirt) geraram explicações diferentes um dos outros.
Based on the results generated in the analyzes of the ``C1'' and ``C2'' models, the models from these two clusters can be considered as the most difficult to explain, as in general, the XAI methods used (including \textit{eXirt}) generated different explanations from each other.

\subsubsection{Models by cluster ``C3''}

%Antes de analisar os resultados do pipeline para o cluster ``C3'', deve-se levar em conta uma principal característica deste cluster, que é o fato de ser composto por datasets com menores números de instâncias. Este menor número de instância deveria impactar os resultados do eXirt, pois com menas instâncias se tem a aplicação de um ``teste menor'' no modelo, porém nenhum tipo de impacto negativo foi notado nos resultados.
Before analyzing the results of the pipeline for the ``C3'' cluster, an important characteristic of this cluster must be taken into account, which is that it is made up of datasets with smaller numbers of instances. This lower number of instances should have an impact on the \textit{eXirt} results, as fewer instances reflect a ``minor test'' applied to the model, therefore no type of negative impact was noted on the results.

\begin{figure*}[!h]
\begin{center}
\includegraphics[scale=0.55]{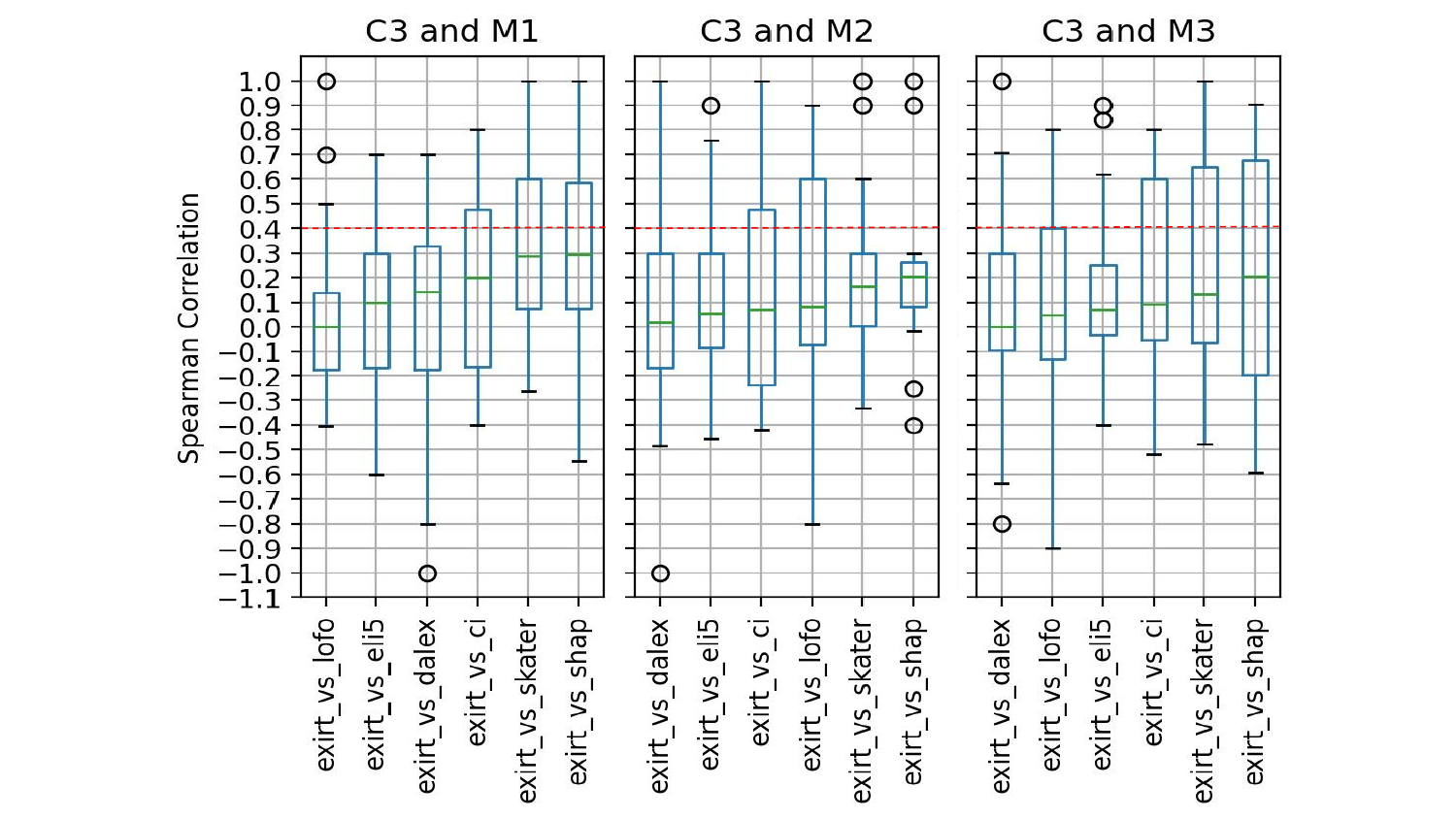}
\caption{Results of \textit{Spearman} correlations for models ``M1'', ``M2'', and ``M3'' based on the datasets cluster ``C3''. The dashed red line divides the correlations between above and below ``moderate''.}
\label{fig_resume_c3}
\end{center}
\end{figure*}

%newnewnew
%As análises a seguir, a partir da figura \ref{fig_resume_c3} em diante, focarão nos resultados dos modelos ``M1'', ``M2'' e ``M3'', já que estes apresentaram maiores performances conforme visto acima.
The following analyses, from figure \ref{fig_resume_c3} onwards, will focus on the results of models ``M1'', ``M2'' and ``M3'', as these presented greater performances as seen above.

\begin{figure*}[!h]
\begin{center}
\includegraphics[scale=0.55]{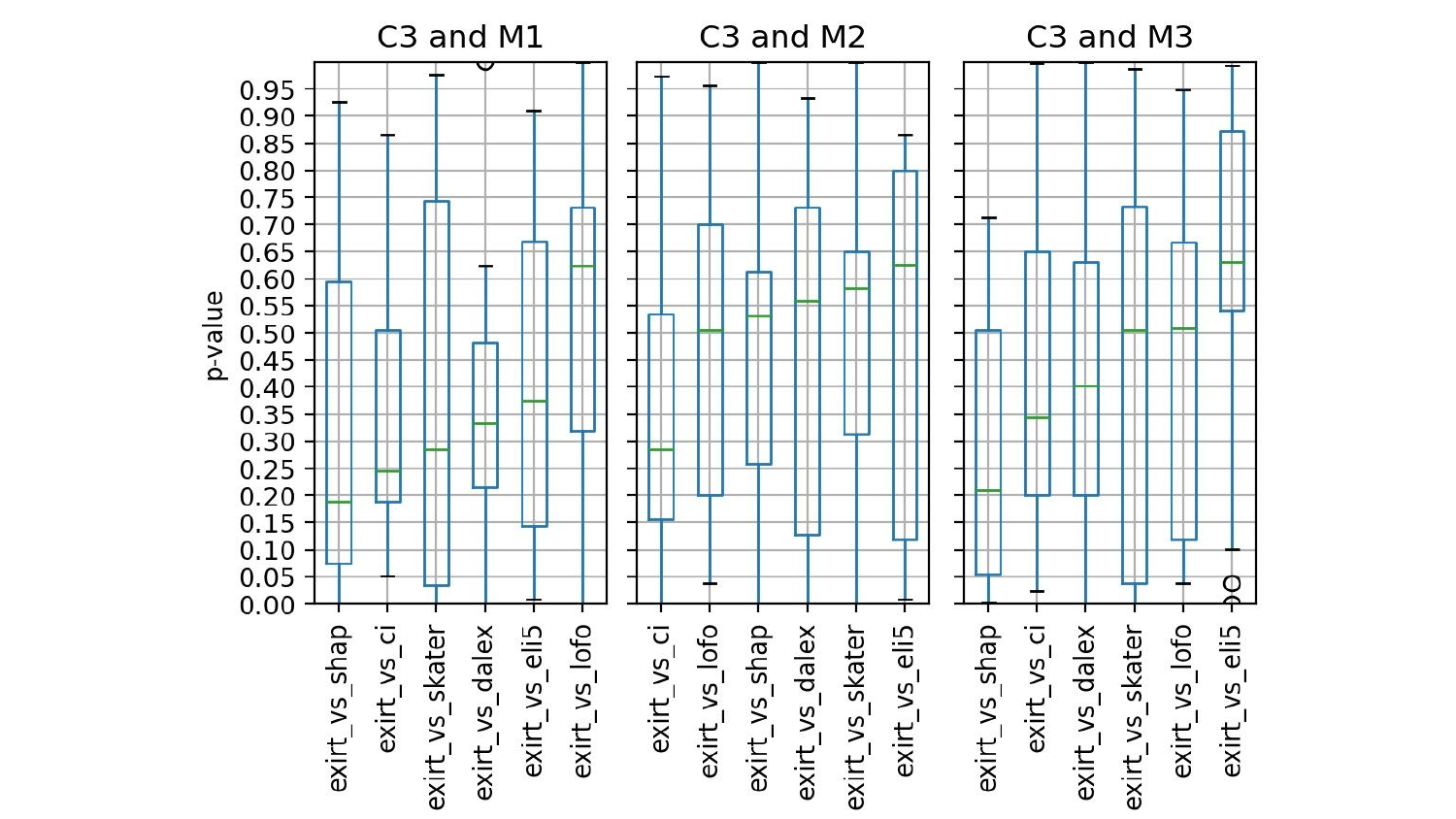}
\caption{Results of \textit{p-values} calculated in \textit{Spearman} correlations for models ``M1'', ``M2'', and ``M3'' based on the datasets cluster ``C3''. Note, values close to 0.05 are more significant. }
\label{fig_resume_c3_p}
\end{center}
\end{figure*}

%newnewnew
%Os resultados apresentados na figura \ref{fig_resume_c3}, mostram correlações com forças menores que ``weak'' ($sp < 0.39$) na maior parte dos resultados (observar linha tracejada vermelha). Porém, os terceiros quartis de algumas comparações ultrapassaram a linha tracejada vermelha, que indica correlação moderada. Assim para um melhor entendimento das correlações apresentadas, recorre-se aos \textit{p-values} gerados em cada comparação, \ref{fig_resume_c3_p}.
The results presented in figure \ref{fig_resume_c3} show correlations with strengths lower than ``weak'' ($sp < 0.39$) in most of the results (note red dashed line). However, the third quartiles of some comparisons exceeded the red dashed line, which indicates moderate correlation. Therefore, for a better understanding of the correlations presented, we use the \textit{p-values} generated in each comparison, figure \ref{fig_resume_c3_p}.

%Observando-se as menores medianas dos \textit{p-values} apresentados na figura \ref{fig_resume_c3_p}, pode-se destacar as seguintes correlações da figura \ref{fig_resume_c3} como mais significantes dentro de um intervalo de confiança específico:
Observing the lowest medians of the \textit{p-values} presented in figure \ref{fig_resume_c3_p}, the following correlations in figure \ref{fig_resume_c3} can be highlighted as most significant within a specific confidence interval:

\begin{itemize}
    \item ``M1'': \textit{shap} (medians of $sp=0.29$ and \textit{p-value} $=0.19 \therefore ic=81\%$);
    \item ``M3'': \textit{shap} (medians of $sp=0.2$ and \textit{p-value} $=0.21 \therefore ic=79\%$).
    \item ``M1'': \textit{ci} (medians of $sp=0.20$ and \textit{p-value} $=0.24 \therefore ic=76\%$);
    \item ``M2'': \textit{ci} (medians of $sp=0.07$ and \textit{p-value} $=0.29 \therefore ic=71\%$).
\end{itemize}

%newnewnew
%Observando-se os resultados listados acima, percebe-se que o intervalo de confiança ($ic$) não são tão expressivos, mostrando que mesmo em comparações onde o \textit{eXirt} apresentou as maiores correlações com os demais métodos de XAI, as correlações não apresentaram forças significante. 
Observing the results listed above, it is clear that the confidence interval ($ic$) are not as expressive, showing that even in comparisons where \textit{eXirt} presented the highest correlations with the other XAI methods, the correlations did not present significant strengths.

%Conforme visto nas figuras \ref{fig_resume_c0,fig_resume_c1,fig_resume_c2,fig_resume_c3}, os resultados encontrados pelo \textit{eXirt} para o processo de criação de explicações baseadas em ranque de relevância de atributos são diferentes dos resultados encontrados pelos demais métodos XAI presentes na literatura, mesmo diante de modelos (dataset + algoritmo) que representam problemas com perfis distintos de complexidade.
As seen in figures \ref{fig_resume_c0} to \ref{fig_resume_c3}, the results found by \textit{eXirt} in the process of creating explanations based on feature relevance rank are different from the results found by other XAI methods in the literature, even in the face of models (dataset + algorithm) that represent problems with different profiles.

%Nos resultados aqui apresentados, focou-se nas comparações de ranques de relevância de atributos de ranques gerados pelo \textit{eXirt} em relação aos ranques dos métodos já existentes na literatura, porém uma comparação completa de todos os pares de ranques gerados podem ser acessador na parte de Informação Suplementar E slides 7, 10, e 13.
In the results presented here, the focus was on comparing the feature relevance ranks generated by \textit{eXirt} with the ranks of existing methods in the literature, but a complete comparison of all pairs of ranks generated can be accessed in the part \textit{Supplementary Information} (E).

%De maneira indireta, os resultados também mostram as dificuldades em torno de pesquisas e avaliações de métodos que se propõe explicar modelos caixa-preta tree-ensemble utilizando ranques globais de relevância de atributos, pois não pode-se simplesmente enfatizar que um método ``A'' é pior ou melhor do que um outro método ``B''. Já que isso, somente poderia ser feito por especialistas, externos a área de computação, conhecedores de cada um dos problemas/datasets analisados.
Indirectly, the results also show the difficulties surrounding research and evaluation of methods that propose to explain tree-ensemble black-box models using global feature relevance ranks, as one cannot simply emphasize that an ``A'' method is worse or better than another ``B'' method. Since this could only be done by specialists, external to the computing area, knowledgeable about each of the analyzed problems/datasets. 

%O resultados também mostram os problemas que os métodos de XAI geram ao explicar modelos caixa-preta tree-ensemble, pois nem mesmo os métodos já presentes na literatura --- e já comumente utilizados pela comunidade de computação --- geram explicações idênticas entre si, deixando o usuário humano que necessita de explicações com dúvidas de qual explicação confiar.
The results also show the problems and challenges that XAI methods generate when explaining tree-ensemble black-box models, since the methods present in the literature do not generate identical explanations, leaving the human user with doubts about which explanation to trust.

%%\color{cyan}
\subsubsection{Details of \textit{eXirt} global explanations}

%Até o momento, foram apresentados resultados referentes a correlações de ranques de relevância de atributo encontrados a partir de 4 diferentes modelos baseados em 41 datasets de diferentes problemas. Porém, buscando-se apresentar resultados ainda mais exclarecedores, sobre como os ranques do \textit{eXirt} podem ser similares ou mesmo diferentes dos ranques gerados por outros métodos XAI, foram escolhidos 2 modelos, da etapa anteriorer de resultados, a fim de ser apresentar de maneira visual como os ranques são compostos.
At this time, we have presented results referring to correlations of feature relevance ranks found from 4 different models based on 41 datasets of different problems. Therefore, seeking to present more detailed results, on how the \textit{eXirt} ranks can be similar or even different in relation to the ranks generated by other XAI methods, we have chosen 2 models with explanations more reliable, from the previous stage of results in order to be presented in a manner visual as the ranks are composed.

%Para isso, selecionou-se o modelo ``M2'' que se basea no dataset ``credit-g'' do cluster ``C0'' e também o modelo ``m1'' que se basea no dataset ``diabetes'' do cluster ``C3'', pelo fato de gerarem explicações mais estáveis e representarem problemas do mundo real.
For this, the ``M2'' model was selected which is based on the ``credit-g'' dataset of the ``C0'' cluster and also the ``M1'' model which is based on the ``diabetes'' dataset of the cluster ``C3'', because they presented more reliable explanations (according to previous analyses).

\begin{figure*}[!h]
\begin{center}
\includegraphics[scale=0.6]{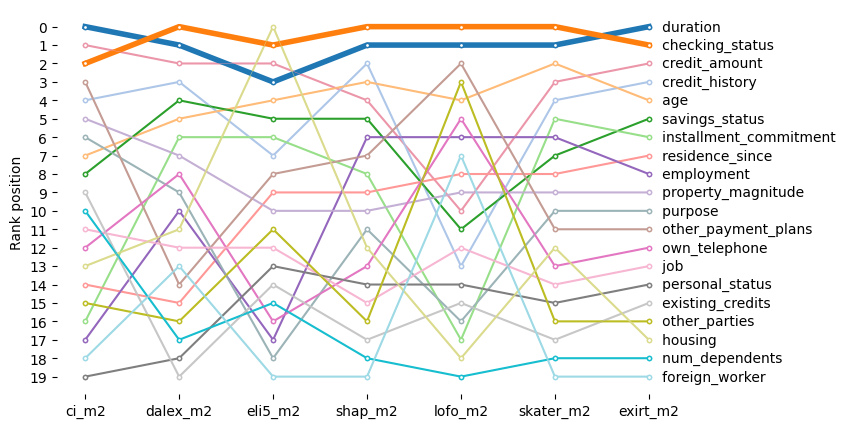}
\caption{Comparisons of feature relevance ranks generated by each XAI method from the ``M2'' model and ``credit-g'' dataset (cluster ``C0'').}
\label{fig_ranks_credit-g}
\end{center}
\end{figure*}

%Na figura \ref{fig_ranks_credit-g} é apresentada uma comparação visual dos ranques de relevância de atributos gerados a partir da aplicação dos métodos de XAI \textit{eXirt, skater, lofo, shap, eli5, dalex} e \textit{ci} no modelo ``M2'' que é baseado no dataset ``credit-g'' (cluster ``C0''). No eixo ``x'' são apresentados os métodos em ordem crescente de correlação com o \textit{eXirt} (usando resultados apresentado na figura \ref{fig_resume_c0} modelo ``M2''). Já no eixo ``y'' são apresentadoas as posições de cada feature nos ranques gerados.
Figure \ref{fig_ranks_credit-g} presents a visual comparison of feature relevance ranks generated from the application of XAI methods \textit{eXirt, skater, lofo, shap, eli5, dalex,} and \textit{ci } in model ``M2'' which is based on dataset ``credit-g'' (cluster ``C0''). On the ``x'' axis, methods are presented in ascending order of correlation with the \textit{eXirt} (using results presented in figure \ref{fig_resume_c0} model ``M2''). On the ``y'' axis, the positions of each feature in the generated ranks are presented.

%De maneira geral, pelo menos duas features aparecem dentre as primeiras posições em cada um dos ranques gerados, são elas: ``duration'' e ``check-status'', figura \ref{fig_ranks_credit-g} linhas mais grossas. Estes resultados mostram, que estas duas features apresentam maiores capacidades de explicar o modelo analisado, dado que todas os métodos de XAI apontam estas como relentes (posições mais próximas a 0).
In general, at least two features appear among the first positions in each of the generated ranks, they are: ``duration'' and ``check-status'', figure \ref{fig_ranks_credit-g} thicker lines. These results show that these two features are more capable of explaining the analyzed model, given that all XAI methods point to these as relevant (positions closer to 0).

%removido
%Ainda com relação ao apresentado na figura \ref{fig_ranks_credit-g} é possível notar há elevada variação das posições de features de ranque para ranque. Isso, ainda é mais evidente ao se observar o comportamento da posições da relevância das features em posições maiores que 3.
%Still regarding figure \ref{fig_ranks_credit-g}, it is possible to notice a high variation of feature positions from rank to rank. This is even more evident when observing the behavior of the feature relevance positions in positions greater than 3.

\begin{figure*}[!h]
\begin{center}
\includegraphics[scale=0.7]{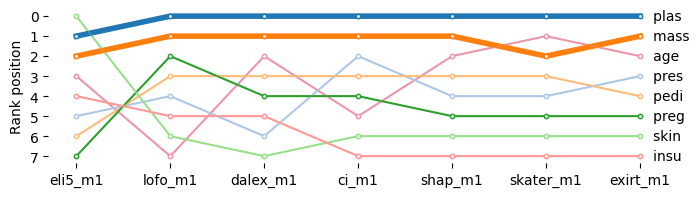}
\caption{Comparisons of feature relevance ranks generated by each XAI method from model ``M1'' and dataset ``diabetes'' (cluster ``C3'').}
\label{fig_ranks_diabetes}
\end{center}
\end{figure*}

%Na figura \ref{fig_ranks_diabetes} é apresentada novamente uma comparação visual dos ranques de relevância de atributos gerados a partir da aplicação dos métodos de XAI \textit{eXirt, skater, lofo, shap, eli5, dalex} e \textit{ci} no modelo ``M1'' que é baseado no dataset ``diabetes'' (cluster ``C3''). No eixo ``x'' são apresentados os métodos em ordem crescente de correlação com o \textit{eXirt} (usando resultados apresentado na figura \ref{fig_resume_c3} modelo ``M1''). Já no eixo ``y'' são apresentadoas as posições de cada feature nos ranques gerados.
Figure \ref{fig_ranks_diabetes} presents a visual comparison of feature relevance ranks generated from the application of the XAI methods \textit{eXirt, skater, lofo, shap, eli5, dalex,} and \textit{ci} to the ``M1'' model which is based on the ``diabetes'' dataset (cluster ``C3''). On the ``x'' axis, methods are presented in ascending order of correlation with \textit{eXirt} (using results presented in figure \ref{fig_resume_c3} model ``M1''). On the ``y'' axis, the positions of each feature in the generated ranks are presented.

%Mesmo se tratando de um modelo com menas features, figura \ref{fig_ranks_diabetes} linhas mais grossas, pode-se notar novamente que duas features se destacam aparecendo constantemente entre as principais posições de todos os ranques gerados, as features são: ``plas'' e ``mass''. Isso retrata os desafios em realizar explicações similares, mesmo para modelos que apresentam baixo número de entradas.
Even in the case of a model with fewer features, figure \ref{fig_ranks_diabetes} thick lines, it can be noted again that two features stand out constantly appearing among the main positions of all ranks generated, the features are: ``plas'' and ``mass''. This portrays the challenges in carrying out similar explanations, even for models that have a low number of entries.

%removido
%Estes resultados mais detalhados sobre as composições de cada ranque gerado, retratam as dificuldades que os métodos de XAI voltados para geração de explicação baseada em ranque de relevância de atributos apresentam, pois mesmo em situações onde existem maiores concordâncias das explicações geradas (entre diferentes métodos), percebi-se que tais explicações ainda estão longe de serem iguais.
%These more detailed results on the compositions of each rank generated, portray the difficulties and challenges that the XAI methods aimed at generating an explanation based on rank of feature relevance present, because even in situations where there is greater agreement of the explanations generated (between different methods), it was noticed that such explanations are still far from being equal.

%Em ambos os casos, pode-se notar que o \textit{eXirt} conseguiu identificar as features com maiores relevâncias, apontadas pelos métodos já existentes. Mostrando que esta metodo apresenta processos de gerações de explicações sólidos e que as explicações geradas são similares (porém, não idênticos) aos métodos já existentes na literatura.
In both cases, it can be noted that the \textit{eXirt} was able to identify the features with greater relevance, pointed out by the existing methods. Showing that this method presents solid explanation generation processes and that the generated explanations are similar (but not identical) to the existing methods in the literature.

%\color{teal}

\subsection{The \textit{eXirt} exclusive local and global explanations to trust}\label{exirt_and_trust}

%Até o presente momento, foram apresentados resultados das explicações globais geradas pelo \textit{eXirt} e os demais métodos presentes na literatura, apresentando também alguns insights sobre quais seriam as explicações mais confiáveis. Porém, a partir de agora serão apresentando resultados mais detalhados sobre os diferenciais do \textit{eXirt} em relação aos demais métodos de XAI, por meio de explicações locais baseadas na propriedade da Teoria de Resposta ao Item.
Until now, results of the global explanations generated by \textit{eXirt} and the other methods present in the literature have been presented, along with insights on which explanations would be more reliable. However, now detailed results will be presented on the \textit{eXirt} differentials, which are the local explanations based on the \textit{Item Response Theory} property.

%removido
%Os resultados apresentados a seguir mostram valores de discriminação, dificuldade e adivinhação que apontam para modelos mais estáveis e também confiáveis, que são fáceis de serem identificados devido a perspectiva do\textit{IRT}. Não serão abordadas simplesmente dados de performance do modelo, mas sim será explicado como cada modelo acerta as instâncias repassadas a ele.
%The results presented below show discrimination, difficulty and guessing values that point to more stable and also reliable models, which are easy to identify due to the \textit{IRT} perspective.

%Optou-se por selecionar 4 datasets (um de cada cluster) a fim de se diversificar os resultados (frente as diferentes propriedades de cada dataset). Os datasets escolhidos foram: ``credit-g'', ``ozone-level-8hr'', ``PhisingWebsites'', and ``diabetes''.
It was decided to select 2 datasets (randomly, one from each cluster) in order to diversify the results (in view of the different properties of each dataset). The chosen datasets were ``credit-g'' and ``diabetes''.

%Como primeira análise voltada a identificação de modelos mais estáveis e assim confiáveis, pode-se apresentar os percentuais de discriminação, dificuldade e adivinhação encontrados nas instâncias dos modelos, tabela \ref{tab_ddg}.
As a first analysis aimed at identifying more stable and reliable models, we can present the percentages of discrimination, difficulty and guesswork found in the instances of the models, Table \ref{tab_ddg}.

\begin{table}[!h]
\begin{center}
\caption{Discrimination, Difficulty and Guessing values for models ``M1'' to ``M4'' generated from 2 selected datasets.\\}
\resizebox{.75\textwidth}{!}{%
\begin{tabular}{rccc}
\hline
\shortstack{Model \\(Dataset + Algorithm)} & \multicolumn{1}{c}{Discrimination} & \multicolumn{1}{c}{Difficulty} & \multicolumn{1}{c}{Guessing} \\ \hline
Credit-g + M1                                                          & 64\%                                & 37\%                            & 38\%                         \\
Credit-g + M2                                                          & 96\%                                & 5\%                             & 15\%                         \\
Credit-g + M3                                                          & 96\%                                & 6\%                             & 15\%                         \\
Credit-g + M4                                                          & 97\%                                & 2\%                             & 22\%                         \\ %\hline
Diabetes + M1                                                          & 45\%                                & 16\%                            & 14\%                         \\
Diabetes + M2                                                          & 16\%                                & 29\%                            & 8\%                          \\
Diabetes + M3                                                          & 13\%                                & 31\%                            & 13\%                         \\
Diabetes + M4                                                          & 38\%                                & 42\%                            & 19\%                         \\ \hline
\end{tabular}
}
\label{tab_ddg}
\end{center}
\end{table}

%Nesta tabela \ref{tab_ddg}, os percentuais apresentados foram calculados com base nas estimativas dos parâmetros de item calculadas pelo \textit{eXirt} em cada execução do pipeline. Os percentuais apresentados são referentes a quantidade de instâncias com valores altos (diferentes de 0) para discriminação, dificuldade e adivinhação que foram encontrados a partir da aplicação do \textit{eXirt} no modelo.
In this Table \ref{tab_ddg}, the percentages presented were calculated based on the item parameters estimations calculated by \textit{eXirt} in each execution of the pipeline. The percentages presented refer to the number of instances with high values considering the thresholds: discrimination $> 0.75$, difficulty $> 1$, and guessing $> 0.2$, as done in \citep{cardoso2020decoding_irt}.

%Observando a tabela \ref{tab_ddg} linhas referentes aos modelos baseados no dataset ``credit-g'', pode-se identificar os modelos ``M2'' e ``M3'' como sendo os mais estáveis e confiáveis, uma vez que eles possuiram: elevados valores de discriminação (significado: as perturbações aplicadas a eles foram suficientes para distinguir-se modelos habilidosos e modelos não habilidosos dentre os respondentes), baixos valores de dificuldade (significado: os modelos apresentaram dificuldade em responder somente 5\% a 6\% das instâncias submetidas a eles), e menores valores de adivinhação (significado: estes modelos adivinharam pelo menos 15\% das instâncias submetidas a eles). Ou seja, é mais fácil um usuário humano confiar em um modelo que apresenta pouca dificuldade em responder instâncias de um dataset e também apresenta baixo acertos por sorte.
Observing the Table \ref{tab_ddg} rows referring to the models based on the ``credit-g'' dataset, the ``M2'' and ``M3'' models are identified as being the most stable and reliable, since they present: a high number of discriminative instances (meaning: the perturbations applied to them were sufficient to distinguish skilled models from non-skilled models among the respondents), few difficult instances (meaning: the models had difficulty in answering only 5\% of 6\% of the instances submitted to them), and smaller amounts of instances correct by guessing (meaning: these models guessed at least 15\% of the instances submitted to them). That is, it is easier for a human user to trust a model that presents little difficulty in responding to dataset instances and low guessing successes.

The results of the models based on the ``diabetes'' dataset, Table \ref{tab_ddg}, are presented. In them, the model ``M1'' as the most stable and reliable, as this model obtained the highest percentages of discrimination (45\%), lower percentages of difficulty (16\%) and intermediate values of guessing (14\%).

%Note, nem sempre os valores elevados de adivinhação devem ser entendidos como ruins, pois dependendo do problema a ser resolvido pelo modelo o ato de adivinhar algumas instâncias pode ser algo desejavel. Porém, esta pesquisa ressalta que valores altos de adivinhação (próximos a 50\%, por exemplo) podem significar problemas de vieses existentes no modelo.
Note, high guessing values should not always be understood as bad, because depending on the problem to be solved by the model, the act of the model guessing some instances may be desirable. However, this research points out that high guessing values (close to 50\%, for example) can mean problems with existing biases in the model.

%Apesar das análises dos percentuais de parâmetros de item ajudarem na identificação de modelos mais confiáveis, este tipo de análise apresenta limitações que são superadas a partir de análises envolvendo as Curvas Características de Itens, que será apresentada a seguir.
Although analyzes of item parameter percentages help to identify more reliable models, table \ref{tab_ddg}, this type of analysis has limitations that are overcome through analyzes involving Item Characteristic Curves, which will be presented below.

%\color{black}

\subsubsection{The \textit{eXirt} \textit{ICC} local and global explanations}

\begin{figure*}[h]
\begin{center}
\includegraphics[scale=0.75]{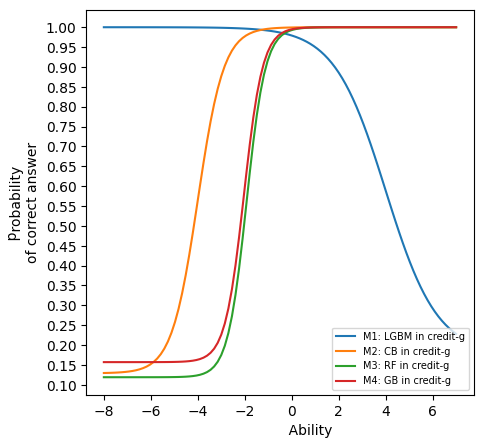}
\caption{\textit{Item Characteristic Curve} for a random instance of the ``credit-g'' dataset.}
\label{fig_icc_credit-g}
\end{center}
\end{figure*}

\begin{figure*}[!b]
\begin{center}
\hspace*{-0.3in}
\includegraphics[scale=0.6]{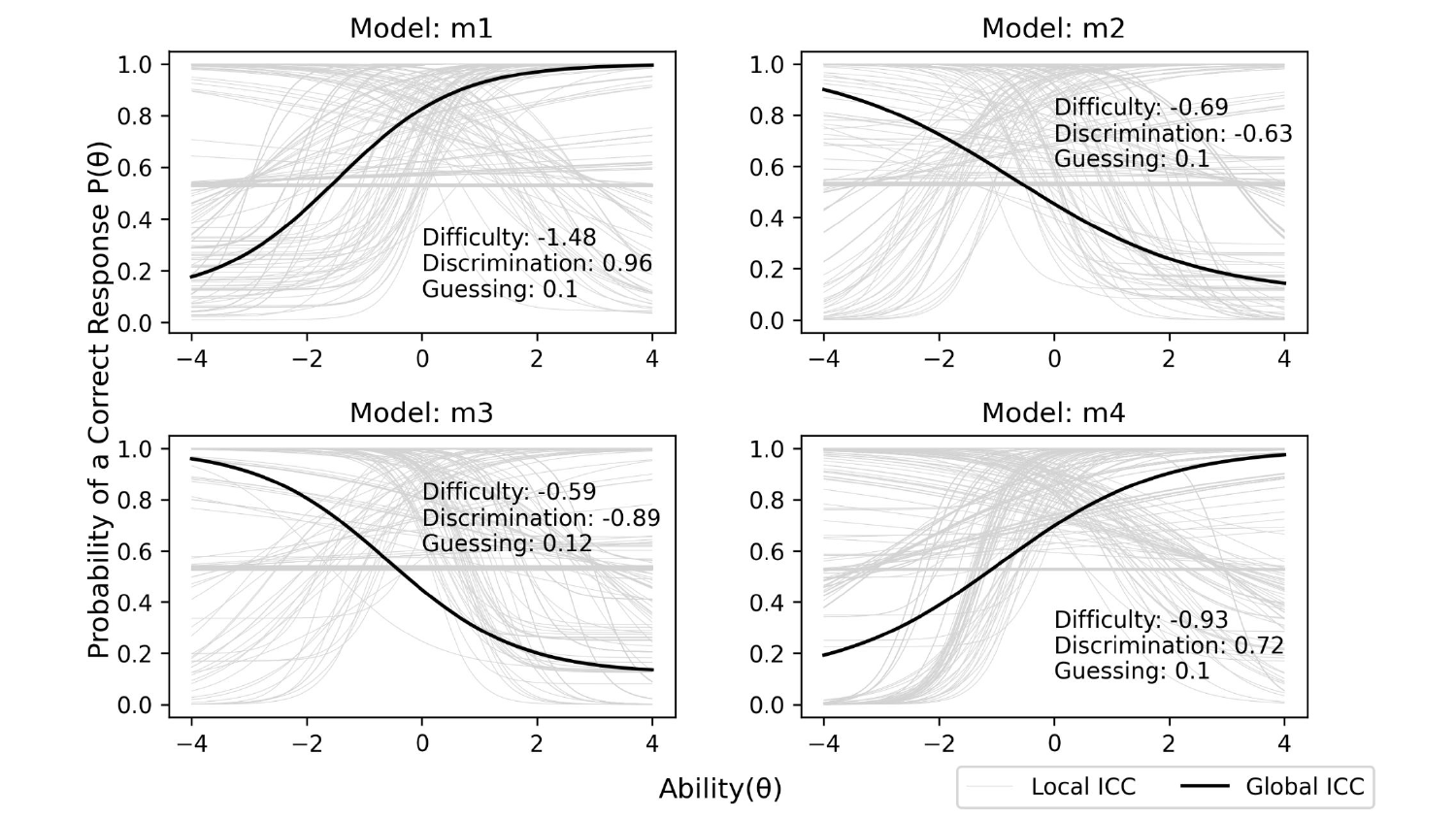}
\caption{\textit{Item Characteristic Curve (ICC)} for all instances of the ``diabetes'' dataset. The gray lines are the local explanations and the black lines (thicker) are the global explanations for each model.}
\label{fig_icc_diabetes}
\end{center}
\end{figure*}

%Buscando-se  ir ainda mais fundo em relação aos explicações locais que o \textit{eXirt} é capaz de gerar, a seguir são apresentadas as Curvas Características dos Items - ICC, a nível de instância, com os valores de discriminação, dificuldade e adivinhação para os modelos ``M1'' a ``M4'' baseados no dataset ``credit-g'', figuras \ref{fig_icc_credit-g} e \ref{fig_icc_credit-g_geral}. 
Seeking to go even deeper in relation to the local explanations that \textit{eXirt} is able to generate, the \textit{Item Characteristic Curves (ICC)} are presented below, at the instance level, with the discrimination, difficulty and guessing values for the models ``M1'' to ``M4'' based on dataset ``credit-g'', figures \ref{fig_icc_credit-g} and \ref{fig_icc_diabetes}.

%As visualizações citadas, mostram de maneira detalhada (nível de instância) como cada modelo responde as instâncias do dataset em questão. Este nível de explicação é o mais detalhado possível que o \textit{eXirt} consegur fornecer, pois permite ao usuário saber: se o modelo considera a instância discriminativa ou não, se o modelo teve dificuldade ou não pra responder a instância, ou até mesmo saber se o modelo simplesmente adivinhou determinada instância. Permitindo com isso, um maior entendimento quanto a confiança que o usuário pode empregar no modelo.
The aforementioned views show in detail (instance level) how each model responds to dataset instances. This level of explanation is the most detailed possible that \textit{eXirt} can provide, as it allows the user to know: If the model considers the instance discriminative or not; Whether or not the model had difficulty responding to the instance; Or if the model simply guessed a certain instance. Allowing with this, a greater understanding of the confidence that the user can employ in the model.

%Na figura \ref{fig_icc_credit-g}, são apresentadas 4 diferentes ICC, uma para cada modelo ``M1'' a ``M4'', de uma instância aleatória do conjunto de testes. Tomando-se como base os conceitos de\textit{IRT}, pode-se identificar que para esta instância específica:
In figure \ref{fig_icc_credit-g}, four different \textit{ICCs} are presented, one for each model ``M1'' to ``M4'', of a random instance of the test set (this specific instance is the same for all models).
Based on the concepts of \textit{IRT}, one can identify how each model predicted this instance, as follows.

%\item O modelo ``M1'', linha azul, adivinhou a instância (valores mais elevados a esqueda da alinha de ICC), não vindo apresentar valores significativos de dificuldade, e apresentando valores negativos de discriminação (valores decrescentes mais a direita), indicando que o modelo acertou a instância pela presença de algum viés existentes nela, que somente este modelo (entre os demais) identificou;
Model ``M1'',\ref{fig_icc_credit-g} (blue line), guessed the instance (higher values to the left of the ICC line), not presenting significant difficulty values, and presenting negative discrimination values (decreasing values to the right), indicating that the model hit the instance due to the presence of some existing bias in it, which only this model (among the others) identified;
    
%\item O modelo ``M2'', linha laranja, foi o modelo com um dos menores valores de adivinhação (valores mais baixos a esquerda da linha), necessitou de menos habilidade para que o modelo conseguisse alcançar o máximo de probabilidade de acerto e assim apresenta menor dificuldade (linha alcança rapidamente o topo do eixo ``y''), e apresentou um significativo valor de discriminação positiva (ângulo de inclinação ascendente da linha de ICC);
Model ``M2'', \ref{fig_icc_credit-g} (orange line), was the model with one of the lowest guessing values (lower values to the left of the ICC line), required less skill for the model to achieve the maximum probability of success and thus presents less difficulty (line quickly reaches the top of the ``y'' axis), and presented a significant value of positive discrimination (angle of upward inclination of the ICC line);
    
%\item O modelo ``M3'', linha verde, foi o modelo com o menor valor de adivinhação (valores mais baixos a esquerda da linha de ICC), necessitou de maiores valores de  habilidade para conseguir alcançar o máximo de probabilidade de acerto e assim apresenta maior dificuldade (linha alcança mais lentamente o topo do eixo ``y''), e apresentou um dos maiores valores de discriminação positiva (ângulo de inclinação ascendente da linha de ICC mais ingrime);
Model ``M3'', \ref{fig_icc_credit-g} (green line), was the model with the lowest guessing value (lower values to the left of the ICC line), needed higher skill values to achieve the maximum probability of success and therefore, it presents greater difficulty (line reaches the top of the ``y'' axis more slowly), and presented one of the highest values of positive discrimination (steepest angle of upward slope of the ICC line);

%\item O modelo ``M4'', linha vermelha, foi o modelo com o maior valor de adivinhação depois do ``M1'', necessitou de um dos maiores valores de habilidade para conseguir alcançar o máximo de probabilidade (inha alcança lentamente o topo do eixo ``y''), e apresentou valores de discriminação muito próximos aos valores encontrados pelo modelo ``M3''.
Model ``M4'', \ref{fig_icc_credit-g} (red line), was the model with the highest guess value after ``M1'', required one of the highest skill values to achieve maximum probability (line slowly reaches the top of the ``y'' axis), and presented discrimination values very close to the values found by the ``M3'' model.

%Como pode ser visto acima, as explicações a nível de instância que o \textit{eXirt} é capaz de gerar fornecem inssights que mostram características latentes do modelo analisado, principalmente quanto sua estabilidade e confiabilidade. Sendo que tais explicações, por serem a nível de instância, podem ser avaliadas para um dataset inteiro (ou mesmo splits como traino e teste).
As can be seen above, the instance-level explanations that \textit{eXirt} is able to generate insights that show latent characteristics of the analyzed model, mainly regarding its stability and reliability. Since such explanations, because they are at the instance level, can be evaluated for an entire dataset (or even splits such as training and testing).

%Ou seja, cabe ao usuário humano o julgamento de qual modelo é mais estável e confiável, tomando-se como base resultados de explicações locais gerados pelo eXirt.
That is, it is up to the human user to decide which model is more stable and reliable, based on the results of local explanations generated by eXirt.

%newnewnew
%Sabe-se que o julgamento humano, no processo da escolha de um modelo, muitas das vezes é orientado pela performance dele, porém em situações onde os valores de performances dos modelos são muito próximos, tem-se uma tarefa difícil para um humano conseguir distinguir qual é o modelo mais confiável dentre os testados.
It is known that human judgment, in the process of choosing a model, is often guided by the performance of this model, however in situations where the performance values of the models are close, it is a difficult task for a human to be able to distinguish which is the most reliable model among those tested.

%%\color{violet}
%Visando apresentar uma alternativa a este problema, o \textit{eXirt} é capaz de gerar calcular as médias dos valores de discriminação, dificuldade e adivinhação gerados durante o processo de explicação dos modelos (sobre o conjunto de teste) e assim apresentar uma explicação global, baseada na \textit{ICC} capaz de fornecer informações globais a um indivíduo humano, permitindo-o selecionar qual modelo é mais confiável.
Aiming to present an alternative to this problem, \textit{eXirt} is capable of calculating the averages of the discrimination, difficulty and guessing values generated during the model explanation process (over the test set) and thus presenting a global explanation , based on \textit{ICC} capable of providing global information to a human individual about the reliability of each model.

%Para ilustrar este formato de explicação do \textit{eXirt}, pode-se utilizar o dataset ``diabetes'', this dataset is originally from the National Institute of Diabetes and Digestive and Kidney Diseases, the objective is to predict based on diagnostic measurements whether a patient has diabetes. Os quantitativos de features e instâncias são respectivamentes 9 e 768. As performances dos modelos ``M1'', ``M2'', ``M3'' e ``M4'' criados a partir deste dataset (utilizando um split 70\%|30\% de treino de teste) são mostrados na tabela \ref{tab_diabetes}.
To illustrate this \textit{eXirt} explanation format, we can use the ``diabetes'' dataset, this dataset is originally from the National Institute of Diabetes and Digestive and Kidney Diseases, the objective is to predict based on diagnostic measurements whether a patient has diabetes. The numbers of features and instances are 9 and 768 respectively. The performances of the models ``M1'', ``M2'', ``M3'' and ``M4'' created from this dataset (using a split 70\%|30\% of test training) are shown in the table \ref{tab_diabetes}.

\begin{table}[!h]
\begin{center}
\caption{Accuracy, precision and recall data for models M1 to M4 based on the biabetes dataset.}
\resizebox{.5\textwidth}{!}{%
\begin{tabular}{cccc}
\hline
\textbf{Model} & \textbf{Accuracy} & \textbf{Precision} & \textbf{Recall} \\ \hline
M1             & 0.71              & 0.59               & 0.59            \\
M2             & 0.74              & 0.64               & 0.60             \\
M3             & 0.75              & 0.69               & 0.59            \\
M4             & 0.74              & 0.64               & 0.60             \\ \hline
\end{tabular}
}
\end{center}
\label{tab_diabetes}
\end{table}

%Conforme a tabela \ref{tab_diabetes}, os valores de accuracy, precision and recall, são relativamente próximos, mostrando uma mínima diferença entre os modelos de aprendizagem de máquina ``M1'' a ``M4''. Porém, ao se analisar os dados gerados pelo \textit{eXirt}, figura \ref{fig_icc_diabetes}, pode-se entender melhor como cada modelo aprende o conjunto de dados, permitindo-se a definição de qual deles é mais confiável. 
According to the table \ref{tab_diabetes}, the accuracy, precision and recall values are relatively close, showing a minimal difference between the machine learning models ``M1'' to ``M4''. However, when analyzing the data generated by \textit{eXirt}, figure \ref{fig_icc_diabetes}, it is possible to better understand how each model predicts the test dataset, allowing the definition of which one is more reliable.

%Como pode-se observar na figura \ref{fig_icc_diabetes}, o modelo mais confiável é o ``M1'', uma vez que ele apresentou menor média de dificuldade $= -1.48$, maior média de discriminação $=0.96$, juntamente com uma das menores médias de adivinhação $= 0.1$. Em outras palavras, este é o modelo que teve menor média de dificuldade para responder as predições, na média é o que melhor discrimina as instâncias do conjunto de teste, e acerta as instâncias fáceis do conjunto de teste (errando somente algumas mais difíceis) evidenciando que seu apredizado é sólido.
As can be seen in figure \ref{fig_icc_diabetes}, the most reliable model is ``M1'', since it presented the lowest mean difficulty $= -1.48$, the highest mean discrimination $=0.96$, along with with one of the lowest guessing averages $= 0.1$. In other words, this is the model that had the lowest average difficulty in answering the predictions, on average it is the one that best discriminates the instances of the test set, and gets the easy instances of the test set right (missing only some more difficult ones), showing that your learning is solid.

%Como segundo modelo mais confiável, destaca-se o ``M4'', uma vez que ele aprensetou a segunda menor média de dificuldade $=-0.93$, a segunda maior média de discriminação $=0.72$, juntamente com uma das menores médias de adivinhação $=0.1$.
As the second most reliable model, ``M4'' is pointed out, since it presented the second lowest average of difficulty $=-0.93$, the second highest average of discrimination $=0.72$, together with one of the lowest averages of guess $=0.1$.

%Como modelos menos confiáveis para o dataset em questão, estão os modelos ``M2'' e ``M3'',  uma vez que eles apresentam as maiores médias de dificuldade com valores respectivos iguais a $-0.69$ e $0.59$, apresentam também valores das médias de discriminação respectivamente $-0.63$ e $-0.89$ e médias de adivinhação iguais respectivamente a $0.1$ e $0.12$. Em outras palavras, apesar de terem apresentado dificuldade relativamente baixa em realizar o processo de predição, os valores de discriminação negativos indicam a possível presença de viés nos dois modelos, pois uma discriminação negativa significa modelos habilidades mais baixas tiveram maiores chances de acertar as predições do conjunto de teste.
The models ``M2'' and ``M3'' are the least reliable models for the dataset in question, since they present the highest averages of difficulty with respective values equal to $-0.69$ and $0.59$, they also present discrimination average values respectively $-0.63$ and $-0.89$ and guessing averages equal to $0.1$ and $0.12$ respectively. In other words, despite having presented relatively low difficulty in carrying out the prediction process, the negative discrimination values indicate the possible presence of bias in both models, as negative discrimination means models with lower abilities had a greater chance of getting the predictions right.

%As explicações locais e globais baseadas na \textit{Curva Caracteristica de Item} e \textit{Explainable-by-Example} fornecem informações do modelo que nenhum outro método de XAI, entre os estudos, é capaz de fornecer. Sendo assim, estes dois formatos de explicações são os grandes diferenciais deste novo método em relação aos demais, além das explicações baseadas em ranque de relevência de atributos comprovadamente com valores diferenciados. 
Local and global explanations based on \textit{Item Characteristic Curve} and \textit{Explainable-by-Example} provide model information that no other XAI method can generate. Therefore, these two formats of explanations are the main differences of this new method in relation to the others, in addition to explanations based on the relevance ranking of attributes proven to have different values.

%Os resultados apresentados nesta seção, reforçam os desafios da que a área de Explainable Artifical Intelligence enfrenta com a dificil tarefa de explicar localmente os modelos black-box tree-ensemble, já que cada modelo é capaz de generalizar o dado/problema a partir de uma perspectiva.
The results presented in this section reinforce the challenges that the Explainable Artificial Intelligence area faces with the difficult task of explaining locally the black-box tree-ensemble models, since each model is capable of generalizing the data/problem from a different perspective, whether this perspective is local or global.

%\color{black}
%%\color{cyan}
\section{Conclusions}\label{label_conclusoes}

%Diante de todas as análises realizadas, esta pesquisa atinge seu objetivo ao apresentar uma proposta inovadora de método de XAI, denominada \textit{eXirt}, que é capaz de realizar o processo de explicação de modelos de aprendizado de máquina tree-ensemble de maneira global utilizando ranques de relevância de feature e de maneira local/global utilizando \textit{Curva Característica de Item} e \textit{Explainable-by-Example}, todas baseadas na \textit{Teoria de Resposta ao Item}. 
In view of all the analyzes carried out, this research achieves its objective by presenting an innovative proposal for an XAI method, called \textit{eXirt}, which is capable of carrying out the process of explaining tree-ensemble machine learning models in a globally based on feature relevance ranks and locally/globally using \textit{Item Characteristic Curve} and \textit{Explainable-by-Example}, all based on \textit{Item Response Theory}.

%Conforme visto, recorreu-se as comparações de ranques de relevância de atributo gerados pelo eXirt e os demais métodos de XAI, visando medir de maneira quantitativa o quanto as explicações baseadas na teoria de resposta ao item são similares as explicações já existentes na atual literatura. Ficando evidente, que o \textit{eXirt} gerou explicações consideravelmente diferentes dos métodos \textit{Ciu, Dalex, Eli5, Lofo, Shap} and \textit{Skater} na maior parte dos experimentos.
As seen, we used comparisons of attribute relevance ranks generated by \textit{eXirt} and other XAI methods, aiming to quantitatively measure how similar explanations based on \textit{IRT} are to existing explanations in current literature. It is evident that \textit{eXirt} generated considerably different explanations from the \textit{Ciu, Dalex, Eli5, Lofo, Shap} and \textit{Skater} methods in a significant part of the experiments.

%O fato do eXirt mostrar diferentes e inovadoras formas de se explicar modelos caixa-preta tree-ensemble evidencia que a Teoria de Resposta ao Item foi capaz de permitir ao método eXirt a explicação de modelos através de uma perspectiva ainda não utilizada pelos demais métodos de XAI da atualidade.
The fact that \textit{eXirt} shows different and innovative ways of explaining tree-ensemble black box models, shows that \textit{IRT} was able to allow the proposed method to explain models through a perspective that has not yet been used by other current XAI methods.

%Esta pesquisa motiva a utilização das explicações geradas pelo \textit{eXirt} por parte da comunidade de machine learning, pois os ranques de relevância de feature fornecem uma visão geral de como as features de entrada são relevantes para o modelo proposto e as explicações locais e globais baseadas em na \textit{Curva Característica de Item} e \textit{Explainable-by-Example} fornecem informações que ajudam a um indivíduo humano a escolher qual modelo e até mesmo quais instâncias de predições são mais confiáveis.
This research motivates the use of \textit{eXirt} by the machine learning community, as feature relevance ranks provide an overview of how input features are relevant to the proposed model and local and global explanations based on in \textit{Item Characteristic Curve} and \textit{Explainable-by-Example} provide information that helps a human individual choose which model and even which instances of predictions are most reliable.
%%\color{black}

\section{Future works}\label{label_trabalhos_furutos}

\begin{itemize}

\item Develop an interface for the \textit{eXirt} seeking the interaction of the human with its explanations, thus enabling the creation of a collaborative explanation between man and machine.

\item Expand the analyses with \textit{eXirt} for other types of algorithms, which are not exclusively tree-ensemble by analyzing the potential of this measure to become Model Agnostic, since the whole methodology proposed by \textit{eXirt} is compatible with the nature of this type of measure.

\end{itemize}

\section*{Acknowledgements}

%Esta pesquisa reconhece o esforço de MSc. José Ribeiro (primeiro autor), como pessoa responsável por realizar o planejamento escrita e execução desta pesquisa de maneira exitoza, através da total orientação do Dr. Ronnie Alves que incançavelmente discutil os diversos pontos metodológicos em volta da pesquisa, revisando todo o trabalho até aqui. Recenhece-se também os esforços dos demais autores, em diferentes atividades relacionadas a produção deste artigo, sendo: Lucas Cardoso e Vitor Cirilo os autores responsáveis por revisões aprofundadas sobre o assunto IRT; Raíssa Silva autora responsável por aprofundadas revisões relacionadas a XAI; Nikolas Carneiro autor responsável pela revisão e modelagem das visualizações presentes no artigo.
This research recognizes the effort of MSc. José Ribeiro (first author), as the person responsible for successfully carrying out the written planning and execution of this research, through the full guidance of Dr. Ronnie Alves who tirelessly discussed the various methodological points around the research, reviewing all the work so far. We also recognize the efforts of the other authors, in different activities related to the production of this article, namely: Lucas Cardoso and Vitor Cirilo, the authors responsible for reviews on the subject of \textit{IRT}; Raíssa Silva author responsible for reviews related to \textit{XAI}; Nikolas Carneiro author responsible for reviewing and modeling the visualizations present in the article.

\section*{CRediT author statement}

\textbf{José Ribeiro:} Conceptualization, Methodology, Software, Validation, Formal analysis, Investigation, Writing - Original Draft. \textbf{Lucas Carsoso:} Formal analysis, Software, Resources, Data Curation, Writing - Review \& Editing. \textbf{Raíssa Silva:} Software, Data Curation, Writing - Review \& Editing. \textbf{Níkolas Carneiro:} Writing - Review \& Editing, Visualization. \textbf{Vitor Santos:} Writing - Review \& Editing. \textbf{Ronnie Alves:} Supervision, Writing - Review \& Editing.

\section*{Funding}

\begin{itemize}
    \item Federal Institute of Education, Science and Technology of Pará - IFPA;
    
    \item Coordination for the Improvement of Higher Education Personnel - Brazil (CAPES) [Financing Code 001];
    
    \item Vale Institute of Technology - ITV: [grant number R100603.DTL.07] and  [grant number R100603.DTL.08];

    %\item Montpellier University: [grant number XXXX];

    %\item La Ligue Contre le Cancer: [grant number XXXX].
    
\end{itemize}

%As instituições acima listadas, não interferiram no processo de submissão deste artigo ao journal.
The institutions listed above did not interfere in any part of the article's methodology and also in the process of submitting this article to the journal.

\section*{Conflicts of interest/Competing interests}

It is declared that there are no conflicts of interest between the authors and their institutions belonging to any part of this research.

%\section*{Declaration of Generative AI and AI-assisted technologies in the writing process}
%Os autores declaram a não utilização de AI Generativa na escrita deste artigo.
%No Generative AI was used in writing this article.

\section*{Ethics approval}

Not applicable.

\section*{Consent for publication}

All individuals and institutions involved in the research in question are in agreement with the publication of this article in the journal.

\section*{Supplementary information}\label{sup_info}

% \begin{itemize}
%     \item[A] - General repository of research:
%     \begin{itemize}
%         \item \url{https://github.com/josesousaribeiro/eXirt-XAI-Pipeline};
%     \end{itemize}
    
%     \item[B] - Dataset properties with normalized values:
%     \begin{itemize}
%         \item \url{https://github.com/josesousaribeiro/eXirt-XAI-Pipeline/blob/main/data/df\_dataset\_properties\_norm.csv};
%     \end{itemize}
    
%     \item[C] - Dataset properties with binarized values:
%     \begin{itemize}
%         \item \url{https://github.com/josesousaribeiro/eXirt-XAI-Pipeline/blob/main/data/df\_dataset\_properties\_binarized.csv};
%     \end{itemize}
    
%     \item[D] - Repository for reproducibility of pipeline:
%     \begin{itemize}
%         \item \url{https://github.com/josesousaribeiro/eXirt-XAI-Pipeline/blob/main/code/eXirt\_pipeline\_v0\_3\_2\_m1\_to\_m4.ipynb};
%     \end{itemize}
    
%     \item[E] - Extra analysis illustrations:
%     \begin{itemize}
%         \item \url{https://github.com/josesousaribeiro/eXirt-XAI-Pipeline/blob/main/doc/Supplementary\%20Material\%20Based\%20on\%20Illustrations.pdf}.
%     \end{itemize}
% \end{itemize}

\begin{itemize}
    \item[A] - General repository of research:
    \begin{itemize}
        \item \url{https://github.com/josesousaribeiro/eXirt-XAI-Pipeline};
    \end{itemize}
    
    \item[B] - Dataset properties with normalized values:
    \begin{itemize}
        \item \url{https://github.com/josesousaribeiro/eXirt-XAI-Pipeline/blob/main/data/df\_dataset\_properties\_norm.csv};
    \end{itemize}
    
    \item[C] - Dataset properties with binarized values:
    \begin{itemize}
        \item \url{https://github.com/josesousaribeiro/eXirt-XAI-Pipeline/blob/main/data/df\_dataset\_properties\_binarized.csv};
    \end{itemize}
    
    \item[D] - Repository for reproducibility of pipeline:
    \begin{itemize}
        \item \url{https://github.com/josesousaribeiro/eXirt-XAI-Pipeline/blob/main/code/pipeline_xai.py};
    \end{itemize}
    
    \item[E]\label{sup_E} - Extra analysis illustrations:
    \begin{itemize}
        \item \url{https://github.com/josesousaribeiro/eXirt-XAI-Pipeline/blob/main/doc/Supplementary\%20Material\%20Based\%20on\%20Illustrations.pdf}.
    \end{itemize}
    \item[F]\label{sup_f} - The \textit{eXirt} distribution:
    \begin{itemize}
        \item \url{https://github.com/josesousaribeiro/eXirt}.
    \end{itemize}
\end{itemize}


\bibliographystyle{apalike}
\bibliography{main}

\begin{thebibliography}{}

\bibitem[Abdi and Valentin, 2007]{ref_mca}
Abdi, H. and Valentin, D. (2007).
\newblock Multiple correspondence analysis.
\newblock {\em Encyclopedia of measurement and statistics}, 2(4):651--657.

\bibitem[Andrade et~al., 2000]{de2000teoria_irt_ref}
Andrade, D.~F., Tavares, H.~R., and da~Cunha~Valle, R. (2000).
\newblock Teoria da resposta ao item: conceitos e aplica{\c{c}}{\~o}es.
\newblock {\em ABE, Sao Paulo}.

\bibitem[Apley and Zhu, 2020]{alibi_ale_ref}
Apley, D.~W. and Zhu, J. (2020).
\newblock Visualizing the effects of predictor variables in black box supervised learning models.
\newblock {\em Journal of the Royal Statistical Society: Series B (Statistical Methodology)}, 82(4):1059--1086.

\bibitem[Araujo~Santos et~al., 2023]{araujo2023quest_vitor}
Araujo~Santos, V.~C., Cardoso, L., and Alves, R. (2023).
\newblock The quest for the reliability of machine learning models in binary classification on tabular data.
\newblock {\em Scientific Reports}, 13(1):18464.

\bibitem[Arrieta et~al., 2020]{arrieta_explainable_2019_20}
Arrieta, A., Díaz-Rodríguez, N., Del~Ser, J., Bennetot, A., Tabik, S., Barbado, A., Garcia, S., Gil-Lopez, S., Molina, D., Benjamins, R., Chatila, R., and Herrera, F. (2020).
\newblock Explainable {Artificial} {Intelligence} ({XAI}): {Concepts}, taxonomies, opportunities and challenges toward responsible {AI}.
\newblock {\em Information Fusion}, 58:82--115.

\bibitem[Artusi et~al., 2002]{spearman_ref}
Artusi, R., Verderio, P., and Marubini, E. (2002).
\newblock Bravais-pearson and spearman correlation coefficients: Meaning, test of hypothesis and confidence interval.
\newblock {\em The International Journal of Biological Markers}, 17(2):148--151.
\newblock Publisher: SAGE Publications Ltd STM.

\bibitem[Arya et~al., 2020]{ibm_xai360}
Arya, V., Bellamy, R.~K., Chen, P.-Y., Dhurandhar, A., Hind, M., Hoffman, S.~C., Houde, S., Liao, Q.~V., Luss, R., Mojsilovic, A., et~al. (2020).
\newblock Ai explainability 360: An extensible toolkit for understanding data and machine learning models.
\newblock {\em J. Mach. Learn. Res.}, 21(130):1--6.

\bibitem[Baniecki et~al., 2021]{dalex_python_ref}
Baniecki, H., Kretowicz, W., Piatyszek, P., Wisniewski, J., and Biecek, P. (2021).
\newblock Dalex: responsible machine learning with interactive explainability and fairness in python.
\newblock {\em The Journal of Machine Learning Research}, 22(1):9759--9765.

\bibitem[Baylari and Montazer, 2009]{baylari2009design_irt_esa}
Baylari, A. and Montazer, G.~A. (2009).
\newblock Design a personalized e-learning system based on item response theory and artificial neural network approach.
\newblock {\em Expert Systems with Applications}, 36(4):8013--8021.

\bibitem[Biecek and Burzykowski, 2021]{dalex_book}
Biecek, P. and Burzykowski, T. (2021).
\newblock {\em Explanatory model analysis: explore, explain, and examine predictive models}.
\newblock CRC Press.

\bibitem[Biggio and Roli, 2018]{biggio2018wild}
Biggio, B. and Roli, F. (2018).
\newblock Wild patterns: Ten years after the rise of adversarial machine learning.
\newblock In {\em Proceedings of the 2018 ACM SIGSAC Conference on Computer and Communications Security}, pages 2154--2156.

\bibitem[Birnbaum, 1968]{birnbaum1968some_parameters_irt}
Birnbaum, A.~L. (1968).
\newblock Some latent trait models and their use in inferring an examinee's ability.
\newblock {\em Statistical theories of mental test scores}.

\bibitem[Breiman, 2001]{breiman2001random_perturbation_3}
Breiman, L. (2001).
\newblock Random forests.
\newblock {\em Machine learning}, 45:5--32.

\bibitem[Cardoso et~al., 2022]{cardoso2022explanationbyexample_irt}
Cardoso, L.~F., de~S.~Ribeiro, J., Santos, V. C.~A., Silva, R.~L., Mota, M.~P., Prud{\^e}ncio, R.~B., and Alves, R.~C. (2022).
\newblock Explanation-by-example based on item response theory.
\newblock In {\em Intelligent Systems: 11th Brazilian Conference, BRACIS 2022, Campinas, Brazil, November 28--December 1, 2022, Proceedings, Part I}, pages 283--297. Springer.

\bibitem[Cardoso et~al., 2020]{cardoso2020decoding_irt}
Cardoso, L.~F., Santos, V.~C., Franc{\^e}s, R. S.~K., Prud{\^e}ncio, R.~B., and Alves, R.~C. (2020).
\newblock Decoding machine learning benchmarks.
\newblock In {\em Brazilian Conference on Intelligent Systems}, pages 412--425. Springer.

\bibitem[Chadaga et~al., 2023]{chadaga2023artificial_compare}
Chadaga, K., Prabhu, S., Bhat, V., Sampathila, N., Umakanth, S., and Chadaga, R. (2023).
\newblock Artificial intelligence for diagnosis of mild--moderate covid-19 using haematological markers.
\newblock {\em Annals of Medicine}, 55(1):2233541.

\bibitem[Chang et~al., 2018]{chang2018explaining_perturbation_2}
Chang, C.-H., Creager, E., Goldenberg, A., and Duvenaud, D. (2018).
\newblock Explaining image classifiers by counterfactual generation.
\newblock In {\em International Conference on Learning Representations}.

\bibitem[Chatzimparmpas et~al., 2020]{chatzimparmpas2020state_trust}
Chatzimparmpas, A., Martins, R.~M., Jusufi, I., Kucher, K., Rossi, F., and Kerren, A. (2020).
\newblock The state of the art in enhancing trust in machine learning models with the use of visualizations.
\newblock In {\em Computer Graphics Forum}, volume~39, pages 713--756. Wiley Online Library.

\bibitem[Chen et~al., 2006]{chen2006personalized_irt_esa}
Chen, C.-M., Liu, C.-Y., and Chang, M.-H. (2006).
\newblock Personalized curriculum sequencing utilizing modified item response theory for web-based instruction.
\newblock {\em Expert Systems with applications}, 30(2):378--396.

\bibitem[Dem{\v{s}}ar, 2006]{demvsar2006statistical_compare_mult_classifier_friedman}
Dem{\v{s}}ar, J. (2006).
\newblock Statistical comparisons of classifiers over multiple data sets.
\newblock {\em The Journal of Machine learning research}, 7:1--30.

\bibitem[Durniak, 2000]{durniak2000welcome_ieeexplore_repository}
Durniak, A. (2000).
\newblock Welcome to ieee xplore.
\newblock {\em IEEE Power Engineering Review}, 20(11):12.

\bibitem[Fr{\"a}mling, 2020]{ciu_ref}
Fr{\"a}mling, K. (2020).
\newblock Decision theory meets explainable {AI}.
\newblock In {\em International Workshop on Explainable, Transparent Autonomous Agents and Multi-Agent Systems}, pages 57--74. Springer.

\bibitem[Ghahramani, 2015]{ghahramani2015probabilistic}
Ghahramani, Z. (2015).
\newblock Probabilistic machine learning and artificial intelligence.
\newblock {\em Nature}, 521(7553):452--459.

\bibitem[Ghosh et~al., 2023]{ghosh2023role_esa}
Ghosh, I., Alfaro-Cort{\'e}s, E., G{\'a}mez, M., and Garc{\'\i}a-Rubio, N. (2023).
\newblock Role of proliferation covid-19 media chatter in predicting indian stock market: Integrated framework of nonlinear feature transformation and advanced ai.
\newblock {\em Expert Systems with Applications}, 219:119695.

\bibitem[Guidotti et~al., 2018]{guidotti2018survey}
Guidotti, R., Monreale, A., Ruggieri, S., Turini, F., Giannotti, F., and Pedreschi, D. (2018).
\newblock A survey of methods for explaining black box models.
\newblock {\em ACM computing surveys (CSUR)}, 51(5):1--42.

\bibitem[Gunning and Aha, 2019]{darpa_2019}
Gunning, D. and Aha, D. (2019).
\newblock {DARPA}’s {Explainable} {Artificial} {Intelligence} ({XAI}) {Program}.
\newblock {\em AI Magazine}, 40(2):44--58.
\newblock Number: 2.

\bibitem[Haffar et~al., 2022]{haffar2022explaining_ensemblerf_ai}
Haffar, R., S{\'a}nchez, D., and Domingo-Ferrer, J. (2022).
\newblock Explaining predictions and attacks in federated learning via random forests.
\newblock {\em Applied Intelligence}, pages 1--17.

\bibitem[Hambleton et~al., 1991]{hambleton1991fundamentals_irt_2}
Hambleton, R.~K., Swaminathan, H., and Rogers, H.~J. (1991).
\newblock {\em Fundamentals of item response theory}, volume~2.
\newblock Sage.

\bibitem[Hariharan et~al., 2023]{hariharan2023xai_compare}
Hariharan, S., Rejimol~Robinson, R., Prasad, R.~R., Thomas, C., and Balakrishnan, N. (2023).
\newblock Xai for intrusion detection system: comparing explanations based on global and local scope.
\newblock {\em Journal of Computer Virology and Hacking Techniques}, 19(2):217--239.

\bibitem[Holzinger et~al., 2020]{holzinger2020explainable}
Holzinger, A., Saranti, A., Molnar, C., Biecek, P., and Samek, W. (2020).
\newblock Explainable ai methods-a brief overview.
\newblock In {\em International Workshop on Extending Explainable AI Beyond Deep Models and Classifiers}, pages 13--38. Springer.

\bibitem[Hunter, 1998]{hunter1998sciencedirect_repository}
Hunter, K. (1998).
\newblock Sciencedirect™.
\newblock {\em The Serials Librarian}, 33(3-4):287--297.

\bibitem[Ibrahim and Shafiq, 2023]{ibrahim2023explainable_cnn}
Ibrahim, R. and Shafiq, M.~O. (2023).
\newblock Explainable convolutional neural networks: A taxonomy, review, and future directions.
\newblock {\em ACM Computing Surveys}, 55(10):1--37.

\bibitem[Jouis et~al., 2021]{jouis2021_anchors}
Jouis, G., Mouch{\`e}re, H., Picarougne, F., and Hardouin, A. (2021).
\newblock Anchors vs attention: Comparing xai on a real-life use case.
\newblock In Del~Bimbo, A., Cucchiara, R., Sclaroff, S., Farinella, G.~M., Mei, T., Bertini, M., Escalante, H.~J., and Vezzani, R., editors, {\em Pattern Recognition. ICPR International Workshops and Challenges}, pages 219--227, Cham. Springer International Publishing.

\bibitem[Karamizadeh et~al., 2013]{pca_ref}
Karamizadeh, S., Abdullah, S.~M., Manaf, A.~A., Zamani, M., and Hooman, A. (2013).
\newblock An overview of principal component analysis.
\newblock {\em Journal of Signal and Information Processing}, 4(3B):173.

\bibitem[Keeney et~al., 1993]{decisions_1993}
Keeney, R.~L., L, K.~R., and Howard, R. (1993).
\newblock {\em Decisions with {Multiple} {Objectives}: {Preferences} and {Value} {Trade}-{Offs}}.
\newblock Cambridge University Press, Cambridge England ; New York, NY, USA, revised edition edition.

\bibitem[Khan, 2022]{khan2022model_specific}
Khan, A. (2022).
\newblock Model-specific explainable artificial intelligence techniques: State-of-the-art, advantages and limitations.

\bibitem[Kim et~al., 2016]{kim2016examples}
Kim, B., Khanna, R., and Koyejo, O.~O. (2016).
\newblock Examples are not enough, learn to criticize! criticism for interpretability.
\newblock {\em Advances in neural information processing systems}, 29.

\bibitem[Kline et~al., 2021]{kline2021item_irt_ml_4_raissa}
Kline, A.~S., Kline, T.~J., and Lee, J. (2021).
\newblock Item response theory as a feature selection and interpretation tool in the context of machine learning.
\newblock {\em Medical \& Biological Engineering \& Computing}, 59(2):471--482.

\bibitem[Koh and Liang, 2017]{koh2017understanding}
Koh, P.~W. and Liang, P. (2017).
\newblock Understanding black-box predictions via influence functions.
\newblock In {\em International conference on machine learning}, pages 1885--1894. PMLR.

\bibitem[Korobov and Lopuhin, 2021]{eli5_ref}
Korobov, M. and Lopuhin, K. (2021).
\newblock Eli5.
\newblock https://eli5.readthedocs.io/en/latest/index.html.
\newblock {Accessed January 21, 2021}.

\bibitem[Kreiner, 2012]{kreiner2012rasch_params_item}
Kreiner, S. (2012).
\newblock {\em The Rasch Model for Dichotomous Items}, chapter~1, pages 5--26.
\newblock John Wiley \& Sons, Ltd.

\bibitem[Krishna et~al., 2022]{krishna2022disagreement}
Krishna, S., Han, T., Gu, A., Pombra, J., Jabbari, S., Wu, S., and Lakkaraju, H. (2022).
\newblock The disagreement problem in explainable machine learning: A practitioner's perspective.
\newblock {\em arXiv preprint arXiv:2202.01602}.

\bibitem[Lin et~al., 2017]{lin2017structured_attention}
Lin, Z., Feng, M., dos Santos, C., Yu, M., Xiang, B., Zhou, B., and Bengio, Y. (2017).
\newblock A structured self-attentive sentence embedding.
\newblock In {\em International Conference on Learning Representations}. International Conference on Learning Representations, ICLR.

\bibitem[Linardatos et~al., 2021]{review_xai_2021}
Linardatos, P., Papastefanopoulos, V., and Kotsiantis, S. (2021).
\newblock {Explainable AI: A Review of Machine Learning Interpretability Methods}.
\newblock {\em Entropy}, 23(1).

\bibitem[Lipovetsky and Conklin, 2001]{shap_ref}
Lipovetsky, S. and Conklin, M. (2001).
\newblock Analysis of regression in game theory approach.
\newblock {\em Applied Stochastic Models in Business and Industry}, 17(4):319--330.

\bibitem[Lord and Wingersky, 1984]{lord1984comparison}
Lord, F.~M. and Wingersky, M.~S. (1984).
\newblock Comparison of irt true-score and equipercentile observed-score" equatings".
\newblock {\em Applied Psychological Measurement}, 8(4):453--461.

\bibitem[Lundberg, 2023]{shap_doc}
Lundberg, S. (2023).
\newblock Shap {Documentation}.
\newblock https://shap.readthedocs.io/en/latest/.
\newblock original-date: 2023-01-15T08:52:28Z.

\bibitem[Lundberg et~al., 2020a]{xai_local_global_2020}
Lundberg, S.~M., Erion, G., Chen, H., DeGrave, A., Prutkin, J.~M., Nair, B., Katz, R., Himmelfarb, J., Bansal, N., and Lee, S.-I. (2020a).
\newblock From local explanations to global understanding with explainable {AI} for trees.
\newblock {\em Nature Machine Intelligence}, 2(1):56--67.
\newblock Number: 1 Publisher: Nature Publishing Group.

\bibitem[Lundberg et~al., 2020b]{tree_shap_ref}
Lundberg, S.~M., Erion, G., Chen, H., DeGrave, A., Prutkin, J.~M., Nair, B., Katz, R., Himmelfarb, J., Bansal, N., and Lee, S.-I. (2020b).
\newblock From local explanations to global understanding with explainable ai for trees.
\newblock {\em Nature Machine Intelligence}, 2(1):2522--5839.

\bibitem[Maclin and Opitz, 1997]{maclin1997empirical_bagging_boosting}
Maclin, R. and Opitz, D. (1997).
\newblock An empirical evaluation of bagging and boosting.
\newblock {\em AAAI/IAAI}, 1997:546--551.

\bibitem[Magis and Ra{\^\i}che, 2012]{catsim}
Magis, D. and Ra{\^\i}che, G. (2012).
\newblock Random generation of response patterns under computerized adaptive testing with the r package catr.
\newblock {\em Journal of Statistical Software}, 48:1--31.

\bibitem[Mart{\'\i}nez-Plumed et~al., 2016]{martinez2016making_irt_ml_2}
Mart{\'\i}nez-Plumed, F., Prud{\^e}ncio, R.~B., Mart{\'\i}nez-Us{\'o}, A., and Hern{\'a}ndez-Orallo, J. (2016).
\newblock Making sense of item response theory in machine learning.
\newblock In {\em ECAI 2016}, pages 1140--1148. IOS Press.

\bibitem[Mart{\'\i}nez-Plumed et~al., 2019]{martinez2019item_implement_irt}
Mart{\'\i}nez-Plumed, F., Prud{\^e}ncio, R.~B., Mart{\'\i}nez-Us{\'o}, A., and Hern{\'a}ndez-Orallo, J. (2019).
\newblock Item response theory in ai: Analysing machine learning classifiers at the instance level.
\newblock {\em Artificial intelligence}, 271:18--42.

\bibitem[Microsoft, 2021]{lightgbm}
Microsoft (2021).
\newblock {LightGBM}.
\newblock {https://lightgbm.readthedocs.io/en/latest/}.
\newblock {Accessed March 2, 2021}.

\bibitem[Molnar, 2020]{molnar2020interpretable}
Molnar, C. (2020).
\newblock {\em Interpretable Machine Learning}.
\newblock Lulu. com.

\bibitem[Myung, 2003]{myung2003tutorial_mle}
Myung, I.~J. (2003).
\newblock Tutorial on maximum likelihood estimation.
\newblock {\em Journal of mathematical Psychology}, 47(1):90--100.

\bibitem[Natekin and Knoll, 2013]{natekin2013gradientboost_gb}
Natekin, A. and Knoll, A. (2013).
\newblock Gradient boosting machines, a tutorial.
\newblock {\em Frontiers in neurorobotics}, 7:21.

\bibitem[Nori et~al., 2019]{interpretML_arxiv}
Nori, H., Jenkins, S., Koch, P., and Caruana, R. (2019).
\newblock Interpretml: A unified framework for machine learning interpretability.
\newblock {\em arXiv preprint arXiv:1909.09223}.

\bibitem[{OpenML}, 2021]{openml}
{OpenML} (2021).
\newblock https://www.openml.org/search?q=qualities.NumberOfClasses%3A2%2520qualities.NumberOfMissingValues%3A0\&type=data\&sort=runs\&order=desc.
\newblock Accessed March 1, 2023.

\bibitem[Oracle, 2021a]{skater_ref}
Oracle (2021a).
\newblock Skater.
\newblock https://oracle.github.io/Skater/overview.html\#skater.
\newblock {Accessed January 21, 2021}.

\bibitem[Oracle, 2021b]{skater_git}
Oracle (2021b).
\newblock Skater {Git}.
\newblock https://github.com/oracle/Skater.
\newblock {Accessed January 14, 2022}.

\bibitem[Oreski et~al., 2017]{dataset_characteristics_effects}
Oreski, D., Oreski, S., and Klicek, B. (2017).
\newblock Effects of dataset characteristics on the performance of feature selection techniques.
\newblock {\em Applied Soft Computing}, 52:109--119.

\bibitem[{Pandas Developers}, 2022]{cut_ref}
{Pandas Developers} (2022).
\newblock {Pandas - Cut}.
\newblock https://pandas.pydata.org/docs/reference/api/pandas.\\cut.html.
\newblock {Accessed August 13, 2022.}

\bibitem[Pasquali and Primi, 2003]{PASQUALI2003_tracos_latentes}
Pasquali, L. and Primi, R. (2003).
\newblock {Fundamentos da teoria da resposta ao item: TRI}.
\newblock {\em {Avaliação Psicológica: Interamerican Journal of Psychological Assessment}}, 2:99 -- 110.

\bibitem[Prud{\^e}ncio et~al., 2015]{prudencio2015analysis_irt_ml_1}
Prud{\^e}ncio, R.~B., Hern{\'a}ndez-Orallo, J., and Mart{\i}nez-Us{\'o}, A. (2015).
\newblock Analysis of instance hardness in machine learning using item response theory.
\newblock In {\em Second International Workshop on Learning over Multiple Contexts in ECML}.

\bibitem[Reza, 1994]{info_theory_1994}
Reza, F.~M. (1994).
\newblock {\em An {Introduction} to {Information} {Theory}}.
\newblock Courier Corporation.
\newblock Google-Books-ID: RtzpRAiX6OgC.

\bibitem[Ribeiro et~al., 2022]{ribeiro_pred2town_et_al_2021}
Ribeiro, J., Meneses, L., Costa, D., Miranda, W., and Alves, R. (2022).
\newblock Prediction of homicides in urban centers: A machine learning approach.
\newblock In {\em Intelligent Systems and Applications: Proceedings of the 2021 Intelligent Systems Conference (IntelliSys) Volume 3}, pages 344--361. Springer.

\bibitem[Ribeiro et~al., 2021]{ribeiro_complexity_et_al_2021}
Ribeiro, J., Silva, R., Cardoso, L., and Alves, R. (2021).
\newblock Does dataset complexity matters for model explainers?
\newblock In {\em 2021 IEEE International Conference on Big Data (Big Data)}, pages 5257--5265.

\bibitem[Ribeiro et~al., 2016]{lime_ref}
Ribeiro, M.~T., Singh, S., and Guestrin, C. (2016).
\newblock ``why should {I} trust you?'': Explaining the predictions of any classifier.
\newblock In {\em Proceedings of the 22nd {ACM} {SIGKDD} International Conference on Knowledge Discovery and Data Mining, San Francisco, CA, USA, August 13-17, 2016}, pages 1135--1144.

\bibitem[Ribeiro et~al., 2018]{anchors_aaai18}
Ribeiro, M.~T., Singh, S., and Guestrin, C. (2018).
\newblock Anchors: High-precision model-agnostic explanations.
\newblock In {\em AAAI Conference on Artificial Intelligence (AAAI)}.

\bibitem[Robnik-{\v{S}}ikonja and Bohanec, 2018]{robnik2018perturbation_1}
Robnik-{\v{S}}ikonja, M. and Bohanec, M. (2018).
\newblock Perturbation-based explanations of prediction models.
\newblock {\em Human and Machine Learning: Visible, Explainable, Trustworthy and Transparent}, pages 159--175.

\bibitem[Roseline and Geetha, 2021]{lofo_ref}
Roseline, S.~A. and Geetha, S. (2021).
\newblock Android malware detection and classification using lofo feature selection and tree-based models.
\newblock In {\em Journal of Physics: Conference Series}, volume 1911, page 012031. IOP Publishing.

\bibitem[Roth, 1988]{roth1988shapley}
Roth, A.~E. (1988).
\newblock {\em The Shapley value: essays in honor of Lloyd S. Shapley}.
\newblock Cambridge University Press.

\bibitem[Rousseeuw, 1987]{silhouettes}
Rousseeuw, P.~J. (1987).
\newblock Silhouettes: a graphical aid to the interpretation and validation of cluster analysis.
\newblock {\em Journal of computational and applied mathematics}, 20:53--65.

\bibitem[Sahatova and Balabaeva, 2022]{sahatova2022overview_comparacao_shap_lime}
Sahatova, K. and Balabaeva, K. (2022).
\newblock An overview and comparison of xai methods for object detection in computer tomography.
\newblock {\em Procedia Computer Science}, 212:209--219.

\bibitem[Samek et~al., 2021]{samek2021explaining_dnn}
Samek, W., Montavon, G., Lapuschkin, S., Anders, C.~J., and M{\"u}ller, K.-R. (2021).
\newblock Explaining deep neural networks and beyond: A review of methods and applications.
\newblock {\em Proceedings of the IEEE}, 109(3):247--278.

\bibitem[{Scikit-learn Developers}, 2021]{kmeans}
{Scikit-learn Developers} (2021).
\newblock {Scikit-learn KMeans}.
\newblock https://scikit-learn.org/stable/modules/generated/sklearn.cluster.KMeans.html.
\newblock {Accessed March 2, 2021}.

\bibitem[{Scikit-learn Developers}, 2022]{permutation_feature_importance_perturbation_3}
{Scikit-learn Developers} (2022).
\newblock {Permutation Feature Importance}.
\newblock https://scikit-learn.org/stable/modules/permutation\_importance.html\#:\~:text=The\%\\20permutation\%20feature\%20importance\%20is,model\%20depends\%20\\on\%20the\%20feature.
\newblock {Accessed April 20, 2022}.

\bibitem[Shalev-Shwartz and Ben-David, 2014]{shalev2014understanding}
Shalev-Shwartz, S. and Ben-David, S. (2014).
\newblock {\em Understanding machine learning: From theory to algorithms}.
\newblock Cambridge university press.

\bibitem[Shojaei et~al., 2023]{shojaei2023evolutionary_esa}
Shojaei, S., Abadeh, M.~S., and Momeni, Z. (2023).
\newblock An evolutionary explainable deep learning approach for alzheimer's mri classification.
\newblock {\em Expert Systems with Applications}, 220:119709.

\bibitem[Sokol and Flach, 2020]{peterflach_humanintheloop}
Sokol, K. and Flach, P. (2020).
\newblock One explanation does not fit all.
\newblock {\em KI-K{\"u}nstliche Intelligenz}, 34(2):235--250.

\bibitem[TeamHG-Memex, 2021]{eli5_git}
TeamHG-Memex (2021).
\newblock Eli5 {Git}.
\newblock {https://github.com/TeamHG-Memex/eli5}.
\newblock {A}ccessed January 21, 2021.

\bibitem[{Trends Developers}, 2023]{coogle_trends}
{Trends Developers} (2023).
\newblock Google trends survey ``{XAI}''.
\newblock https://trends.google.com.br/trends/explore?cat=174\&amp;.
\newblock {A}ccessed in December 20, 2023.

\bibitem[Vine, 2006]{vine2006googlescholar_repository}
Vine, R. (2006).
\newblock Google scholar.
\newblock {\em Journal of the Medical Library Association}, 94(1):97.

\bibitem[Wachter et~al., 2017]{wachter2017counterfactual}
Wachter, S., Mittelstadt, B., and Russell, C. (2017).
\newblock Counterfactual explanations without opening the black box: Automated decisions and the gdpr.
\newblock {\em Harv. JL \& Tech.}, 31:841.

\bibitem[Wang et~al., 2021]{wang2021trust_esa}
Wang, Y., Tian, L., and Wu, Z. (2021).
\newblock Trust modeling based on probabilistic linguistic term sets and the multimoora method.
\newblock {\em Expert Systems with Applications}, 165:113817.

\bibitem[Yandex, 2021]{catboost}
Yandex (2021).
\newblock {CatBoost}.
\newblock {https://catboost.ai}.
\newblock {Accessed March 2, 2021}.

\bibitem[Zhou, 2021]{zhou2021machine_book}
Zhou, Z.-H. (2021).
\newblock {\em Machine learning}.
\newblock Springer Nature.

\end{thebibliography}

\end{document}